\documentclass{egpubl}
\usepackage{EGVMV16}

\pdfoutput=1

\electronicVersion 

\usepackage[pdftex]{graphicx} \pdfcompresslevel=9
\PrintedOrElectronic

\usepackage[toc,page]{appendix}

\usepackage{t1enc,dfadobe}
\usepackage{egweblnk}
\usepackage{cite}

\usepackage{xspace}
\usepackage{color}
\usepackage[cmex10]{amsmath}

\usepackage{fixltx2e}
\usepackage[noend]{algpseudocode}
\MakeRobust{\Call}

\usepackage{subcaption}
\usepackage{relsize}
\usepackage{siunitx}

\newcommand{\vect}[1]{\boldsymbol{#1}}

\newcommand{\figref}[1]{Fig.~\ref{#1}}
\newcommand{\secref}[1]{\S\ref{#1}}

\newcommand{\dataset}[1]{\textsmaller{\textsc{#1}}}
\newcommand{\func}[1]{\textsc{#1}}

\newcommand{\Image}{\vect{I}}
\newcommand{\SrchImage}{\Image_s}
\newcommand{\TmplImage}{\Image_t}
\newcommand{\SrchPtch}{\vect{S}}
\newcommand{\TmplPtch}{\vect{T}}
\newcommand{\NccResp}{\vect{N}}
\newcommand{\GaborFilt}{\vect{G}}
\newcommand{\GaborFilterBank}{\vect{\mathcal{G}}}
\newcommand{\convolve}{\ast}
\newcommand{\GaborResp}{\vect{H}}
\newcommand{\Pairs}{\vect{\mathcal{P}}}

\makeatletter
\DeclareRobustCommand\onedot{\futurelet\@let@token\@onedot}
\def\@onedot{\ifx\@let@token.\else.\null\fi\xspace}
\def\eg{\emph{e.g}\onedot} 
\def\ie{\emph{i.e}\onedot} 
 
\def\etc{\emph{etc}\onedot} 
 \def\dof{d.o.f\onedot}
\def\etal{\emph{et al}\onedot}
\makeatother
\newcommand{\degree}{\ensuremath{^\circ}}

\begin{document}

\teaser{
  \includegraphics[width=\linewidth]{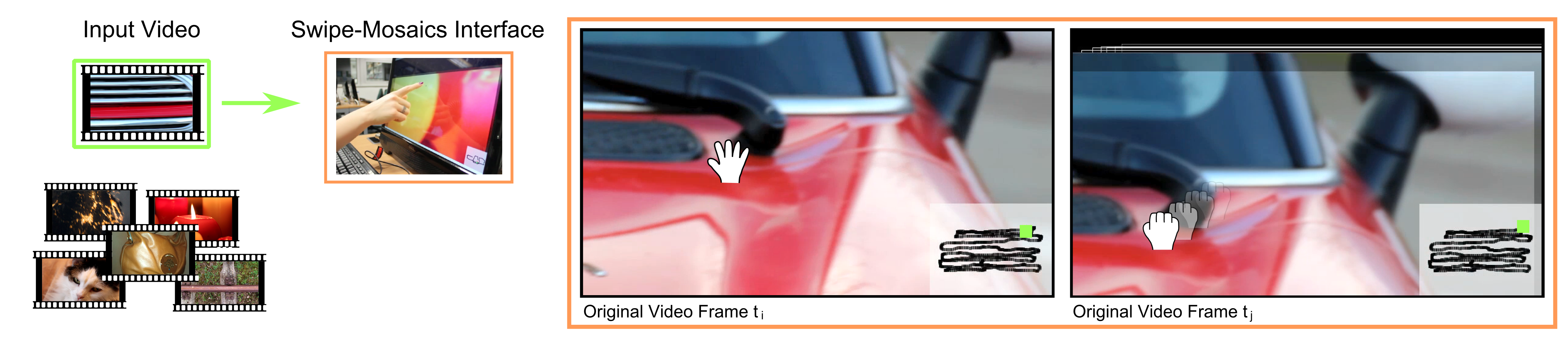}  
  \centering
  \caption{\footnotesize \emph{Left:} input videos containing ``difficult'' phenomena are used as inputs to our system.
  \emph{Right:} the Swipe Mosaic interface allowing navigation over an image sequence.
  We propose Swipe Mosaics as an algorithm and associated interface which
  composites video frames into a content-centric navigable visualization.
  Some videos can already be browsed \emph{spatially} by using existing mosaicing or IBR methods.
  Our system broadens the range of usable videos because it is trained to tolerate scene motion, parallax, repeated structure, and lack of texture.
  }
  \label{fig:bannerImage}    
}

\title{Swipe Mosaics from Video}
\author{Malcolm~Reynolds, Tom~S.~F.~Haines and~Gabriel~J.~Brostow}

\maketitle

\begin{abstract}
A panoramic image mosaic is an attractive visualization for viewing many overlapping photos, but its images must be both captured and processed correctly to produce an acceptable composite.
We propose Swipe Mosaics, an interactive visualization that places the individual video frames on a $2$D planar map that represents the layout of the physical scene.
Compared to traditional panoramic mosaics, our \emph{capture} is easier because the user can both translate the camera center and film moving subjects.
Processing and display degrade gracefully if the footage lacks distinct, overlapping, non-repeating texture.
Our proposed visual odometry algorithm produces a distribution over $(x,y)$ translations for image pairs.
Inferring a \emph{distribution} of possible camera motions allows us to better cope with parallax, lack of texture, dynamic scenes, and other phenomena that hurt deterministic reconstruction techniques.
Robustness is obtained by training on synthetic scenes with known camera motions.
We show that Swipe Mosaics are easy to generate, support a wide range of difficult scenes, and are useful for documenting a scene for closer inspection.
\end{abstract}

\section{Introduction}
The success of Microsoft's Photosynth~\cite{PhotosynthURL} demonstrates that people wish to capture environments for later navigation.
In the case of a video it is intuitive to navigate \emph{spatially}, rather than in the \emph{temporal} order it was captured.
The works of \cite{GoldmanGCSS08}, \cite{karrer2008a}, \cite{videomanip-CHI08}, and \cite{Nguyen2013direct} explored the \emph{direct manipulation} of video. 
They map a user's click-drag strokes to a sequence of frames elsewhere in the timeline (with variations).
The location of the click and the direction of the mouse indicate which pixels and what point or optical flow trajectory to query in the sequence as a whole.
 We seek a similar direct user interaction for spatial navigation of scenes, which preserves the film's points of view and the veracity of the images.
For example, imagine needing to inspect the gold handbag in \figref{fig:bannerImage} to place a bid in an online auction, or record scratches after a car accident.
For example, imagine needing to inspect the car in \figref{fig:bannerImage} to place a bid for it in an online auction, or to examine scratches after a crash.
+Our system allows casually captured video footage to be automatically converted, under some simple assumptions, into a navigable ``Swipe Mosaic''.

Image Based Rendering (IBR) techniques can be used both to composite static/dynamic mosaics~\cite{Irani:1995:MBR} and for $3$D browsing~\cite{Snavely:Phototourism:2006,Goesele:2010}.
They depend on either accurate optical flow estimation or interests points, for $2$D/$3$D pose estimation.
Both optical flow and interest points require texture.
However, many everyday scenes lack texture, or otherwise break the assumptions of current IBR and direct video manipulation systems.
Our proposed IBR approach gracefully degrades for difficult scenes, maintaining both rendering quality and user interaction.
Towards the objective of intuitively navigating video, we present the following contributions:
\begin{itemize}
  \item A regressor model trained with synthetic data, that learns the relationship between image pairs and their 2D Euclidean transform.
  
  \item A layout method that uses the probabilistic pairwise predictions from the regressor to produce a 2D location for each image, including detecting and optimizing loop-closures.
  
  \item A Swipe Mosaic interface, shown in~\figref{fig:bannerImage}, to display the video frames, allowing the user to perform content centric navigation
  and inspection by ``swiping'' scene elements.
  The interface can run as either a native application or in a web browser, and can be used on a smart phone.
  
\end{itemize}
In contrast to regular panoramic image mosaicing approaches, our system can analyze and visualize hand-held camera footage with parallax, blur, textureless areas, specular areas, and moving subjects.
The visualization quality degrades gracefully in the case of especially difficult scenes.

\section{Related Work}
\label{sec:related_work}
Since the genesis of Image Based Rendering (IBR) for synthetic data~\cite{Chen:1993}, steady progress has been made toward beautiful and useful renderings from real world footage.
Footage usually comes from multiple viewpoints, so progress is inherently dependent on having accurate estimates of relative camera poses.
Here we summarize the most relevant interactive IBR approaches, starting with techniques for estimating the needed camera parameters.

\paragraph*{Camera Poses}
A comprehensive summary of methods for converting video frames into planar and cylindrical mosaics is presented in \cite{Szeliski96videomosaics}, while \cite{Szeliski:1997:CFV} cover spherical mosaics.
They explain how stitching an image mosaic is easiest when all the images can be related to each other by homographies.
This relation can exist when the camera is translated parallel to a planar scene, or when undergoing pure rotation.
Szeliski also motivates and demonstrates robust ways of registering images to each other without matching detected interest points, such as through coarse-to-fine matching and phase correlation.
Such registration benefits from either manual or interest-point based initialization, and assumes that the scene is textured.
Textured scenes ensure convergence when minimizing the residual difference in the intensities of overlapping pixels.
Texture can also help when mosaicing an image sequence, because optical flow is strongly correlated with visual odometry~\cite{Campbell04techniquesfor}.
\cite{Peleg:2000:MAM} show that estimating camera motion and warping to enforce a consistent parallel optical flow direction allows one to combine columns of pixels onto a $2$D manifold, and not necessarily onto a planar, cylindrical, or spherical mosaic.
Optical flow estimates are most accurate when the scene is textured, and \cite{MacAodhaPAMI2012} have a helpful system to compute the uncertainty of the estimated $(u,v)$ flow components.
We too benefit from texture in the scene, but are less reliant on it.

Initializing camera poses can be difficult in practice, even in textured scenes.
Hardware attached to the camera can help~\cite{Adams:2010:Frankencam}, as demonstrated by \cite{YouNeumannGyro01} who fused visual cues with gyroscope data and \cite{klein04tightly} who used an inertial sensor to mitigate blur.
\cite{KopfGigaPixel07} actively controlled the camera pose using a motorized telescope mount to stitch mosaics of thousands of photos.
There are numerous other hybrid systems which fuse other data with images, but even \cite{KopfGigaPixel07}, \cite{WagnerMLS10}, and the Photosynth App~\cite{PhotosynthURL} rely on interest point matching to register their images.
The SIFT detection and features of \cite{Lowe04:ijcv} remain the standard by which interest point detection and matching is measured~\cite{TuytelaarsM07}.
Finding enough matching interest points in an image collection means that photos can be registered to each other, adjusted for exposure, and blended into a large mosaic~\cite{Brown03iccv}.
At least four points must be matched to compute the projective transform between two images, but in practice $10$'s and $100$'s of points are used with RANSAC~\cite{RANSAC:Fischler:1981} to robustly calculate an answer.
The same approach and inflated number of distinct interest points is normal for estimating the translation and rotation of the 2D Euclidean transform, even though two corresponding points is enough, and solutions with corresponding lines and curves also exist~\cite{HartleyZisserman}.
The key issues are that large areas of real images have light or sparse texture, and that seemingly corresponding points may not represent the same $3$D point in the world because of scene motion, motion blur, reflection, or repeated structures~\cite{Szeliski:2006:IAS}.

When building mosaics or other IBR and multi-view scene models, camera pose estimation is overwhelmingly seen as a self-contained problem.
Even \cite{Davis:1998:MSM}, whose system was designed to cope with moderately-sized moving objects and rotation-only cameras, performs global optimization by treating all the estimated pairwise camera-transforms as equally good.
In contrast, our regressor (\secref{sub:inference}) reports high uncertainty for less textured or more dynamic scenes, and the subsequent layout computation (\secref{sub:layout}) incorporates this  uncertainty.
Swipe Mosaic visualizations can better cope with difficult (though typical) footage because we work with distributions rather than committing too early to interest-point matches or specific Euclidean transform parameters.

Probabilistic distributions on locations have been applied before, such as to help a ``teleporting'' robot with a range sensor localize itself in a known floorplan~\cite{Thrun:RobustLocalization:2001}.
Probabilistic models are increasingly employed in Structure from Motion (SfM) too~\cite{davison2007monoslam}.
SfM classically requires running RANSAC over more suggested interest-point matches than the Euclidean transform (five are needed at minimum).
SfM then estimates $3$D camera poses and $3$D scene point locations, and finally optimizes these estimates globally using repeated steps of Bundle Adjustment (BA)~\cite{HartleyZisserman}.
The stages of SfM are normally deterministic and notoriously computationally expensive, but we are particularly inspired by the recent work of~\cite{Crandall:DCOforSfM:2011} who use a less costly optimization to compute an initialization for a single iteration of BA.
They convert the deterministic pairwise estimates to probabilistic constraints on a graphical model, which they solve with Loopy Belief Propagation~\cite{murphy1999loopy}.
The probabilistic approach gives a principled method of incorporating other information, such as geotags.
Instead of replacing the final half of the BA pipeline with a probabilistic system, we propose to model pose probabilistically from the beginning.

\begin{figure}[t]
  \centering
  \includegraphics[width=\columnwidth]{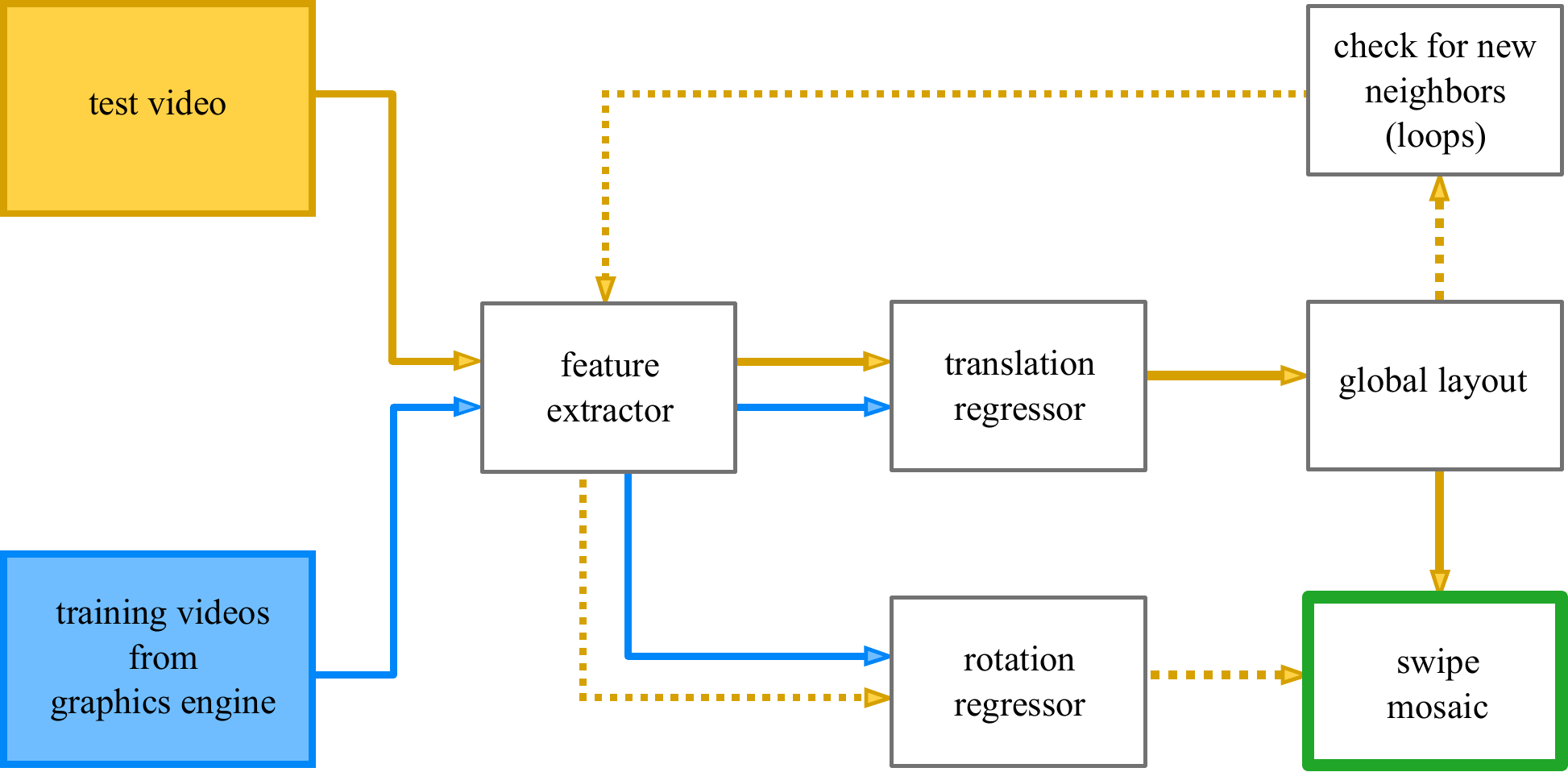}
  \caption{\footnotesize System diagram illustrating how a video is analyzed to generate a Swipe Mosaic.
  Blue lines indicate the offline training process.
  Dotted yellow lines indicate post processing steps, which take place after an initial layout is found.}
  \label{fig:overall_system_pipeline}
\end{figure}

\paragraph*{Rendering \& Interaction}
Much like the direct manipulation works mentioned already and our own interface, Dynamic Mosaics~\cite{GargS12} prominently display for interaction a current frame from the input footage.
Their rendering method occupies a middle ground between ours and that of classic image mosaics, in that they dynamically stitch onto that frame \emph{some}  spatially neighboring frames, choosing neighbors which share a large number of inlier correspondences.
This obviously limits the variety of scenes which can be displayed, so they have an alternate mode based on the similarity transform, which requires somewhat fewer correspondences.
We require no explicit correspondence points.

Interest-point based registration with subsequent Bundle Adjustment has allowed numerous interesting IBR prototypes to emerge.
Panoramic Video Textures (PVT)~\cite{agarwala2005panoramic} register and play video clips inside an otherwise static cylindrical panorama. 
A competing PVT system~\cite{rav2005dynamosaics} allows parts of the $XYT$-volume to be played back in different order, \eg making explosions look like implosions.
Also reliant on interest point matches but with an alternative optimization to BA \cite{Liu:2011:SVS} are able to stabilize shaky videos to follow different target trajectories. 

The Lumigraph~\cite{gortler1996lumigraph} and Light Field Rendering~\cite{levoy1996light} cleverly allow the user to recombine the rays captured by an array of cameras.
Interfaces allow users to navigate the plenoptic function spatially, and to simulate new focal lengths.
\cite{shum1999rendering} showed a hardware based system for capturing a reduced-size $3$D plenoptic function.
The recent system of \cite{davis2012unstructured} massively simplifies the process of capturing light fields by giving fast feedback about what parts of the static scene have been adequately filmed.
They employ the PTAM~\cite{klein07parallel} real-time SfM system which registers their cameras if enough interest points are available, and the camera does the characteristic ``SLAM wiggle''~\shortcite{Holmes:UKF:2009}.

\cite{agarwala2006photographing} discuss the differences between strip panorama systems, and  propose a multiviewpoint panorama which stitches together large regions of photos that were shot with a hand-held camera.
The strength of their interface is that users can override the stitching to (de)emphasize perspective effects in different parts of the scene.
Their system relies on the Bundler SfM system~\cite{Snavely:Phototourism:2006} for camera registration.
The Street Slide system of \cite{Kopf2010} shows another interface to multiviewpoint panoramas, which was part of our motivation for a $2$D interface.
The Photo Tourism work of \cite{Snavely:Phototourism:2006} and \cite{SGSS-siggraph08} was instrumental both for releasing Bundler and the insight that sufficiently large photo collections could be browsed in $3$D.
When images show the same objects or objects in-the-round, the viewer's transitions are rendered smoothly, and \cite{Goesele:2010} offer especially smoothed transition effects for images that are very far apart in $3$D.
These systems prefer to cull low-texture and low-quality images, and endeavor to eliminate moving objects from their collections.
In contrast, our users are filming video of something specific for interaction in a $2$D swipe interface, need \emph{that} sequence to work, and may not have the benefit of static scenes and distinct interest points.

\section{Swipe Mosaic Construction}
\label{sec:implementation}

Our system takes as an input a video sequence or temporally ordered set of images $\{\Image_1, \Image_2, \dots, \Image_N\}$ and presents them in a new type of interactive mosaic.
Valid inputs to our system include scenes which could be used to create a panoramic mosaic, but also include scenes containing significant parallax and dynamic objects, so the Swipe Mosaic avoids trying to stitch all the inputs together seamlessly.
As an overview of our approach, we first select pairs of images and make predictions of the \emph{relative} camera motion for each pair, before combining those predictions using a global least squares optimization.
Predictions are made with random regression forests, trained on synthetic data.
The predictions form a distribution over possible camera motions.
The layout algorithm locates the images on a $2$D manifold so they can be visualized using our Swipe Mosaic interface.
Finally, several postprocessing steps may be performed to further improve the viewing experience.
The overall pipeline of our visual odometry regressor and layout system is shown in \figref{fig:overall_system_pipeline}.

Pair selection generates a set $\Pairs = \{(j_1, k_1), \dots \}$, following which camera motion will be estimated between image pairs $\{(\Image_{j_1}, \Image_{k_1}), (\Image_{j_2}, \Image_{k_2}), \dots \}$.
A number of strategies can be employed to select pair indices -- some selection is necessary as comparing $O(n^2)$ pairs is computationally infeasible for large sequences.
It is possible to anticipate loops in the ordered set by finding image pairs which are temporally distant but show the same location.
Possible techniques for modeling such similarity include SIFT matching~\cite{Lowe04:ijcv}, GIST scene descriptors~\cite{oliva2006building}, simple L2 intensity distance, or geodesic distance models such as Isomap~\cite{tenenbaum2000isomap}.
We evaluated these methods but achieved superior results by initially picking only close temporal neighbors, and finding loop closures at a later stage (\secref{sub:post_processing}).

\subsection{Learning and Inference on Image Pairs}
\label{sub:inference}

We seek a probabilistic estimate of the camera motion between a pair of images.
To that end, we use Regression Random Forests (RRF)~\cite{criminisi2011decision}, in turn based on Random Forests (RF)~\cite{Bleiman:RandomForests2001, Ho:Random:1995}. 
Other supervised learning algorithms could have been used, but RRFs produce inherently probabilistic multivariate output making them an excellent fit.
Testing on unseen data produces a distribution of predictions, one from each tree. We fit a Gaussian to these predictions to obtain a parametric distribution, but in principle, the raw distribution could be used.
As well as their probabilistic nature, RRFs train and test quickly, can handle high dimensional feature vectors, and are trivially parallelizeable.
RF algorithms have been successfully applied in a range of applications, including human pose recognition~\cite{Shotton:PoseRecognition:2011} and supervised mesh segmentation~\cite{kalogerakis2010learning}.
Relative interframe motion is modeled here using the $2$D Euclidean transform, so whether training or testing, the label-space consists of three degrees of freedom: two for translation and one for rotation.
In practice, we build an RRF for translation and an essentially identical RRF for rotation to reduce the amount of training data needed.
The rotational RRF is trained to predict small camera rotations around the optical axis and is used for postprocessing.
Differences between the two RRFs are highlighted in \secref{sub:post_processing} and \secref{sec:results}.

\begin{figure}
 \centering
 \includegraphics[width=\columnwidth]{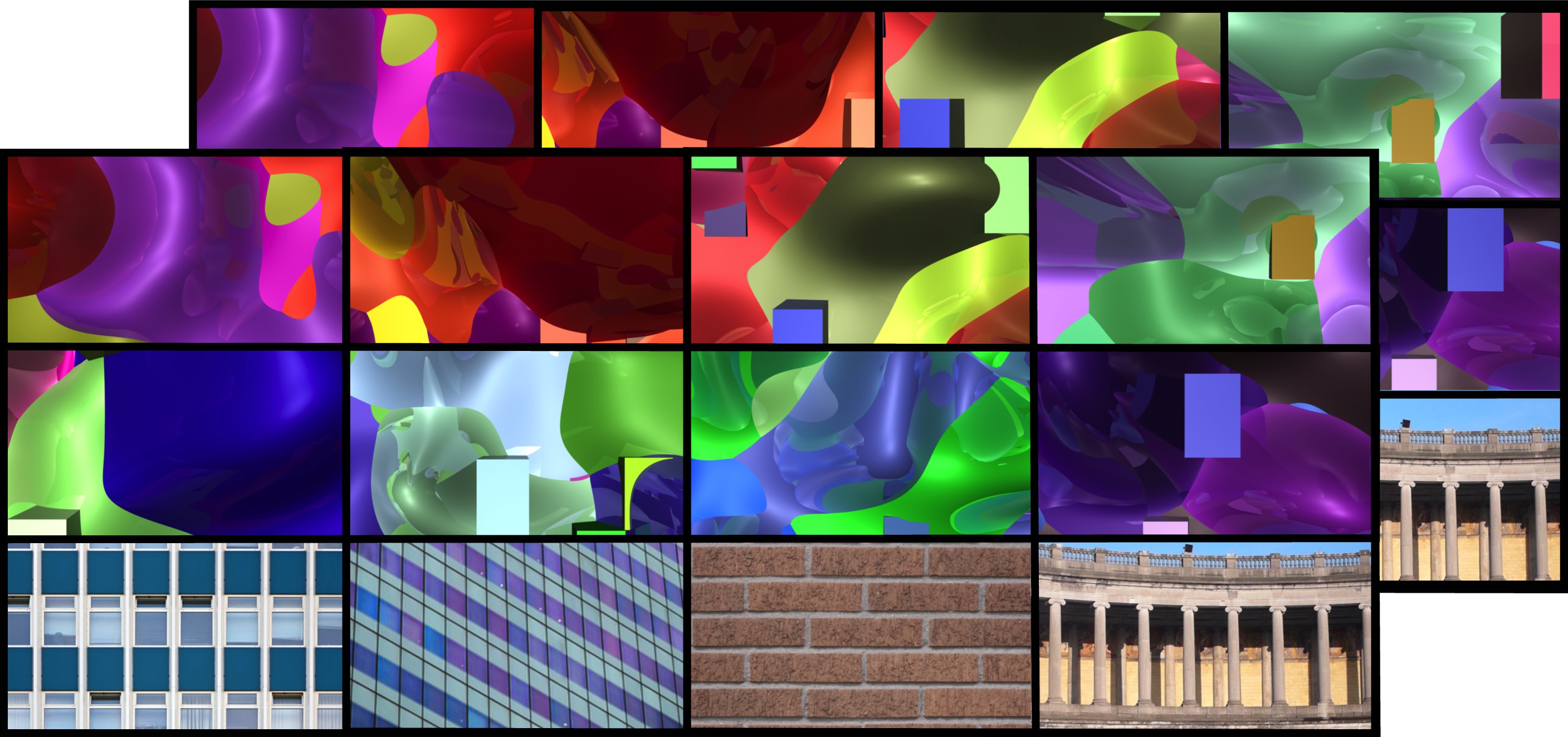}
 \caption{\footnotesize The RRF is trained on thousands of two-frame image sequences, with known camera transforms.
 To obtain sufficient quantity and variety of camera moves and scenes, we generated the training data using a custom-built but simple graphics engine.
 The top $2$ rows show a few examples of the procedurally generated scenes with depth variations and dynamic scene elements.
 The bottom row contains real images from Flickr that were mapped onto flat moving surfaces to generate training data with repeated textures.
 }
 \label{fig:training_data_examples}
\end{figure}

\subsubsection{Training Data Acquisition}
\label{ssub:training_data_acquisition}

Capturing real-world video data with ground-truth camera motion is error-prone and time consuming even with specialized equipment.
After a variety of attempts, including using multi-camera rigs and improvised outdoor motion capture, we eventually chose to generate synthetic image pairs with known camera motion.
The RRFs are able to learn how different $2$D translations and rotations appear when the world is shiny, smooth, bumpy, repetitive, and when distracting objects are moving about.
We did not render motion blur, but this is certainly possible.
Synthesizing training data with graphics techniques has previously proved successful~\cite{Shotton:PoseRecognition:2011,MacAodhaPAMI2012}, despite the obvious risk that the resulting regressor or classifier may only be accurate on artificial-looking scenes.
Aiming for large variations in shape and appearance, we rendered a family of random landscapes consisting of both angular pillars and smooth NURBS surfaces, with shape variability generated by randomly moving the pillars and deforming the surfaces.
Appearance variability was achieved by rendering each object with a random color and reflectivity.
We render two images of each landscape, with a random in-plane camera translation as the only difference between them.
The generated images include both texture rich and texture poor regions, and irregular curved edges between NURBS surfaces, which are elusive to many interest-point detectors.

A benefit of our supervised learning approach is that if deficiencies are found in the future, it is possible to augment the training set and improve model performance.
During development of our system, it was determined that regularly repeating structures posed difficulties for the system.
We augmented the training set by adding ``billboard'' datasets which replaced the random landscape previously described with a textured polygon, containing one of a set of images of repeated structure which were obtained from Flickr and other Creative Commons sources.
Example frame pairs from our training data are shown in~\figref{fig:training_data_examples}.

\subsubsection{Feature Computation}
\label{ssub:feature_computation}

\begin{figure}
 \centering
 \begin{subfigure}[b]{\columnwidth}
  \centering
  \includegraphics[height=1.9cm]{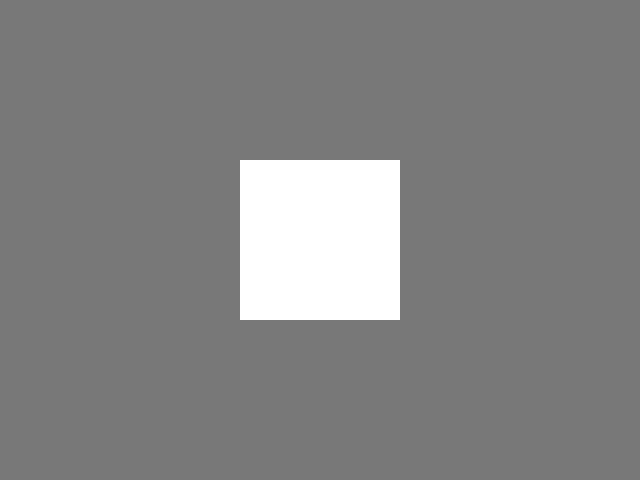}
  \includegraphics[height=1.9cm]{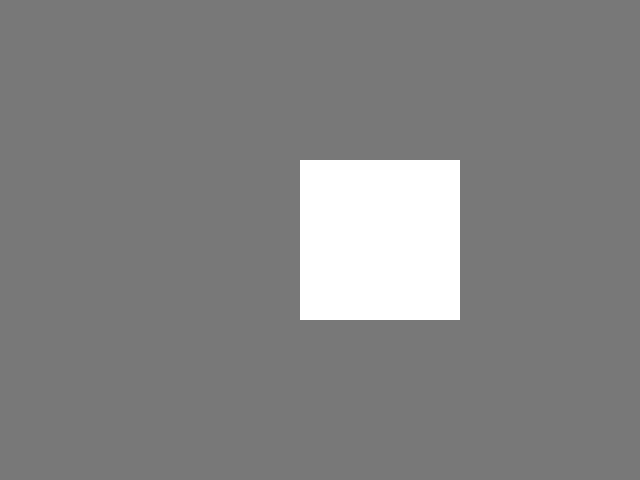}
  \includegraphics[height=1.9cm]{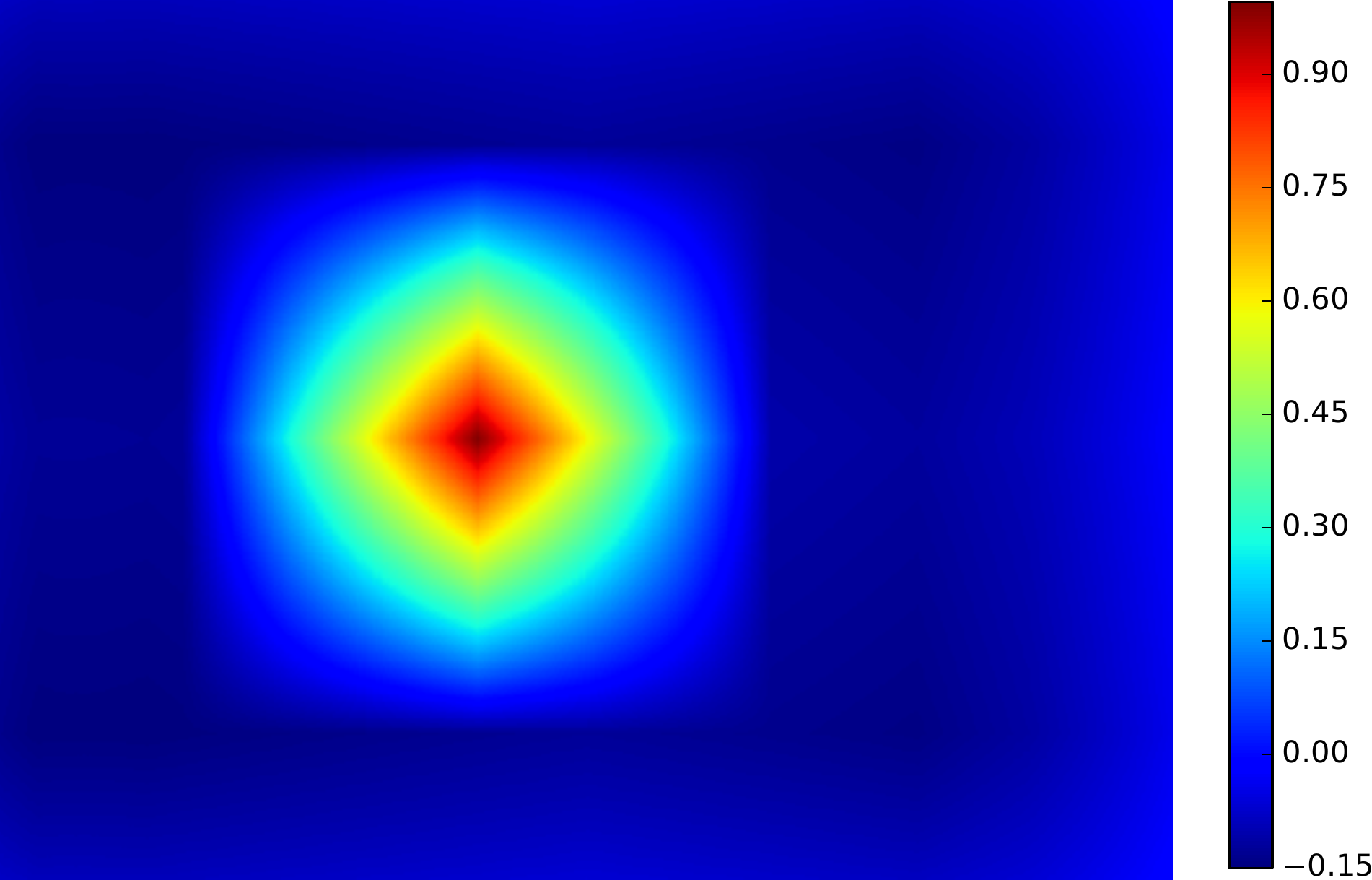}
  \caption{Image pair (left/middle) with a strong texture, producing a unimodal NCC response (right).}
  \label{fig:texture_2d}
 \end{subfigure}
 \begin{subfigure}[b]{\columnwidth}
  \centering
  \includegraphics[height=1.9cm]{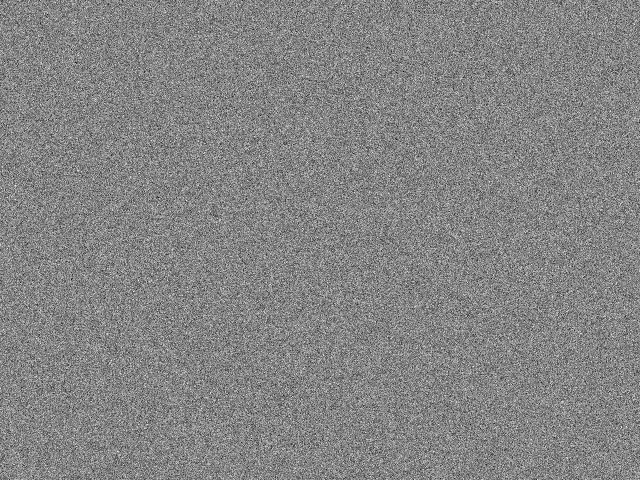}
  \includegraphics[height=1.9cm]{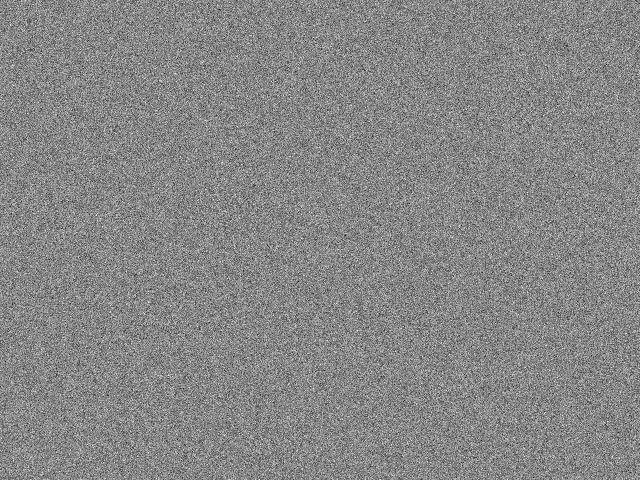}
  \includegraphics[height=1.9cm]{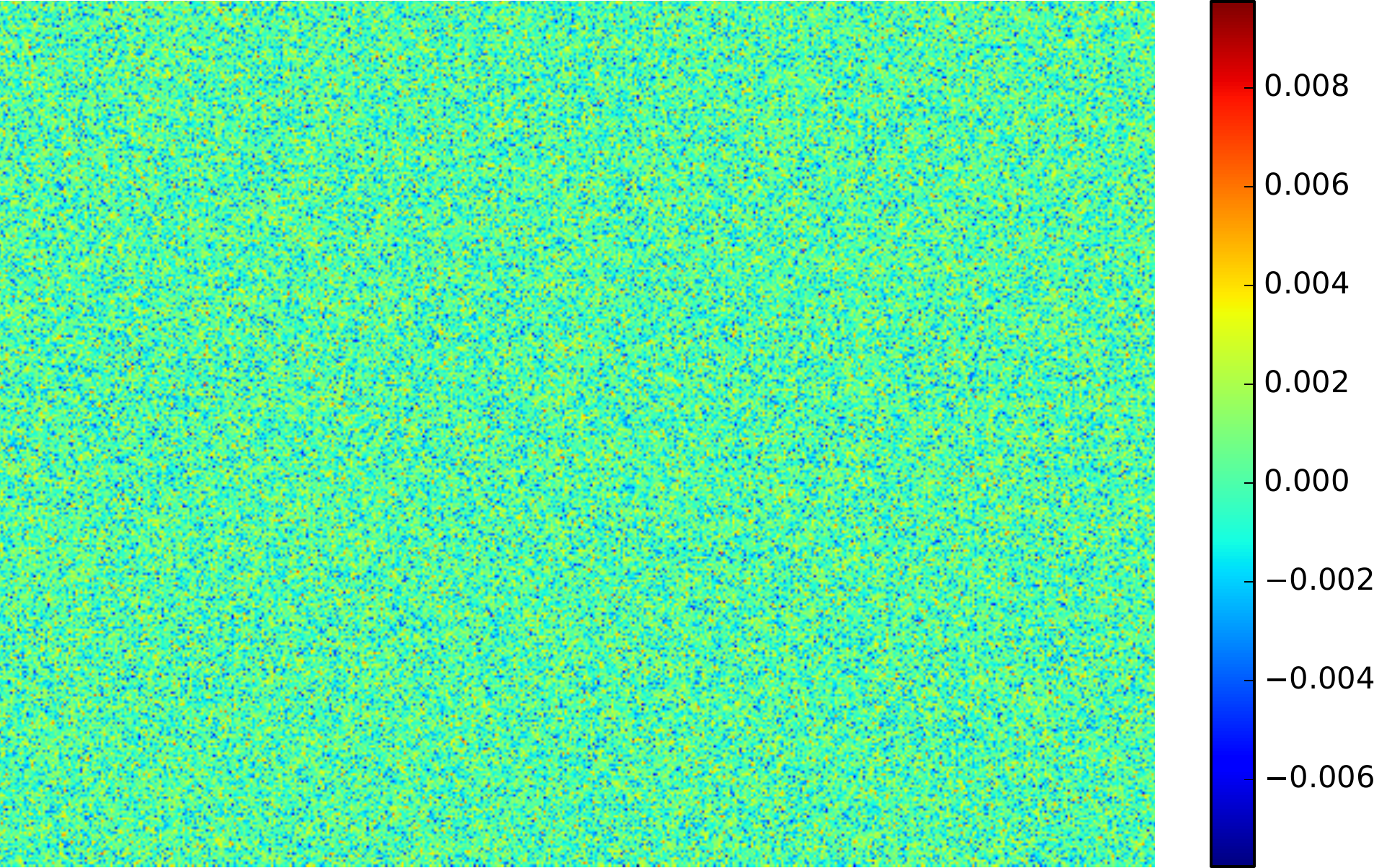}
  \caption{Textureless image pair (left/middle) producing flat NCC response (right).}
  \label{fig:texture_0d}
 \end{subfigure}
 \begin{subfigure}[b]{\columnwidth}
  \centering
  \includegraphics[height=1.9cm]{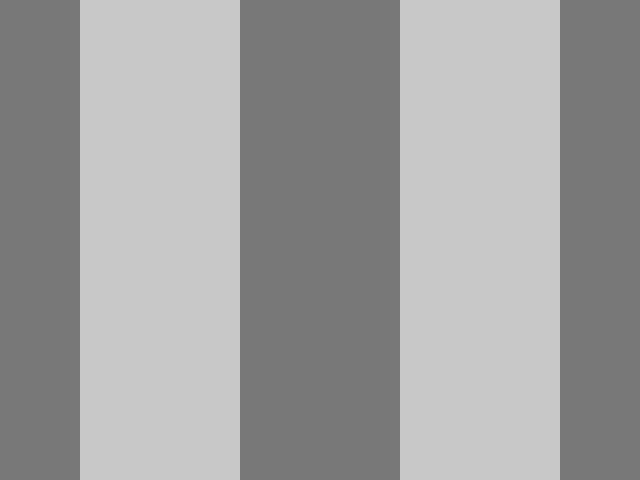}
  \includegraphics[height=1.9cm]{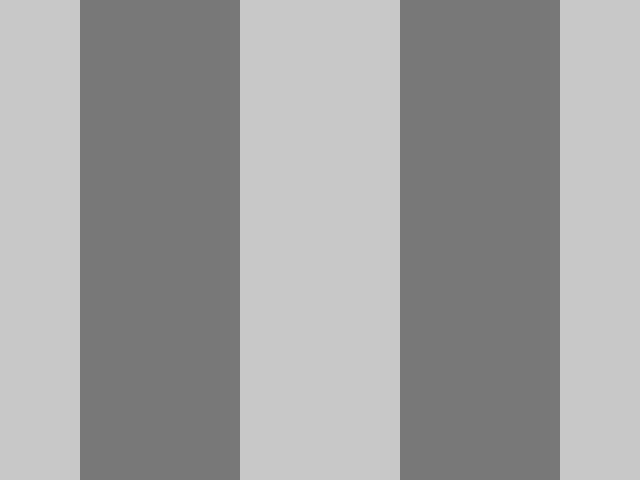}
  \includegraphics[height=1.9cm]{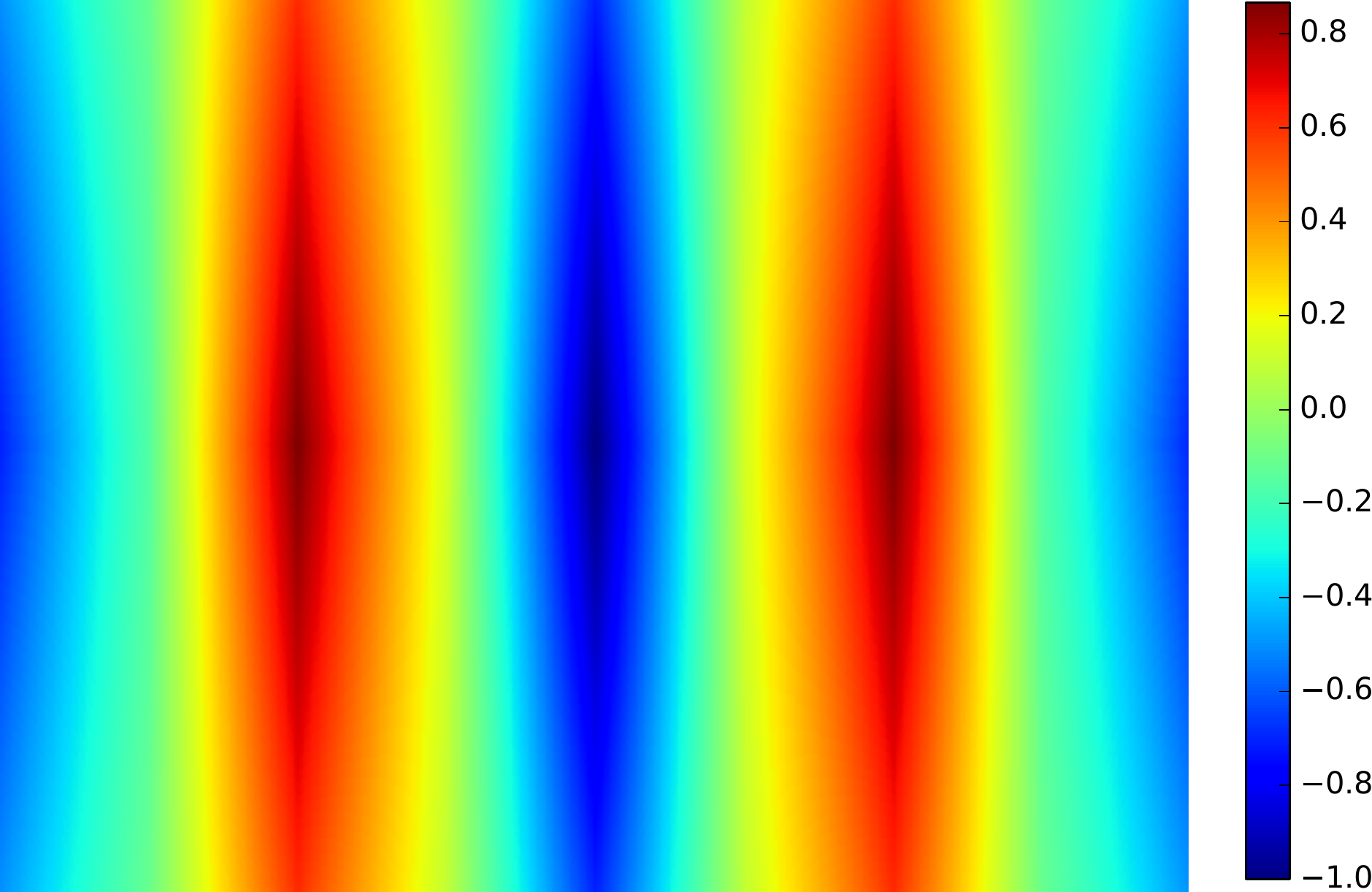}
  \caption{Image pair (left/middle) with repeated structure producing multimodal NCC response (right).}
  \label{fig:texture_md}
 \end{subfigure}
 \caption{Representative types of image pair we may see (left, middle), along with their corresponding NCC response (right).}
 \label{fig:ncc_types}
\end{figure}

A $3599$ dimensional feature vector is extracted from each image pair by encoding the responses of many Normalized Cross Correlation (NCC) comparisons between different segments of the images.
NCC was chosen because it concisely describes many properties of image pairs. 
Images containing similar, distinctive, localizable content produce unimodal NCC responses (\figref{fig:texture_2d}).
Textureless or uniform input images produce approximately flat NCC responses (\figref{fig:texture_0d}).
Images with repeated structure produce periodic NCC responses (\figref{fig:texture_md}).
Our feature extraction aims to detect and encode all these situations, allowing the RRF to learn the mapping between NCC responses and camera movement.

\begin{figure}
 \begin{center}
  \includegraphics[width=\columnwidth]{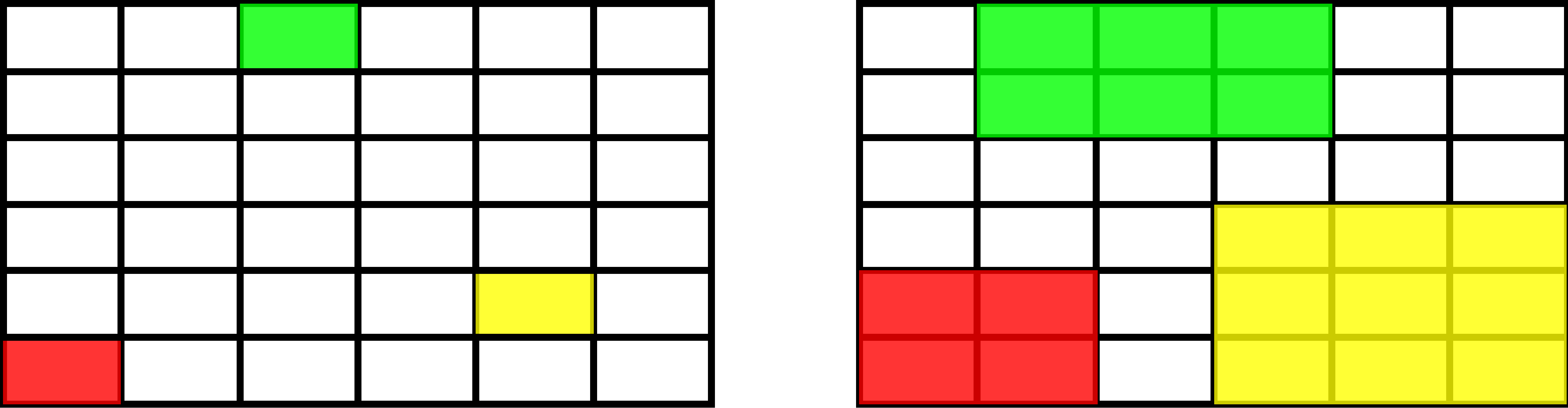}
 \end{center}
 \caption{\footnotesize template (left) and search (right) patches representing some combinations of image regions on which NCC is run ($6 \times 6$ grid level).
 Color reflects correspondence.
 Note the edge truncation behavior.}
 \label{fig:ncc_grid}
\end{figure}

For image index pair $(j,k) \in \Pairs$ we take $\Image_j$ as the template image and $\Image_k$ as the search image.
A pyramid of patches is defined by placing regular $1\times1, 2\times2, 4\times4, 6\times6$ and $8\times8$ grids onto each image (the $6\times6$ case is shown in \figref{fig:ncc_grid}).
This approach of taking patches at different scales and areas at each grid resolution in the template image is compared using NCC to a region of the search image creating a response image $\NccResp$.
To allow for scene movement between the images, each template patch is compared to a larger region in the search image, by expanding out $1$ patch in each direction unless the edge of the image prevents this.
In \figref{fig:ncc_grid} the colored patches indicate representative examples of regions that would be compared.
This first step (\func{FeatExtractImg} in \figref{alg:feature_extract}) produces 121 different NCC responses, each of which is then encoded to a few numbers, the concatenation of which forms our full feature vector.

\algrenewcommand{\algorithmiccomment}[1]{\hskip3em // #1}

\begin{figure} 
\begin{algorithmic}[0]    
    \Procedure{FeatExtractImg}{$\SrchImage, \TmplImage$}    
        \For{$l \in \{1, 2, 4, 6, 8\}$}      
            \For{$x \in \{0 \dots l-1\}$}    
                \For{$y \in \{0 \dots l-1\}$}    
                    \State $\vect{x}, \vect{y} \gets$ \Call{TemplPix}{$l, x, y, \TmplImage.shape$}    
                    \State $\hat{\vect{x}}, \hat{\vect{y}} \gets $ \Call{SearchPix}{$l, x, y, \SrchImage.shape$}    
                    \State $\TmplPtch \gets \TmplImage[\vect{y}, \vect{x}]$ 
                    \State $\SrchPtch \gets \SrchImage[\hat{\vect{y}}, \hat{\vect{x}}]$ 
                    \State $\NccResp \gets $\Call{NCC}{$\SrchPtch, \TmplPtch$} 
                    \State \Call{EncodeNCC}{$\NccResp, l, \SrchPtch.shape$} 
                \EndFor
            \EndFor    
        \EndFor    
    \EndProcedure    
    \Statex    
    \Procedure{EncodeNCC}{$\NccResp, l, shape$}    
        \State \Call{Output}{\Call{Min}{$\NccResp$},\Call{Max}{$\NccResp$},\Call{Mean}{$\NccResp$}}
        \State $x, y \gets$ \Call{PeakCoords}{$\NccResp$}    
        \State $x', y' \gets$ \Call{NormPeak}{$x, y, shape$}    
        \State \Call{Output}{$x', y'$}    
        \For{$p \in {10, 20}$}    
            \State \Call{Output}{\Call{LaplaceCoords}{$\NccResp, x, y, p$}}    
        \EndFor    
        \State \Call{Output}{\Call{NormedHist}{$\NccResp, (-1, 1), 5$}}    
        \If{$l <= 2$}    
            \For{$\GaborFilt \in \GaborFilterBank$}    
                \State $\GaborResp \gets \NccResp \convolve \GaborFilt$    
                \State $a, b \gets$ \Call{Min}{$\GaborResp$},\Call{Max}{$\GaborResp$}    
                \State $c, d \gets$ \Call{Mean}{$\GaborResp$},\Call{Median}{$\GaborResp$}    
                \State \Call{Output}{$a, b, c, d$}    
            \EndFor    
        \EndIf    
    \EndProcedure    
\end{algorithmic} 

\caption{\footnotesize Pseudocode to extract a feature vector from equal sized grayscale images $\Image_s$ and $\Image_t$.    
Indentation denotes structure, as with Python.
Zero based indexing is used.    
$\Image[\vect{y}, \vect{x}]$ is a slicing operation to extract the subwindow defined by pixel index vectors $\vect{x}$ and $\vect{y}$.
\func{TemplPix} returns the pixel indices to extract a small ``template'' window for a given level, window index and image size (\figref{fig:ncc_grid} left).    
\func{SearchPix} operates similarly but produces indices for a larger ``search'' window (\figref{fig:ncc_grid} right).    
\func{Output} appends a variable number of features to the feature vector being built for this image pair (for clarity, there is no variable to represent the feature vector in the pseudocode).
\func{PeakCoords} computes the 2D pixel shift for the peak of the NCC response.
\func{NormPeak} normalises this in relation to the template image size.
\func{LaplaceCoords} computes a Laplacian coordinate descriptor on $\NccResp$ around the peak point $(x,y)$ at a scale $p$.     
\func{NormedHist} returns a 5 bin histogram of the NCC image with bin limits $(-1, 1)$.    
}
\label{alg:feature_extract}    
\end{figure}

A strongly peaked NCC response indicates a likely offset for the scene content (\ie this portion of the scene contains localizable texture).
In this case providing the location of this offset to our machine learning system is crucial, as (for example) if every NCC comparison contained a strongly peaked offset to the left, this is strong evidence that the camera has moved to the right.
If the response is relatively flat, \ie the peak value is close to the mean value, then the patches compared are likely textureless and so no definitive decision can be made about the camera motion.
Note that this absence of certainty is an equally important input to our machine learning technique (it may make the RRF more likely to output a large variance).
The NCC image containing a ``ridge'' (peak only localizable in 1 dimension) indicates certain types of scene geometry (and certain degrees of belief about possible motion) and so must also be concisely encoded.
Each NCC response is encoded as follows (see \func{EncodeNCC} in \figref{alg:feature_extract}), and the concatenation of all these defines our feature vector.

Minimum, maximum and mean values are computed to give an idea of the response distribution, particularly how the peak value compares to the rest of the values.
The $2$D offset of the peak location is found, and the sizes of the input patches are used to convert it to a normalized offset, so that when the input patches are exactly the same and the peak indicates this, the offset $(0, 0)$ will be added to the feature vector.
The shape of the response is captured by computing Laplacian Coordinates around the peak point. 
We apply the Laplacian operator $\frac{1}{4}\begin{bmatrix}1 & -2 & 1\end{bmatrix}$ to $1$D ``slices'' through the 2D surface of $\NccResp$.
These slices all coincide with the peak, and are made at 4 different orientations and two different scales (so the points away from the peak which are multiplied by 1 are either $10$ or $20$ pixels away).
This $8$D descriptor encodes the shape of the peak - if all the numbers are large then the peak is strongly localizable in all directions.
If all the numbers are close to zero this means the peak is very wide, and if (for example) the numbers for horizontal ``slices'' are low and the numbers for vertical slices are large, the peak is a horizontal ``ridge''.
A normalized histogram with 5 equally spaced bins between -1 and 1 is also stored.

The final step, which we only carry out for the $1\times1$ and $2\times2$ grid resolutions, is to run a Gabor filter bank $\GaborFilterBank$ over the image.
The filters $\GaborFilt \in \GaborFilterBank$ vary in orientation, scale and frequency in order to try to capture the multimodal NCC response produced by images with repeating structure.
The min, max, mean and median of each Gabor response is stored, thus completing our feature encoding scheme.
The parameters used to generate the filter bank can be found in the Appendix.

\subsection{Layout to Global Coordinates}
\label{sub:layout}

The pairwise estimates provided by the Random Forest use a relative coordinate system, but to navigate images as a Swipe Mosaic we require image locations in a global coordinate system.
By limiting motion to a $2$D plane and approximating the RRF output as Gaussian, we solve this problem in closed form using linear least squares in a similar spirit to \cite{Olson2006fastiterative}.
For each $(j, k) \in \Pairs$, the RRF gives a mean $\vect{\mu}_{jk} = [\mu_{jk}^x \quad \mu_{jk}^y]^T$ and standard deviation $\vect{\sigma}_{jk} = [\sigma_{jk}^x\quad \sigma_{jk}^y]^T$.
An error function
\begin{align}
 E(\vect{x}) & = \sum_{j, k \in \Pairs}
  \left(\frac{(x_k - x_j) - \mu_{jk}^x}{\sigma_{jk}^x}\right)^2 +
  \left(\frac{(y_k - y_j) - \mu_{jk}^y}{\sigma_{jk}^y}\right)^2
 \label{eq:reg_summation}
\end{align}
is defined on the vector of all camera locations
\begin{align}
 \vect{x} &= \begin{bmatrix} x_1 & y_1 & x_2 & y_2 & \dots & x_N & y_N \end{bmatrix}
\end{align}
by summing squared differences between the \emph{actual} pairwise offsets in $\vect{x}$ and the \emph{predictions}, weighted by the prediction uncertainty.
$E(\vect{x})$ can be written as $e(\vect{x})^T e(\vect{x})$ where
\begin{equation}
 e(\vect{x}) = \begin{bmatrix}
  \frac{(x_{k_1} - x_{j_1}) - \mu_{j_1k_1}^x}{\sigma_{j_1k_1}^x} &
  \frac{(y_{k_1} - y_{j_1}) - \mu_{j_1k_1}^y}{\sigma_{j_1k_1}^y} &
  \dots
 \end{bmatrix}^T.
 \label{eq:reg_error_vector}
\end{equation}
$e(\vect{x})$ can be written as $\vect{A}\vect{x} - \vect{b}$ where each row of $\vect{A}$ contains zeros in all but two entries, at locations to match the variables in $\vect{x}$, containing positive and negative inverse standard deviation, and the corresponding element of $\vect{b}$ is the mean prediction from the forest divided by the standard deviation.
Two more rows are added to $\vect{A}$ and $\vect{b}$ to overdetermine the system by forcing $(x_1, y_1)$ to be $(0, 0)$.
The unique solution for $\vect{x}$ is determined by solving the least squares equation, $\vect{A}\vect{x} = \vect{b}$.

\subsection{Post-processing}
\label{sub:post_processing}

\subsubsection*{Translational Loop Closure}


Errors, however small, will accumulate over long sequences, so that loops in the camera path may not line up correctly.
To mitigate this we automatically find ``loop points'' -- pairs of images which are close in the $2$D coordinate space but temporally distant.
We compute each image's five nearest spatial neighbors, and any neighbors whose timestamps differ by $> n$ become loop points.
For our sequences we set $n$ at 25.
The number of pairs chosen with this technique is scene-dependent, but it is typically orders of magnitude less than the number of temporal pairs used to make the initial layout.
From each loop point image pair, we compute a feature vector and corresponding prediction from the translational RRF.
A new layout is computed from the combined set of temporal and loop-based predictions.

\subsubsection*{Rotational Correction}

The majority of video sequences will contain small variations in rotation about the optical axis, which may be difficult to see when viewing frames in order.
When our viewer (\secref{sub:viewer}) transitions across loop points containing this kind of rotation, even a few degrees disparity is enough to produce a noticeable artifact.
The visualization can render the images with rotation correction, but needs to be provided with the amount to rotate each image by.
A second ``rotational'' RRF was built using the same feature vector as before, but trained to predict optical axis rotation between two images.
Corresponding synthetic training data was rendered using the same method as for translation, with the only difference being the camera undergoing a random rotation around the optical axis instead of a random translation.

Because our training data for this RRF contains image pairs \emph{only} affected by rotation, performance was poor on images containing rotation \emph{and} translation.
To avoid this, we crop each image such that the centers of the cropped images should contain the same scene point, and thus the cropped images differ only in camera rotation.
Our 2D translation prediction from the layout algorithm is used to calculate how much to crop the images by.
As with the translational loop closure, we predict rotations for image pairs which are found at ``loop points''.
For each image pair, features are generated from the cropped images, and the rotational RRF returns a $1$D probabilistic estimate.
Relative rotation estimates are combined using an analagous technique to the least squares layout algorithm (\secref{sub:layout}).
Linear constraints encourage the relative rotational difference between two frames to match the predictions, encourage the rotations to be close to zero (this is a hard constraint for the first image only), and encourage smoothness between temporal neighbors.

\subsection{Swipe Mosaic Interface}
\label{sub:viewer}

We have implemented our viewer interface as both a native desktop application, using Python and OpenGL, and a web app using JavaScript and WebGL, allowing users to browse Swipe Mosaics without installing any software.
Whichever interface is used, Swipe Mosaics are viewed by first loading in images and camera locations.
The interface shows a single image in sharp focus at any one time, and optionally shows blurred surrounding images.
The user navigates by clicking anywhere on the image and dragging (``swiping'').
The swipe direction determines where on the $2$D map the program looks for a new frame to transition to.
If the user swipes enough to the left, we transition to rendering the neighboring frame on the right, in a manner similar to PDF viewers and services such as Google Maps.
A ``minimap'' in one corner of the interface conveys an idea of the overall layout of the scene and highlights the location of the currently displayed image.
As an alternative navigational aid, users can enter ``Picasso view'' which smoothly zooms out and displays all the images overlaid.
These images are not intended to line up perfectly as in a panoramic mosaic, but rather to provide a sense of which directions can be navigated.
While the user presses down and swipes to navigate to a new image, we render smooth transitions using alpha blended crossfades.
The on-screen positions of the current and next frames move smoothly in sync with the mouse, with the intention that if the user clicks on a particular recognizable feature in the scene, that feature will remain close to the cursor no matter where it is moved.
If a rotation vector (see \secref{sub:post_processing}) is provided to the viewer, images are rendered with the corresponding rotational correction.

\section{Experiments}
\label{sec:results}

Experiments were performed on videos filmed by our users on mobile phones, camcorders and SLRs, along with videos from other sources which were not captured with this purpose in mind.
We examine the performance of the pairwise regressor, and the overall Swipe Mosaic rendering and visualization system quality.
We perform qualitative and quantitative comparisons to validate our system's ability to handle a variety of sequences, which are listed along with their defining characteristics in Table \ref{table:test_sequences}.
A number of baseline algorithms are compared to, which represent different approaches to this task in the existing literature.

To reiterate, there are many methods that produce camera paths on simple, textured, static, planar scenes.
When they work, they too could be used to prepare a sequence for use as a Swipe Mosaic.
We compare to specific representative baseline methods to demonstrate that our approaches degrades gracefully with footage  that is less simple, in a variety of ways.

\begin{table}
 \begin{tabular}{l|l}
  Sequence & Characteristics \\ \hline
  \dataset{fence}$\ast$ & Easy sequence, abundant texture \\
  \dataset{mini} & Abundant texture \\ 
  \dataset{lobby}$\ast$ & Abundant texture, demonstrates loop closure \\
  \dataset{facade} & Abundant texture \\
  \dataset{grating} & Partially textureless \\  
  \dataset{rails} & Large scale repeated structure \\
  \dataset{skater} & Dynamic and deforming foreground object \\ 
  \dataset{flowers}$\dagger$ & Dynamic objects, non planar motion, \cite{Goldstein2012videostabilization} failure case \\ 
  \dataset{sculpture} & Little texture, motion blur \\ 
  \dataset{leaves} & Dynamic and deforming scene \\ 
  \dataset{obelisk} & Non planar camera path \\ 
  \dataset{handbag} & Dynamic specularities \\ 
  \dataset{wall} & Ambiguous motion due to repeated structure \\
  \dataset{vinyl} & Motion blur, textureless occluding object\\ 
  \dataset{iss}$\ast$ & Demonstrates Loop closure \\  
  \dataset{dino} & Moving background elements \\ 
  \dataset{prism} & Automatic gain control, CCD overload \\ 
  \dataset{aquatic}$\dagger$ & Scene from movie, contains dynamic objects + parallax \\
  \dataset{freiburg2}$\dagger$ & 6D Ground truth available \\
 \end{tabular}
 \caption{Test sequences along with their defining characteristics.
 $\ast$ - only appears in the Appendix.
 $\dagger$ - captured without intended purpose of building a Swipe Mosaic.}
 \label{table:test_sequences}
\end{table}

\subsection{Regressing Motion Between Image Pairs}

\begin{figure}
 \centering
 \begin{subfigure}[b]{0.3\columnwidth}
  \includegraphics[width=\columnwidth]{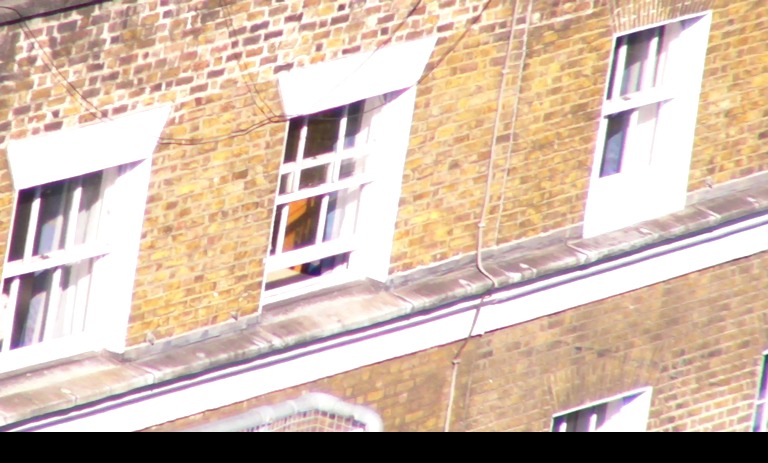}
  \includegraphics[width=\columnwidth]{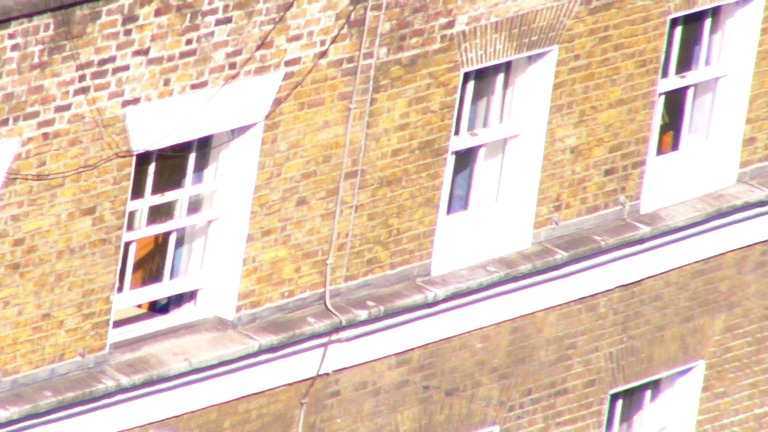}
 \end{subfigure}
 \begin{subfigure}[b]{0.3\columnwidth}
  \includegraphics[width=\columnwidth]{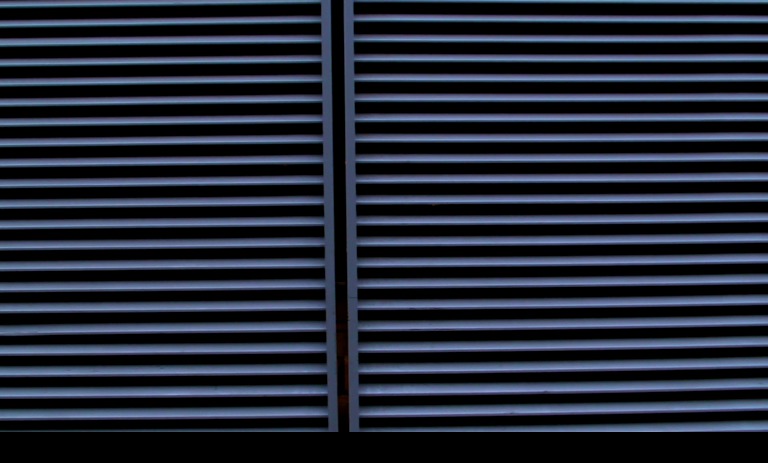}
  \includegraphics[width=\columnwidth]{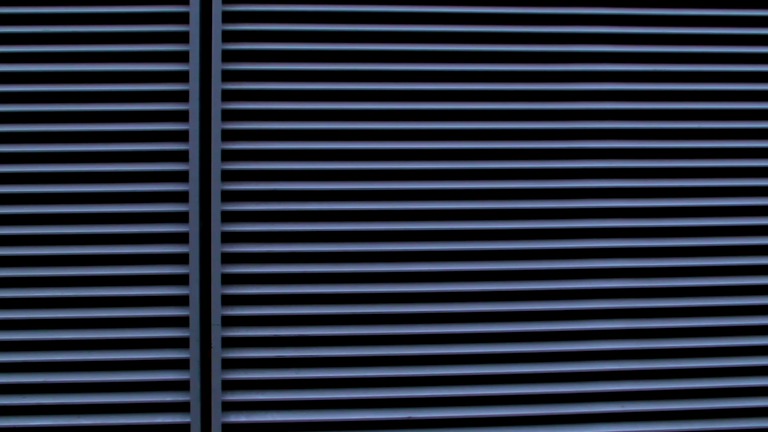}
 \end{subfigure}
 \begin{subfigure}[b]{0.3\columnwidth}
  \includegraphics[width=\columnwidth]{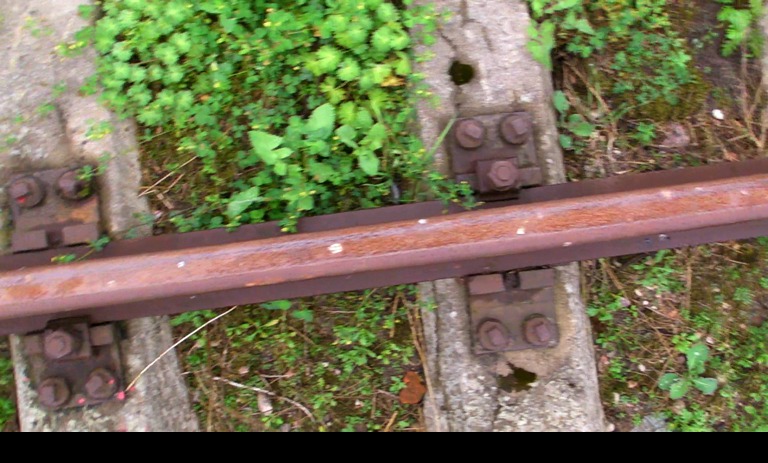}
  \includegraphics[width=\columnwidth]{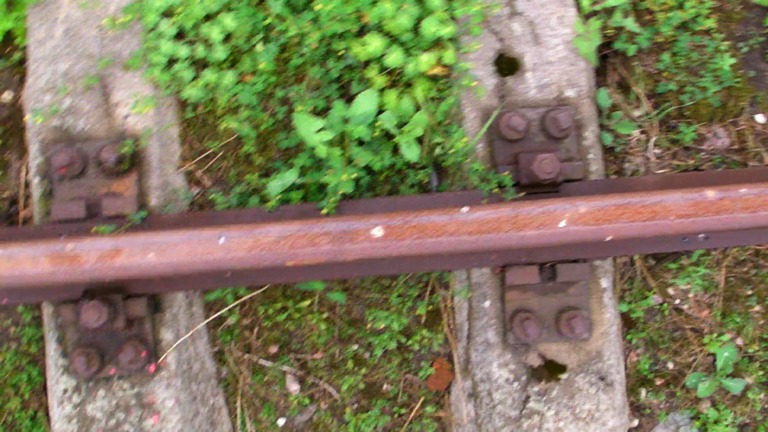}
 \end{subfigure}
 \begin{subfigure}[b]{0.3\columnwidth}
  \includegraphics[width=\columnwidth]{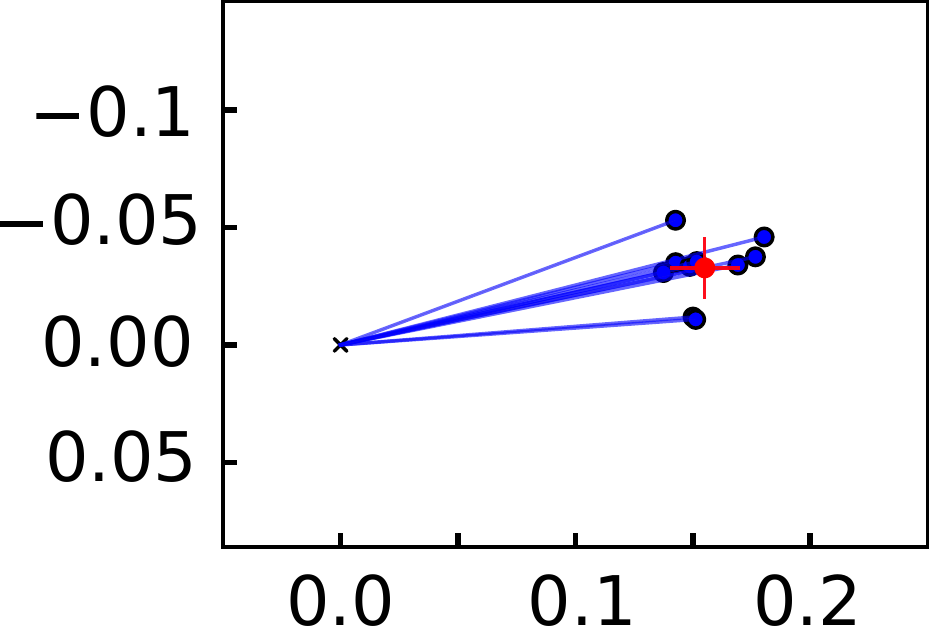}
  \caption{\footnotesize \dataset{facade}}
  \label{fig:label:trans_distribution_easy}
 \end{subfigure}
 \begin{subfigure}[b]{0.3\columnwidth}
  \includegraphics[width=\columnwidth]{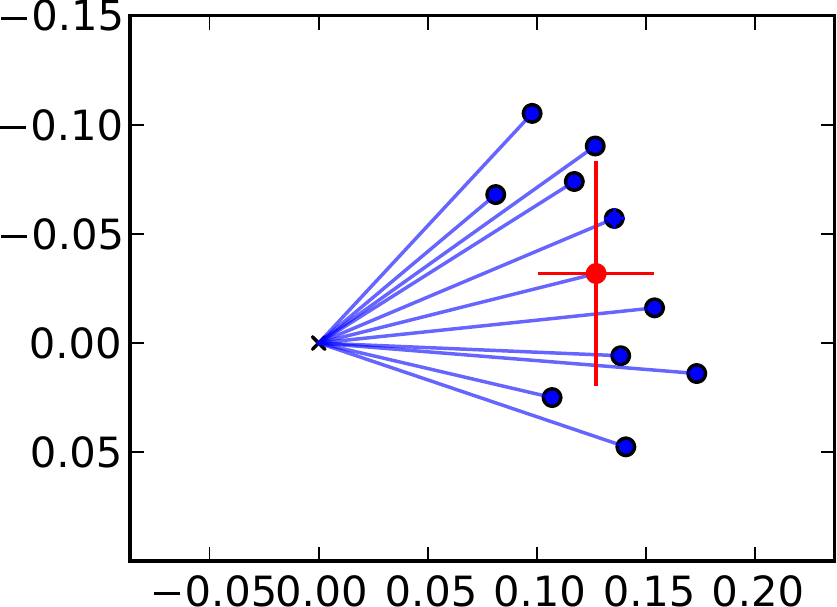}
  \caption{\footnotesize \dataset{wall}}
  \label{fig:label:trans_distribution_hard}
 \end{subfigure}
 \begin{subfigure}[b]{0.3\columnwidth}
  \includegraphics[width=\columnwidth]{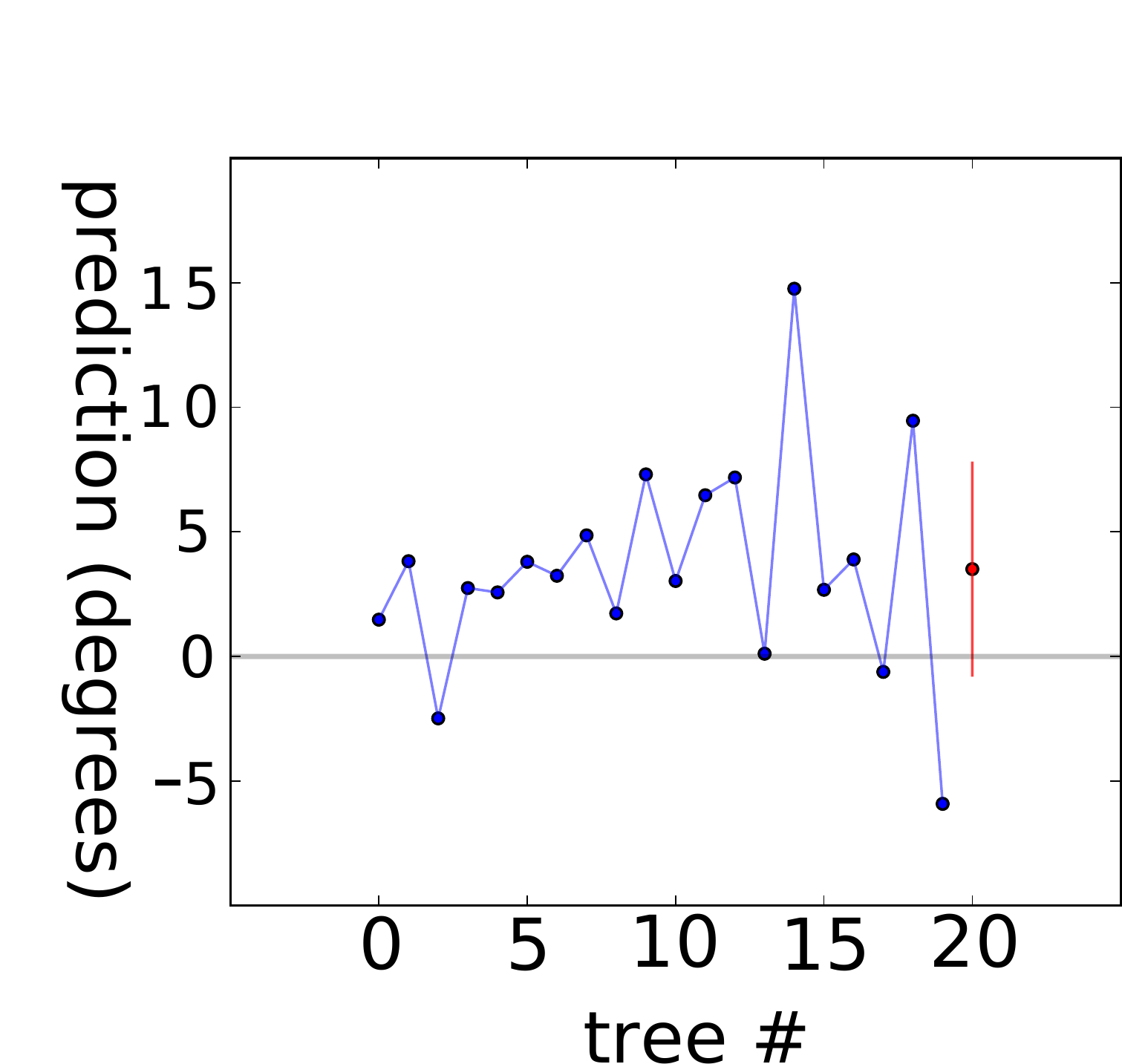}
  \caption{\footnotesize \dataset{rails}}
  \label{fig:label:rot_distribution}
 \end{subfigure}
 \caption{\footnotesize In each column the top two images were inputs resulting in the RRF output at the bottom.
 The graphs in a) and b) show $2$D results from the Translational RRF, and c) shows $1$D results from the Rotational RRF.
 Blue dots show individual tree outputs, red dots and bars show the mean and variance of the fitted Gaussian.}
 \label{fig:RegressionPredictions}
\end{figure}

We start by inspecting what the regressor has learned about the relationship between the computed features and estimating translations.
When images contain unambiguous texture (\figref{fig:label:trans_distribution_easy}) our regressor is confident in both $x$ and $y$.
For scenes with repeated texture but a unique vertical structure (\figref{fig:label:trans_distribution_hard}), the regressor outputs small $\sigma^x$ and large $\sigma^y$, reflecting the uncertainty caused by the aperture problem.
\figref{fig:label:rot_distribution} show the result of using the same type of regressor and features, but training to estimate the in-plane rotation between two images.
We expect the rotational RRF to perform better with a customized feature vector, but the vector designed for translation allowed rotational correction within $\pm5\degree$.

Initial versions of our feature extraction used Optical Flow (rather than NCC) on the input images, before condensing that information into a vector.
Building the feature vector from NCC is not an obviously better choice than using Optical Flow, but we obtained better test-time results with NCC based features.
It is likely that the NCC responses are better correlated with motion-confidence than flow, which has some estimated vector for every pixel.
Significantly, the regressor has the benefit of learning from our graphics engine: it has seen thousands of rendered examples of image pairs, with knowledge of the true $2$D Euclidean transform. 

\begin{figure}
 \begin{center}
  \includegraphics[width=\columnwidth]{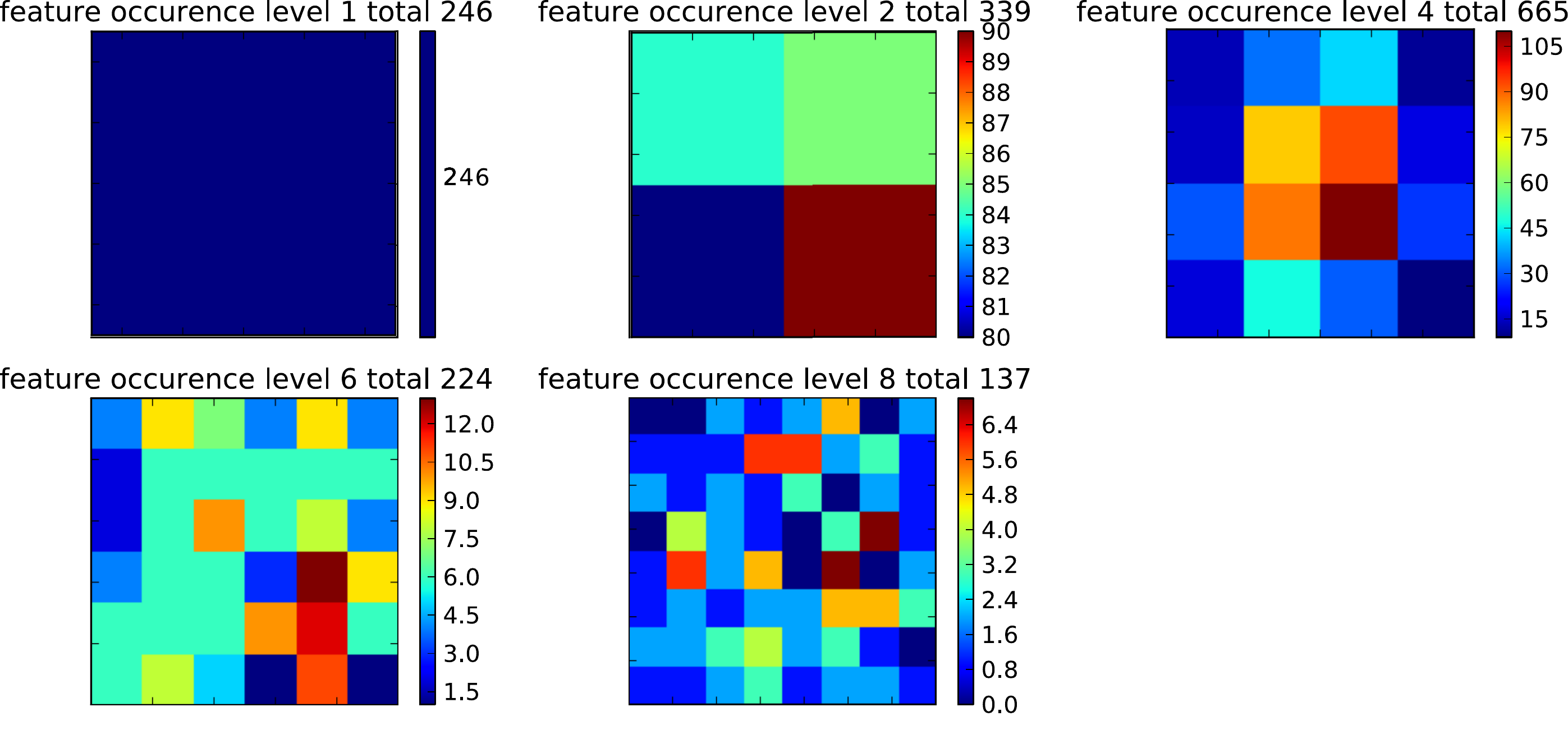}
 \end{center}
 \caption{\footnotesize Spatial histograms showing the quantity and arrangement of features chosen most frequently during training from each level of the NCC grid, for our translational RRF.}
 \label{fig:feature_occurrence_hist}
\end{figure}

The RRF training process selects some elements of the feature vector more frequently than others for estimating the transform parameters at test time.
\figref{fig:feature_occurrence_hist} shows spatial histograms for each scale of the NCC grid, indicating the frequency with which a feature computed from that NCC sub-window was chosen by the forest training process.
Interestingly, the most used level is the $4 \times 4$ grid, and there is a strong peak within that histogram for the most central 4 of the 16 possible NCC sub-windows.
While using a single NCC calculation to compare whole images is a common approach for image alignment, these graphs show that the finer grained NCC sub-windows are providing important extra information to get the right offset.
The top level of features (representing a single global NCC comparison) are chosen by the forest training process 246 times, whilst the $4 \times 4$ resolution features are chosen 665 times.
We know that this is due to these features being more informative, rather than simply more numerous, because the $8\times8$ level, which contains a larger number features than all the other levels put together, is only chosen 121 times.

\subsection{Swipe Mosaic Visualization}

A key property of Swipe Mosaic visualization is being able to grab elements and navigate spatially between temporally distant frames.    
As shown in the video, it is possible to easily maintain a sense of position while navigating within the wider scene.
Possible applications for the system include recording the damage in a car accident for later scrutiny, or examining products when shopping online.

\subsection{Suitable Image Sequences}
\label{sub:variety_of_sequences}

\begin{figure*}
 \begin{subfigure}[t]{0.98\textwidth}
  \begin{subfigure}[t]{0.19\textwidth}
   \includegraphics[width=\columnwidth]{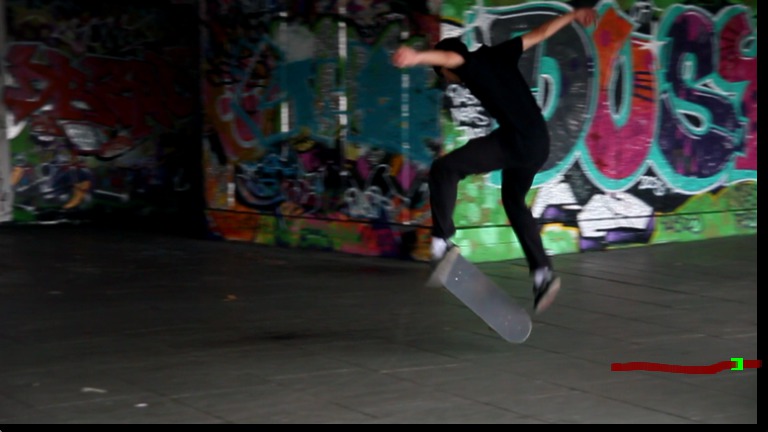}
   \caption{\footnotesize \dataset{skater}}
   \label{fig:result_skater}
  \end{subfigure}
  \begin{subfigure}[t]{0.19\textwidth}
   \includegraphics[width=\columnwidth]{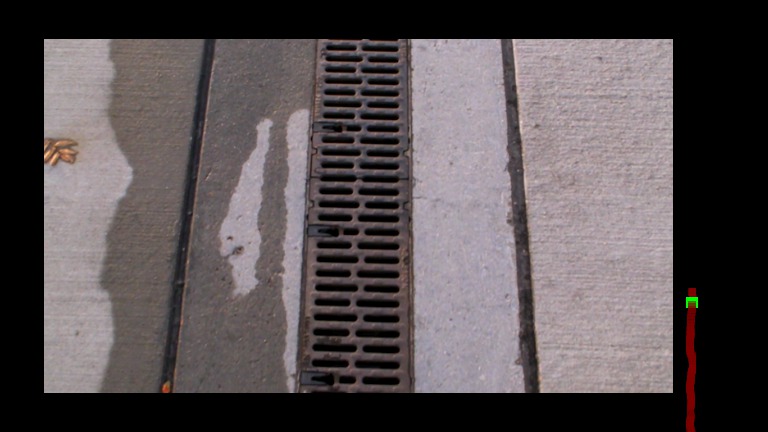}
   \caption{\footnotesize \dataset{grating}}
   \label{fig:result_grating}
  \end{subfigure}
  \begin{subfigure}[t]{0.19\textwidth}
   \includegraphics[width=\columnwidth]{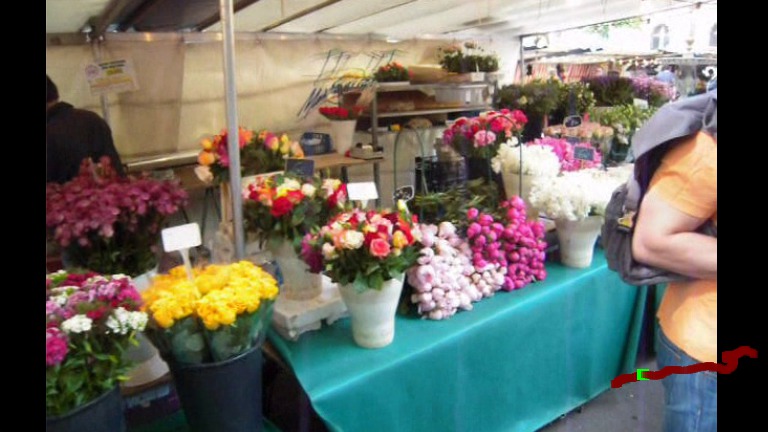}
   \caption{\footnotesize \dataset{flowers}}
   \label{fig:result_flowers}
  \end{subfigure}
  \begin{subfigure}[t]{0.19\textwidth}
   \includegraphics[width=\columnwidth]{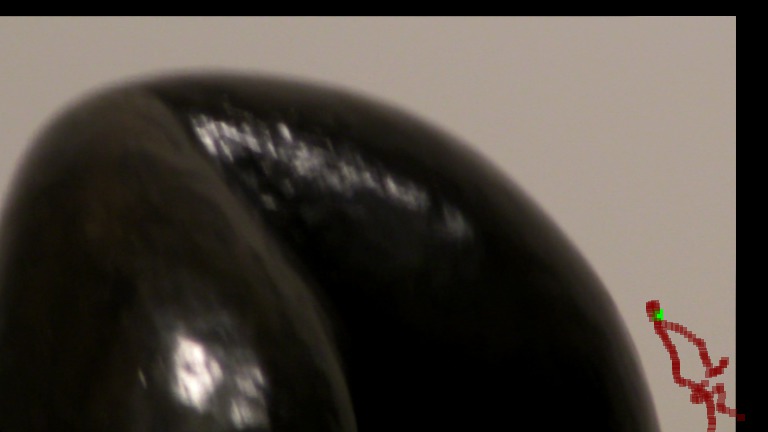}
   \caption{\footnotesize \dataset{sculpture}}
   \label{fig:result_sculpture}
  \end{subfigure}
  \begin{subfigure}[t]{0.19\textwidth}
   \includegraphics[width=\columnwidth]{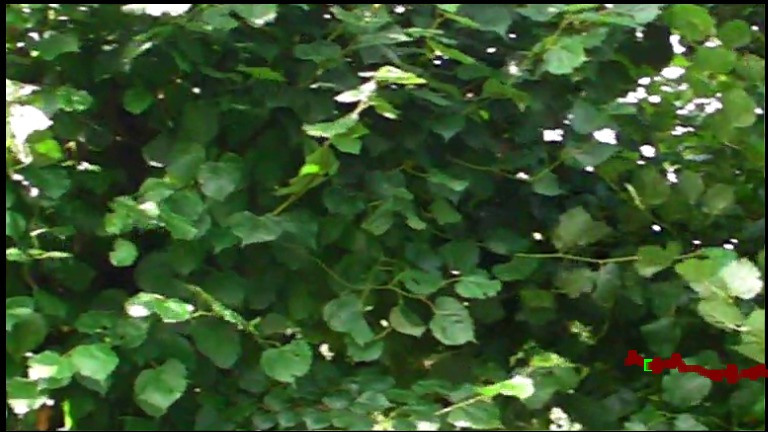}
   \caption{\footnotesize \dataset{leaves}}
   \label{fig:result_leaves}
  \end{subfigure}                            
 \end{subfigure}

 \begin{subfigure}[t]{0.98\textwidth}
  \begin{subfigure}[t]{0.19\textwidth}
   \includegraphics[width=\columnwidth]{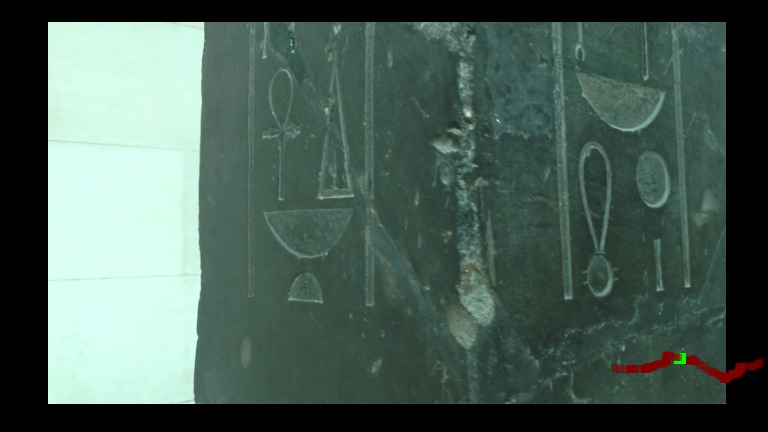}
   \caption{\footnotesize \dataset{obelisk}}
   \label{fig:result_obelisk}
  \end{subfigure}
  \begin{subfigure}[t]{0.19\textwidth}
   \includegraphics[width=\columnwidth]{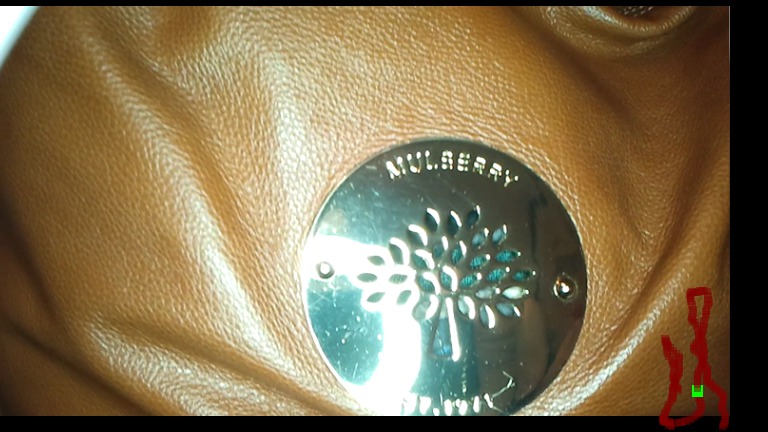}
   \caption{\footnotesize \dataset{handbag}}
   \label{fig:result_handbag}
  \end{subfigure}
  \begin{subfigure}[t]{0.19\textwidth}
   \includegraphics[width=\columnwidth]{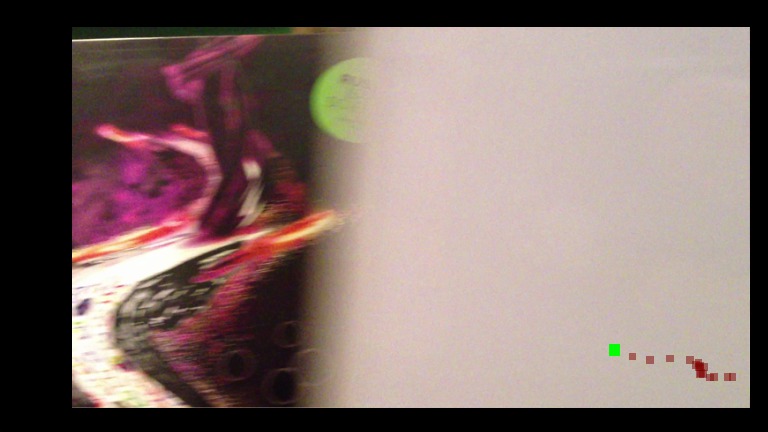}
   \caption{\footnotesize \dataset{vinyl}}
   \label{fig:result_vinyl}
  \end{subfigure}
  \begin{subfigure}[t]{0.19\textwidth}
   \includegraphics[width=\columnwidth]{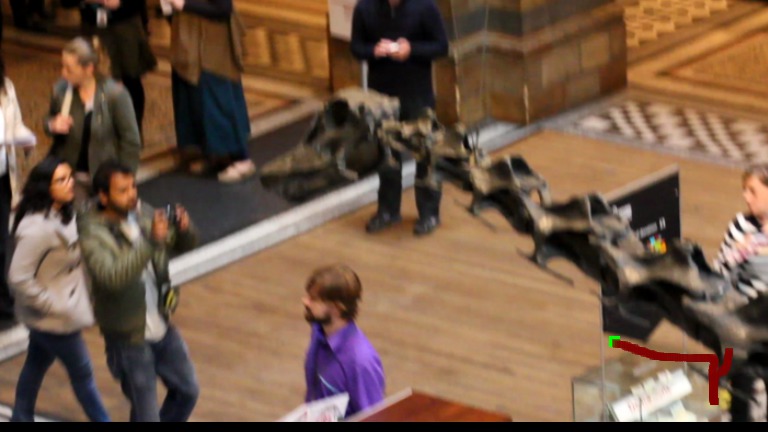}
   \caption{\footnotesize \dataset{dino}}
   \label{fig:result_dino}
  \end{subfigure}
  \begin{subfigure}[t]{0.19\textwidth}
   \includegraphics[width=\columnwidth]{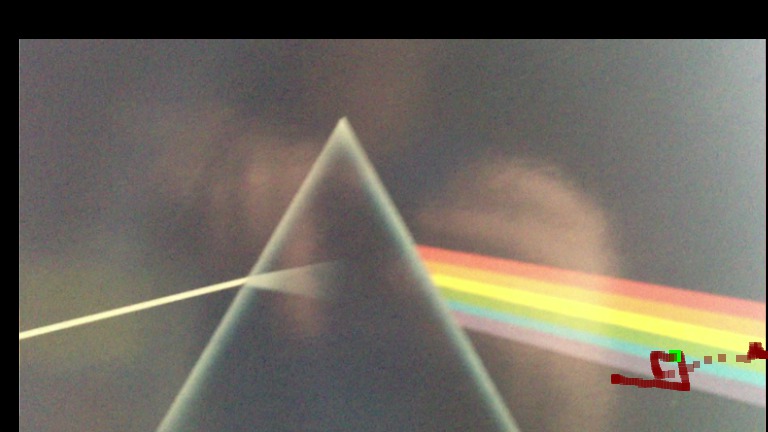}
   \caption{\footnotesize \dataset{prism}}
   \label{fig:result_prism}
  \end{subfigure}                            
 \end{subfigure}
 
 \caption{\footnotesize Screenshots of various sequences loaded in the Swipe Mosaic viewer}
 \label{fig:results}
\end{figure*}

Swipe mosaics are best observed in motion, so we present qualitative results in the video.
Our system performs best when the camera motion and scene geometry are parallel and both approximately planar (as with the training data), but we are robust to deviations from this setup.
\dataset{obelisk} shows that if the object of interest fills much of the screen, we infer a 2D version of the motion as the camera \emph{rotates around} the object.
\dataset{dino} contains people in the background moving in various directions, but the transforms computed allow navigation along the main item of interest.
The level 4 histogram in ~\figref{fig:feature_occurrence_hist} helps explain this; the RRF learns that the center of the image is usually more informative, and so treats the motion implied by central pixels as more informative than that implied by edge pixels.
View dependent effects such as the specularities in the \dataset{handbag} sequence do not lead to incorrect motion estimates.
\dataset{skater}, however, features non rigid movement in the \emph{centre} of the frame and only the \emph{outskirts} imply the (correct) sideways motion.

\begin{figure*}
 \begin{subfigure}[t]{0.19\textwidth}
  \includegraphics[width=\columnwidth]{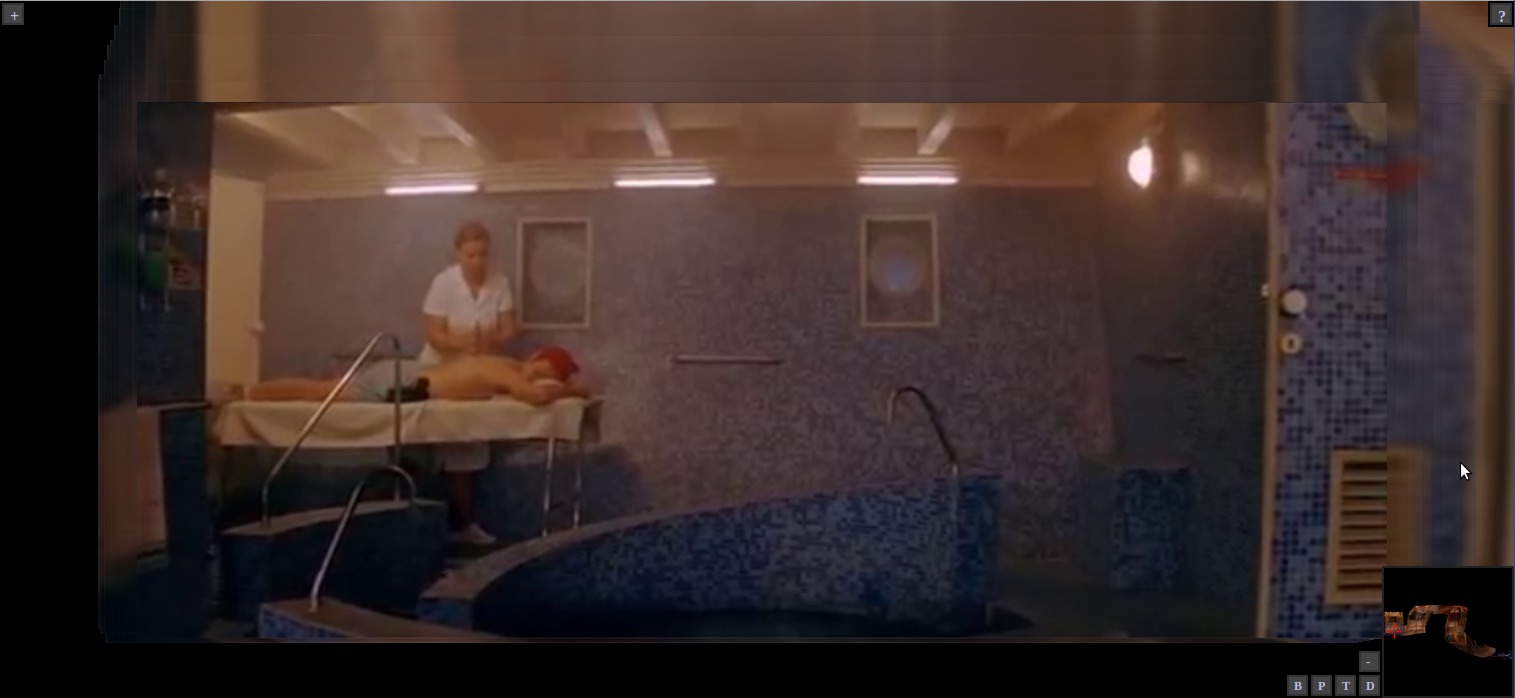}
 \end{subfigure}
 \begin{subfigure}[t]{0.19\textwidth}
  \includegraphics[width=\columnwidth]{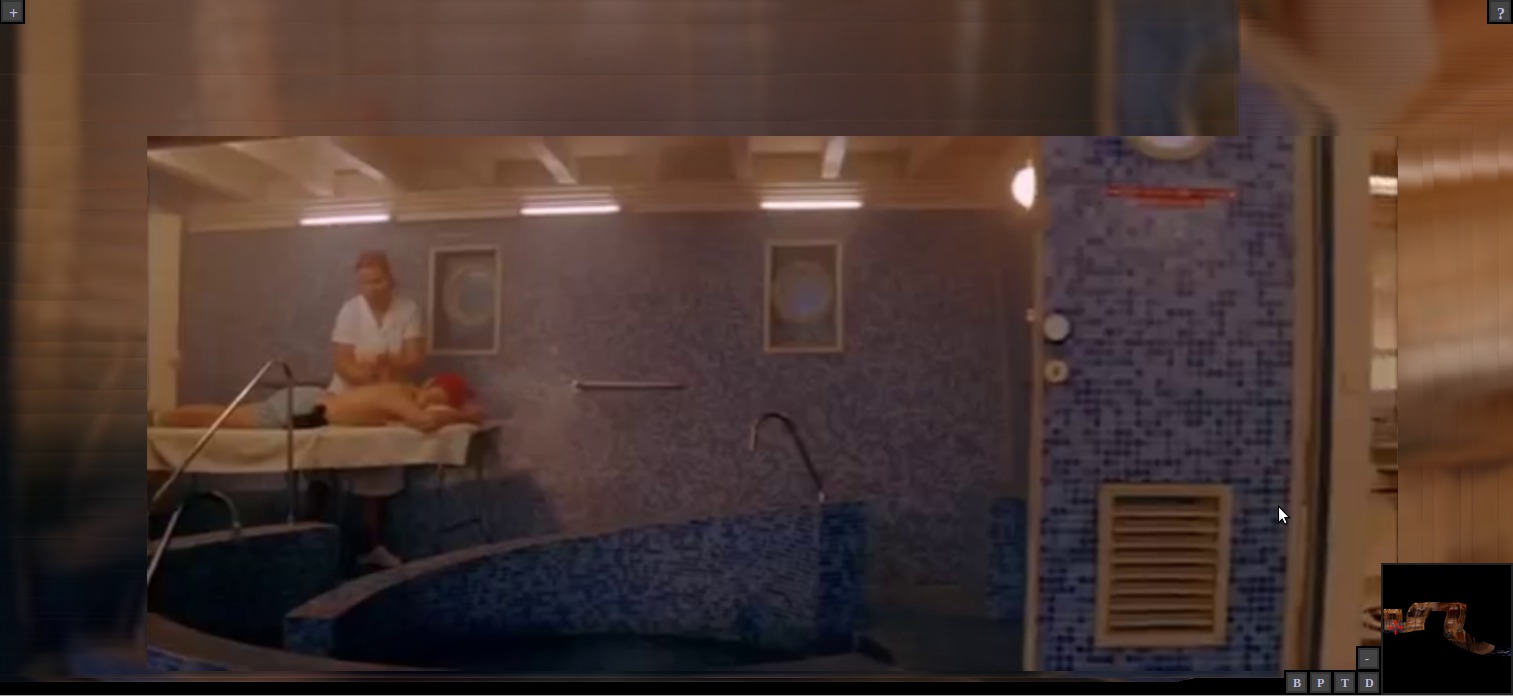}
 \end{subfigure}
 \begin{subfigure}[t]{0.19\textwidth}
  \includegraphics[width=\columnwidth]{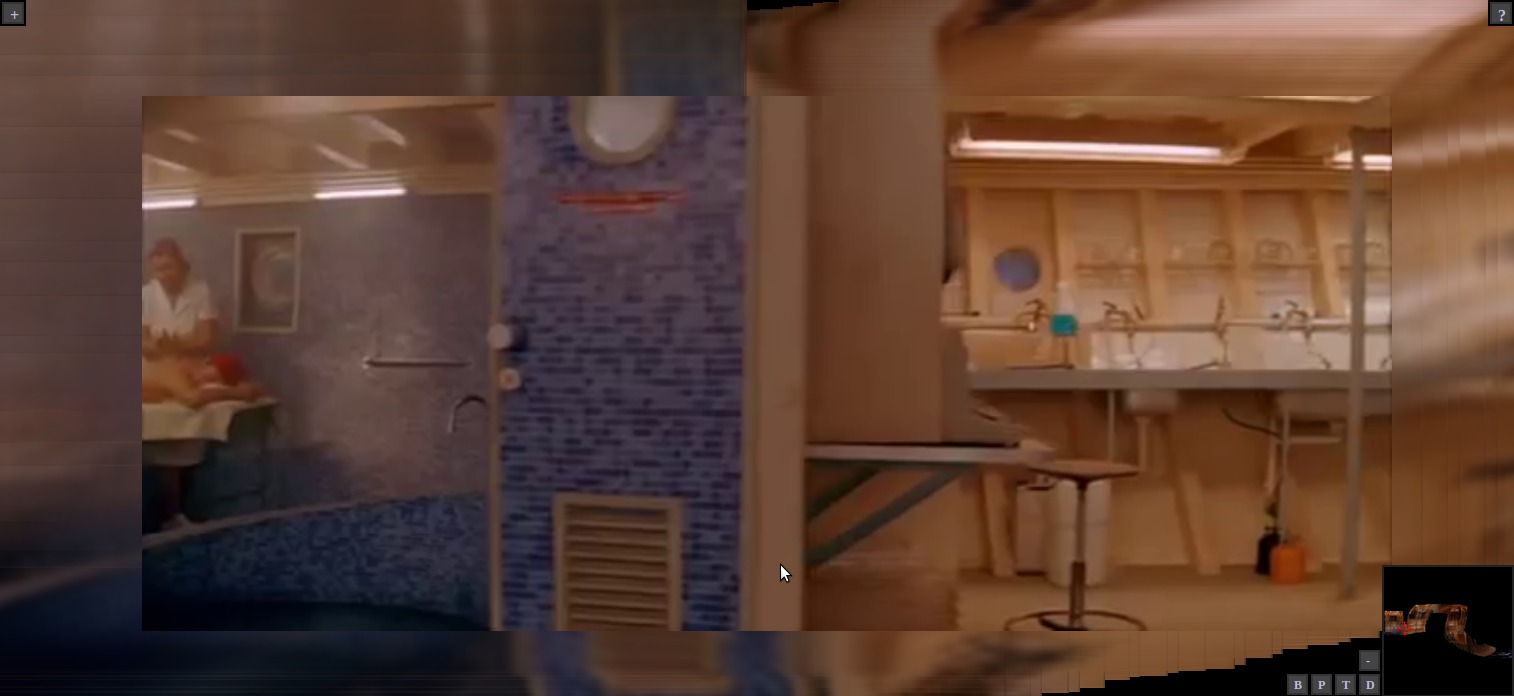}
 \end{subfigure}
 \begin{subfigure}[t]{0.19\textwidth}
  \includegraphics[width=\columnwidth]{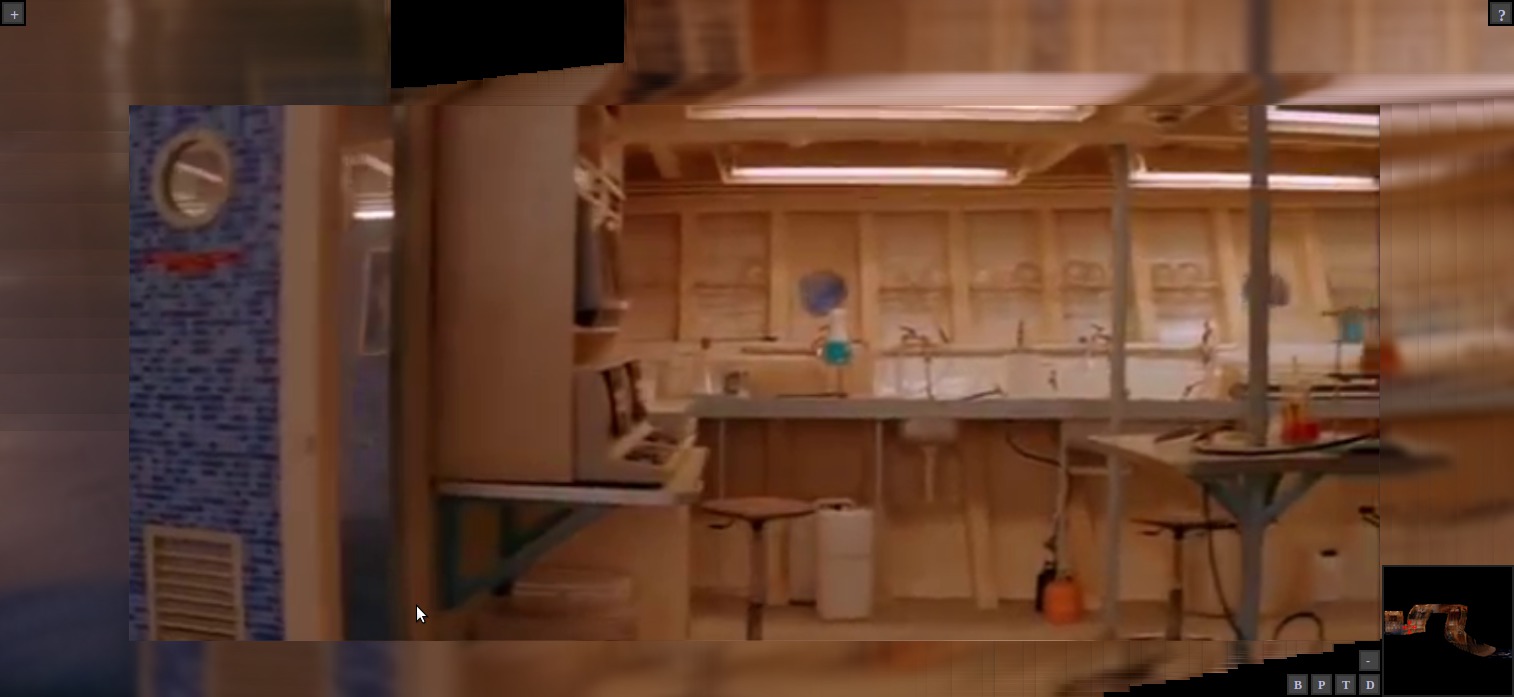}
 \end{subfigure}
 \begin{subfigure}[t]{0.19\textwidth}
  \includegraphics[width=\columnwidth]{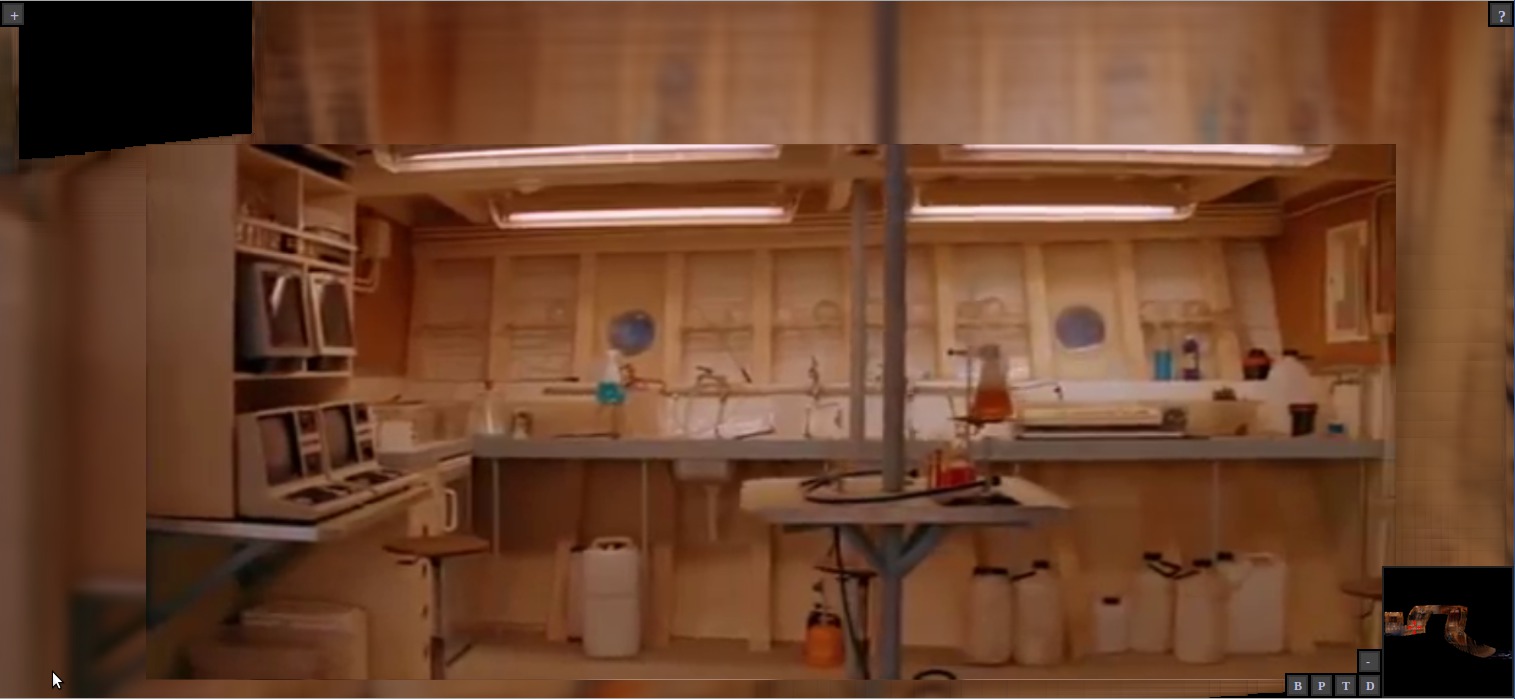}
 \end{subfigure}
 
 \caption{\footnotesize A series of screenshots taken during a single swipe, navigating \dataset{aquatic}, a scene generated from the movie ``The Life Aquatic with Steve Zissou''}
 \label{fig:aquatic_screenshots}
\end{figure*}

To demonstrate the utility of our system on existing sequences filmed by others, we processed a scene from the movie ``The Life Aquatic with Steve Zissou''.
\dataset{aquatic} features the camera panning over a cutaway version of a boat, traveling between rooms and showing different characters.
The camera trajectory roughly matches the assumptions made in our training data, but the scene contains large amounts of parallax due to depth variation, as well as dynamic characters.
We put $600$ frames into our system and built a swipe mosaic which allows intuitive navigation between seven distinct areas.
Sample frames from transitioning ``through'' a wall are shown in \figref{fig:aquatic_screenshots}.
Please see the supplemental video to view this scene in motion.

Sequences such as \dataset{handbag} (\figref{fig:result_handbag}) can be processed successfully despite containing out-of-plane camera translation and strong specularities.
Note that if a loop point featured images with differing scale, this would pose a problem for our system both in terms of the loop closure algorithm and in terms of viewer artifacts.
To test the limits of our system, we ran it on a challenging part of the \dataset{flowers} scene, a failure case from \cite{Goldstein2012videostabilization}, which the authors describe as troublesome due to the pedestrians occluding geometry and cutting feature trajectories.
The camera is moving forwards as well as sideways so that the contents of the flower stall appear to be moving roughly diagonally in image space.
Our system copes with this motion, and we can browse the scene by swiping elements on the flower stall.
This challenging video also demonstrates a failure mode of our system; obstructing bystanders in the image centre, combined with motion that differs significantly from that of our training data, makes for a very difficult scene.

\subsection{Qualitative Evaluation against Baseline Algorithms} 
\label{sub:evaluation_against_baseline_algorithm}

We evaluate the performance of our odometry regressor by comparing to simple NCC based alignment~\cite{Szeliski:2006:IAS}, VisualSfM by Wu~\etal~\shortcite{Wu:SIFTGPU:2007,Wu:MCBA:2011}, Viewfinder Alignment by Adams~\etal~\shortcite{adams2008viewfinderalignment}, Real-time image-based tracking of planes using Efficient Second-order Minimization (ESM) by Benhimane \& Malis~\cite{Benhimane:ESM:2004}, and ``microSfM'' or ``$\mu$SfM'', a new system which uses the methodology of Fundamental Matrix computation but produces a 2D translation.
While these techniques can be applied to a wide range of sequences, we confirmed known circumstances under which each of the baselines failed, and our technique succeeded.
Our baseline comparisons are summarized below; please see the Appendix and video for details.

\begin{figure}
 \begin{center}
  \includegraphics[width=0.8\columnwidth]{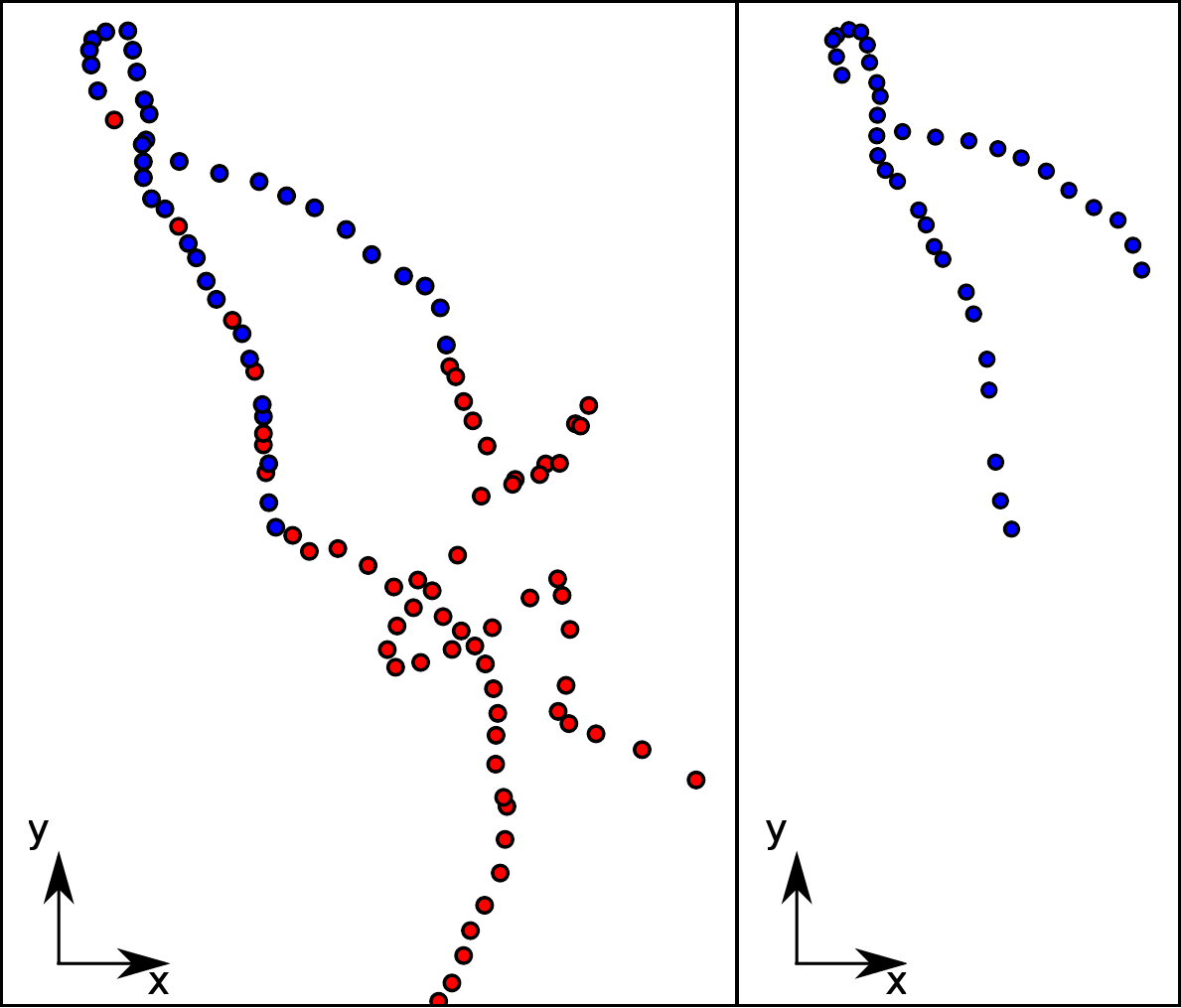}
 \end{center}
 \caption{\footnotesize $2$D camera coordinates (unitless) for \dataset{sculpture} produced by our system (left) and VisualSFM (right).
 Blue points indicate images where both systems gave an estimate of location (note the estimates differ); red points are images where only our system produced an estimate.
 NB: One outlier is not shown in the right hand image.}
 \label{fig:sculpture_ours_and_visualsfm}
\end{figure}

Our NCC based feature encoding is partly inspired by the \textbf{Direct Methods} detailed by Szeliski~\cite{Szeliski:2006:IAS}, which compute the best 2D alignment between images from the location of the peak in their combined NCC response.
This simple method often succeeds, but we found sequences where the NCC method produces incorrect results.
Szeliski~\cite{Szeliski:2006:IAS} notes \emph{``[NCC's] performance degrades for noisy low-contrast regions''}.
For example, in the \dataset{prism} sequence where horizontal movement takes place, the failure occurs on two specific frames where the resulting NCC image had a maximum implying a translation of $(0, 0)$.
The NCC response contained a second mode, corresponding to a more sensible horizontal offset, but the height of this ``correct'' peak was slightly lower than the peak representing zero motion, so it would never be chosen by the alignment algorithm.
Systems deterministically selecting the global NCC peak and ignoring other factors will fail on this scene and similar scenes.
Our RRF incorporates this peak offset information as part of the feature vector.

\textbf{VisualSfM}~\shortcite{Wu:SIFTGPU:2007,Wu:MCBA:2011} is a state of the art SfM system, which can process images either as ``ordered'' (temporally sequential frames) or ``unordered''.
It produces excellent results in general, but struggles when few or misleading feature matches are present. For example, we evaluate on two sequences: \dataset{sculpture} containing motion blur and textureless regions, and \dataset{leaves} containing moving geometry.
VisualSfM failed to return a full, correct result for either scene, whereas in both cases our system produced a full layout suitable for browsing as a Swipe Mosaic.
For \dataset{sculpture}, the best result was with ordered mode, but only 38 camera positions were returned for an input of 101 images (\figref{fig:sculpture_ours_and_visualsfm}).
Camera positions were produced for all 201 frames in \dataset{leaves} using unordered mode, but the location accuracy becomes progressively worse throughout the sequence (see Video).

\textbf{$\mu$SfM} is a system of our own creation using traditional SIFT matching and RANSAC to compute a 2DoF translational offset rather than a 7DoF fundamental matrix.
The \dataset{vinyl} sequence, containing a textureless obstruction close to the camera, causes $\mu$SfM to fail.
Descriptors computed on the obstruction display self-similarity, resulting in noisy matches.
Frames containing the obstruction were laid out far to the right of the rest of the scene, when they should be in the middle.
Our system produces a correct horizontal motion path.
A similar failure occurs in \dataset{prism}, which features a typical home video problem of the camera moving into direct sunlight, and the automatic gain control taking a few frames to adapt.
No interest points were matched between the beginning and end of the sequence, so camera locations cannot be inferred for the whole sequence.
Our system produces a zero mean, high variance estimate (effectively applying a Brownian Motion prior) whenever there is no visual evidence suggesting any particular direction of motion, allowing the resulting (complete) set of camera locations to be browsed as a Swipe Mosaic.

\begin{figure}
 \begin{subfigure}[t]{0.49\textwidth}
  \includegraphics[width=\columnwidth]{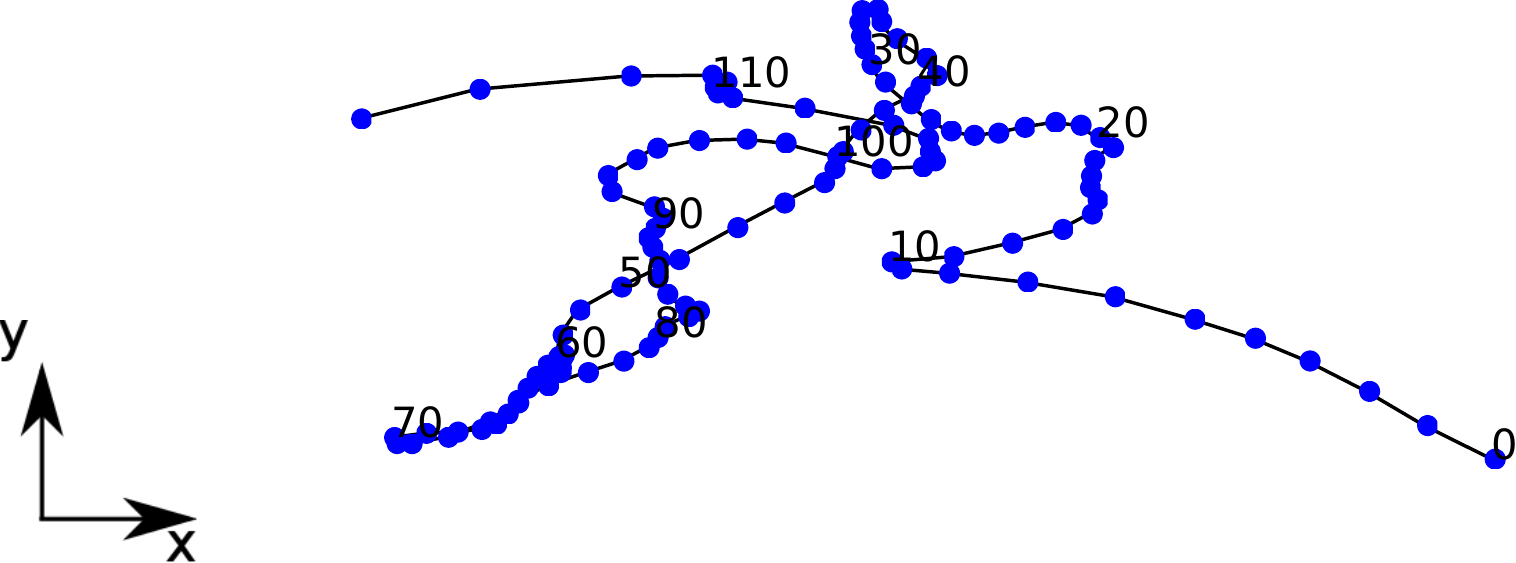}
  \caption{\footnotesize ESM}
  \label{fig:obelisk_esm}
 \end{subfigure}
 \begin{subfigure}[t]{0.49\textwidth}
  \includegraphics[width=\columnwidth]{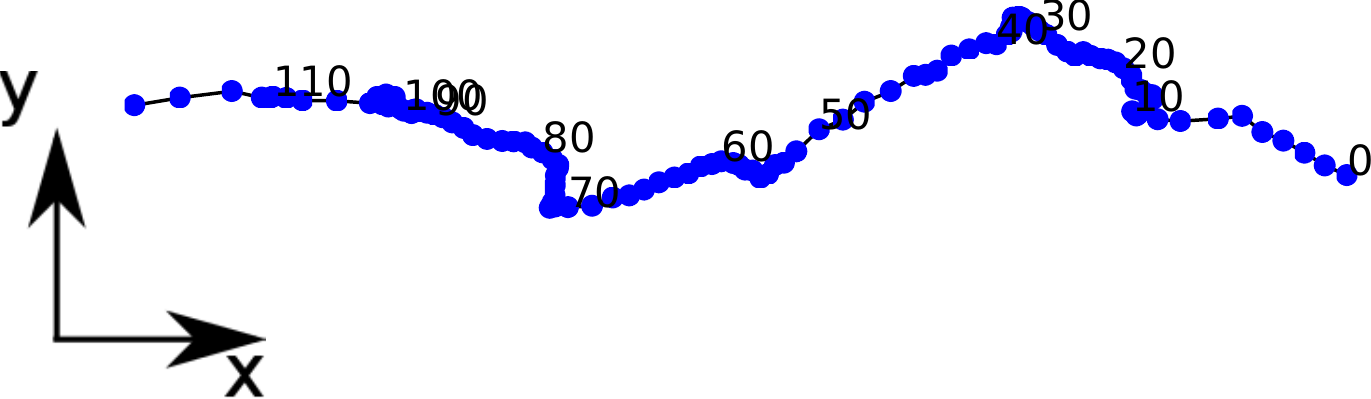}
  \caption{\footnotesize Ours}
  \label{fig:obelisk_pcm}
 \end{subfigure}
 \caption{\footnotesize 2D camera coordinates produced for \dataset{obelisk} using ESM (\figref{fig:obelisk_esm}) and our method (\figref{fig:obelisk_pcm}).
 The true camera motion is approximately constant horizontal translation, coupled with rotation to remain pointed towards the object of interest (see supplemental video).}
 \label{fig:obelisk_comparison}
\end{figure}

\textbf{Real-time image-based tracking of planes using Efficient Second-order Minimization} (ESM)~\cite{Benhimane:ESM:2004} is a direct method which explicitly models the scene as a plane, and searches for a parameterised transform which minimises the sum of squared differences between two images.
The transform can be parameterised as anything from a full homography (8DoF) to a translational transform (2DoF), and the parameters are solved for using an efficient method which achieves Newton method like convergence rates, without having to compute the Hessian.
We used the implementation of ESM available in Ed Rosten's LibCVD project, using 2 DoF to produce a translation between each image pair, before using our layout algorithm.
ESM produces a good Swipe Mosaic result for some of our test sequences, but the explicit parameterisation of the scene as a plane leads to problems when faced with non-planar camera paths or distorting objects.
ESM performed very badly on the \dataset{obelisk} scene, laying out frames which should be very far from each other in roughly the same place.
Our system produced an intuitively navigable Swipe Mosaic.
The camera paths produced by ESM and our method are shown in \figref{fig:obelisk_comparison}.
Note the broadly horizontal linear path produced by our method in contrast to the ESM path which continually crosses itself.
This is due to our method's more gradual degradation as scenes deviate from planar, allowing us to cope with strong perspective deformations.
See the supplemental video for a comparison of the browsing experience of these two solutions.

\textbf{Viewfinder Alignment} (VfA)~\shortcite{adams2008viewfinderalignment} computes constrained transforms between temporally close video frames.
A ``digest'' containing edge information in multiple orientations, and the locations of the strongest detected corners is computed for each frame.
Two digests are matched by aligning the histograms to give a putative 2D shift, and aligning the detected corner locations to evaluate how likely this shift is to be correct.
VfA works well given sufficient strong edges and recognisable corners, but if either are absent it will produce an incorrect transform, or no transform at all.
The \dataset{grating} sequence contains similar strong edge information in all frames, leading VfA to erroneously believe geographically distant frames were close together.
Apart from the edges, sufficient visual cues are present in the image so that both our system and $\mu$SfM inferred a correct result.
Because only the location, and no visual descriptor, is stored for each corner, VfA is prone to linking disparate frames with similar edges.
Low texture regions in \dataset{prism} or \dataset{vinyl} cause no corners to be detected, meaning VfA cannot compute a result.
Further results are included in the Appendix.

\subsection{Quantitative Evaluation Against NCC}
\label{sub:quantitative_evaluation_against_ncc}

\begin{figure*}
 \includegraphics[width=\textwidth]{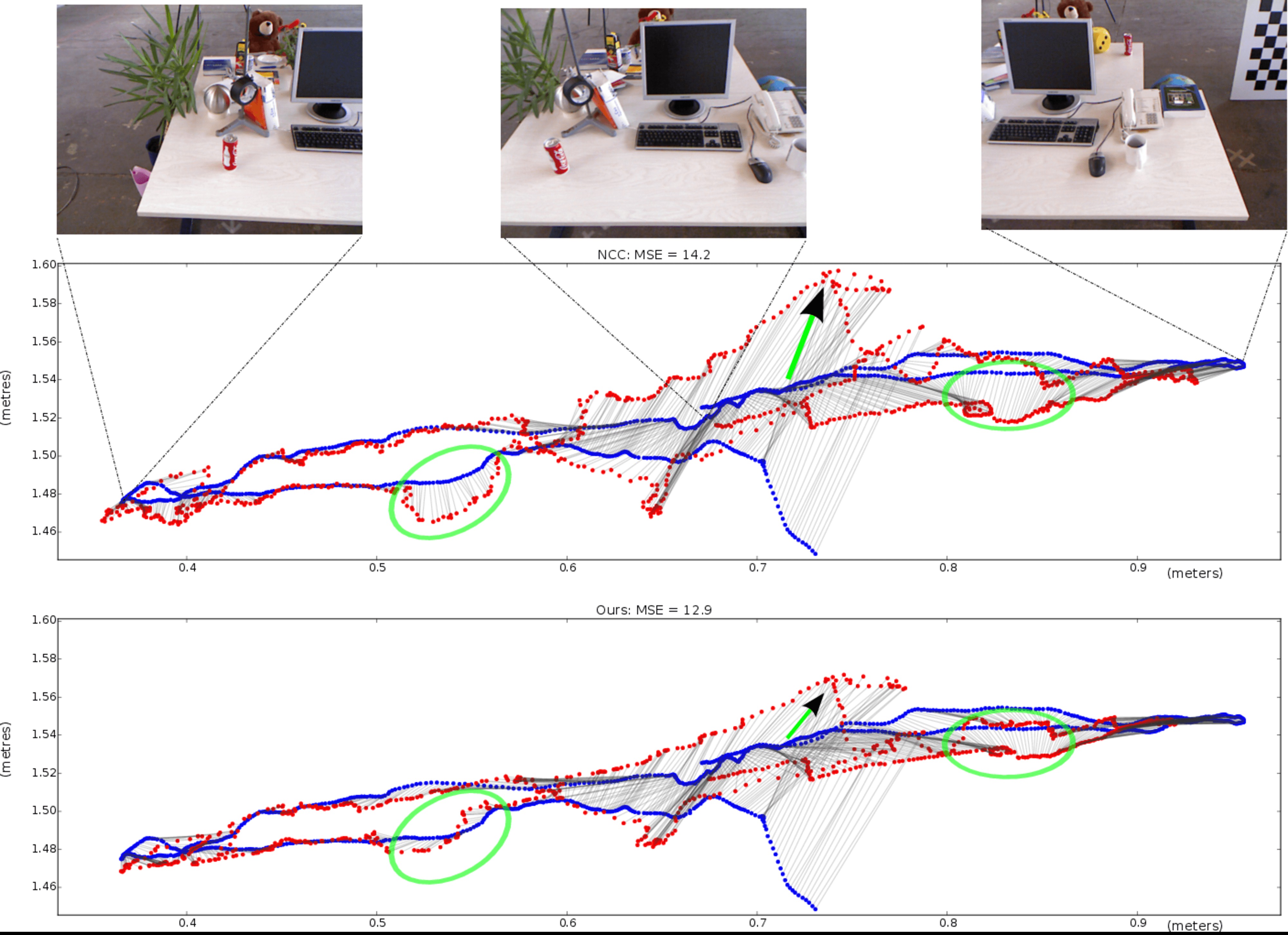}
 \caption{\footnotesize Alignment result for NCC (top) and our system (bottom) for the \dataset{freiburg2} sequence \cite{Sturm2012TUM}.
 Each graph shows the camera locations from the prospective solution (Red) after being aligned using Procrustes to the ground truth locations computed with a ``best fit'' plane (Blue).
 Corresponding camera locations are joined by grey lines.
 Note the highlighted regions (green ellipses / arrows) in which our system provides a solution substantially closer to ground truth.
 Navigating these areas of the NCC solution as a Swipe Mosaic would require users to follow a more distorted trajectory than would seem natural.
 Corresponding frames are shown above to give an idea of the scene makeup.
 Onscreen viewing recommended.}
 \label{fig:ncc_eval_alignment}                          
\end{figure*}

\begin{figure}
 \includegraphics[width=\columnwidth]{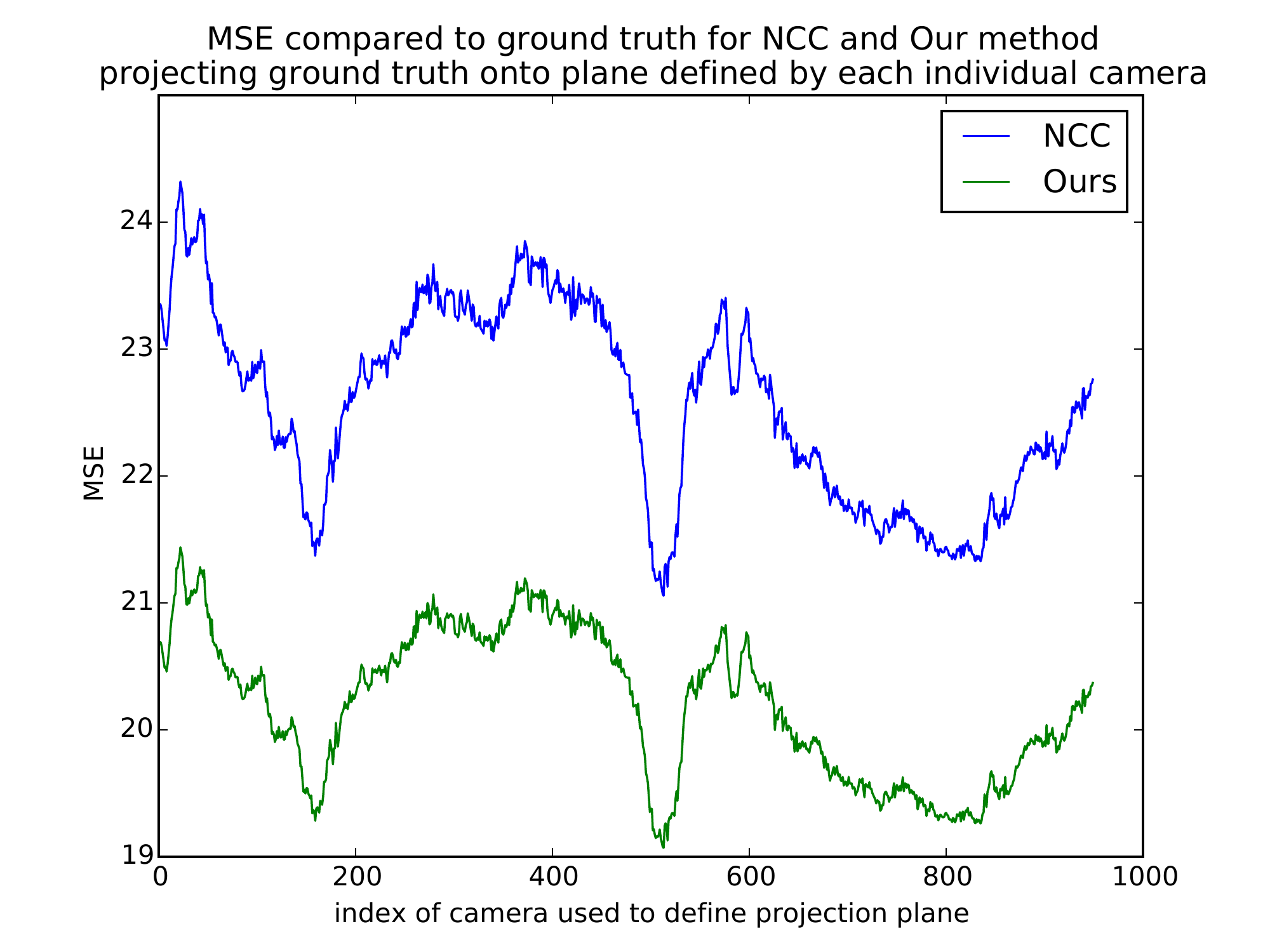}
 \caption{\footnotesize Comparison of Mean Squared Errors between our method and NCC for the whole range of possible individual camera projection planes, for \dataset{freiburg2}.
 Note the self similarity of the lines.
 As we change which ground truth camera defines the projection plane, both solutions (being so similar) have coinciding increases or decreases in performance.}
 \label{fig:ncc_eval_global_sweep}
\end{figure}

While graceful degradation is easy to illustrate qualitatively, it is reasonable to check if regression using our NCC-based feature vectors is actually different than just using NCC directly, at least for best-case in-plane motion and static scenes with negligible motion blur.
To quantitatively evaluate the odometry of our system, we compare on the \dataset{freiburg2} sequence from the TUM~\cite{Sturm2012TUM} dataset.
This dataset consists of Kinect video sequences alongisde 6D ground truth camera positions, generated using $100$Hz active motion capture.
Unlike the rest of the dataset, the camera path in \dataset{freiburg2}'s first 950 frames contains almost exclusively vertical and horizontal translation. 
This is the kind of motion that both NCC and our system should be able to handle.
The appearance of the indoor office scene is unremarkable in terms of texture or dynamic elements.

Both NCC and our system were used to generate $2$D camera locations for the sequence.
Every pair within a 4 frame temporal window was used to generate relative offset predictions.
For both systems, the offsets were fed to our least squares layout algorithm.
To compare the $6$D ground truth poses with each $2$D solution, three steps had to be carried out.
First, a ground truth pose must be established for each RGB frame, as the dataset only provides camera locations from motion capture, and the Kinect and motion capture systems were running unsynchronised at different frequencies.
Secondly, some projection of each $6$D camera pose onto $2$D must be established, and finally the $2$D solutions must be scaled and aligned to assess the correctness of the locations.

To establish a $6$D ground truth location for each RGB frame, we linearly interpolated between the closest two motion capture positions using the globally synched timestamps (available for both motion capture and Kinect readings).
The rotation was encoded as a quaternion during this process, ensuring linear interpolation is a reasonable approach.
The result of this operation is a $6$D camera pose corresponding exactly to each RGB frame.
The $6$D to $2$D projection step is described below, and the final alignment step is carried out using the Procrustes algorithm~\cite{PrinceCVMLI2012}.

\begin{figure}
 \def\svgwidth{\linewidth}
 \input{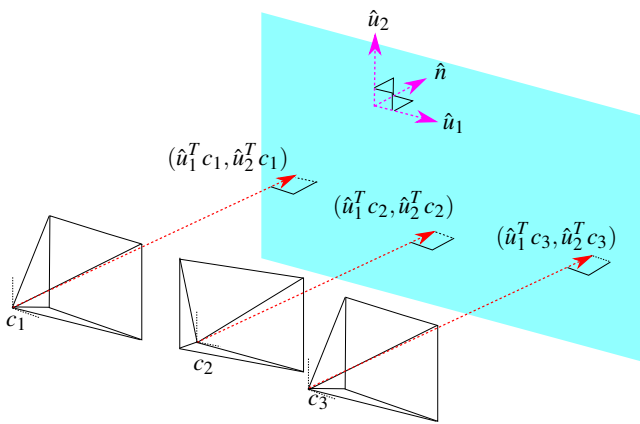}
 \caption{\footnotesize Diagram showing how 3D camera positions $\vect{c}_1, \vect{c}_2, \vect{c}_3$ are projected (red dashed line) onto a 2D coordinate space defined by a plane (blue).
 For the ``best fit'' case, the plane normal $\hat{\vect{n}}$ is generated by averaging the forward vectors of all the cameras.
 The basis vectors $\hat{\vect{u}}_1$ and $\hat{\vect{u}}_2$ lie within the plane and are mutually orthogonal.
 When fitting to the plane defined by an individual camera, $\hat{\vect{u}}_1$ and $\hat{\vect{u}}_2$ are chosen to be parallel to the ``right'' and ``up'' vectors of individual cameras (black dotted lines).}
 \label{fig:projection_onto_plane}
\end{figure}

Given a $3$D plane represented as a unit normal $\hat{\vect{n}}$ and a point on the plane $\vect{p}$, we define two unit vectors $\hat{\vect{u}}_1$ and $\hat{\vect{u}}_2$ such that $\hat{\vect{u}}_1^T\hat{\vect{u}}_2 = \hat{\vect{u}}_1^T\hat{\vect{n}} = \hat{\vect{u}}_2^T\hat{\vect{n}} = 0$.
The exact orientation of these vectors is not important as the subsequent alignment includes a rotation step.
Each camera location $\vect{c}_i$ is projected onto a $2$D location on the plane $(\hat{\vect{u}}_1^T\vect{c}_i, \hat{\vect{u}}_2^T\vect{c}_i)$ (see \figref{fig:projection_onto_plane}).
To define the plane orientation, we tried both a ``best fit'' to all the camera locations, and also tried using the ``up'' and ``right'' vectors of each individual camera in turn.
The ``best fit'' plane was defined by setting $\hat{\vect{n}}$ to the average forward direction of each camera.
The maximum angular difference between the computed normal and the forward vector of any of the cameras is $9.46$ degrees, confirming that the scene contains only minimal rotation and is therefore a good candidate sequence for this comparison.

On the best fitted plane, NCC gave a final Mean Squared Error of $14.2$ \si{cm^2} against our system's $12.9$ \si{cm^2}, an improvement of $9.4\%$.
The alignments resulting from both systems are shown in \figref{fig:ncc_eval_alignment}.
It is interesting to note that both algorithms make similar mistakes, owing perhaps to our system being built on top of NCC based features, but our system produces errors of smaller magnitude, especially towards the center of the horizontal axis, because it incorporates more information than just the top level, entire-image NCC comparison.
As well as comparing with this fitted plane, we tried aligning to each of the planes defined by one individual camera's orientation, by projecting all other cameras onto the local ``right'' and ``up'' vectors.
The results of this are shown in \figref{fig:ncc_eval_global_sweep}.
Unsurprisingly, whichever one of the $950$ cameras we choose, we see an improvement with our system, as shown by the green line always being underneath the blue line. 
This evaluation shows that even for a texture-rich real world sequence, not captured with Swipe Mosaics in mind, our system produces a measurable improvement over the alignment produced by NCC.

Our evaluations show the robustness of our system to visual phenomena found difficult by other systems.
In the presence of dynamic objects, lack of texture, or repeated structure, we are able to compute $2$D locations which enable browsing of the scene, degrading gracefully in the presence of inconclusive visual information.
Many of our sequences were captured by users unfamiliar with the workings of the system, and our success here demonstrates our robustness to input data straying outside the assumptions of the training data.
All results were generated by a single trained translational RRF with $10$ trees and maximum depth $12$.
At each node, $2000$ possible feature splits were considered.
The one-time training took 1h 43m on a Core 2 Duo 2.8 GHz, using both CPU cores.
All experiments were carried out with the same synthetic training set built from $8800$ image pairs.
The rotational RRF was trained with $20$ trees of depth $12$, considering $100$ feature splits per node.
Only $400$ image pairs were used in the training set, as inferring pure rotation is a strictly easier problem than translation, because what appears in the images is not scene geometry dependent.
All images were $768 \times 432$ or $640 \times 480$.
A $2.5$ GHz single core Xeon, computed $121$ NCC matches and the $3599$D features for two images in 15 seconds.
Most useful datasets contain thousands of image pairs, so we used a cluster to process datasets quickly.

\section{Conclusion}
\label{sec:conclusion}

Our results show that Swipe Mosaics can be used to intuitively navigate video sequences containing a range of camera motions and visual content, including those that failed or trouble existing standard baselines.
As seen in the video, even passively observing someone else's Swipe Mosaic interaction provides a good sense of a scene's layout.
Traditional panoramic image mosaics undoubtedly have a cleaner overall appearance than our Picasso-view.
Yet the payoff of browsing video frames through our interface is enormous: footage exhibiting parallax and other view- and lighting-dependent effects can be visualized without the extra user effort needed for most multi-perspective renderers (\eg \cite{rav2008unwrap}), because the pixels need not join up.
Training our RRFs on synthetic data has led to a visual odometry system that achieves our goals.
It manages to estimate visually-acceptable translations and rotation both in textured scenes, where existing methods also work well, but also in much more difficult scenes, where other methods become brittle or fail.
It is certainly possible that our model may learn still better correlations between appearance and pose if, for example, 3 or more frames were examined together, potentially allowing motion models to be incorporated.
Quite significantly from the perspective of potential users, our adapted regressor-layout pairing takes account of ambiguities when computing distributions over possible camera-motions.
This means that our visualization prototype fails more gracefully than existing systems that are designed for somewhat idealized conditions.
In an indirect way, our RRF is looking at thousands of examples to learn its own version of the user-sought visual continuity: which parts of an image pair are correlated with each other, and how?
While still challenging to dissect, the visualization of our RRF in \figref{fig:feature_occurrence_hist} shows that NCC comparisons from particular parts of the image, at particular scales, were learned to be the most informative for this task.

\subsection{Limitations and Future Work}
Some kinds of ``unexpected'' motion (motions not featured in the training set) are handled well by our system, but others are not.
When most scene geometry is moving in a similar manner, we are able to produce sensible Swipe Mosaics despite the presence of, for example, forward motion (which the RRF has not been trained on).
However, as shown by the \dataset{flowers} example, occluders dominating the image center can cause failures.
Our model has learned to rely on the center of the image somewhat
more than other areas, so it is likely that an enhanced feature vector and training set would be required to cope with this problem.
Object-recognition could also be incorporated, \eg to recognize and ignore pedestrians.
Our NCC based feature vector performs well, but other features could replace
or augment it.
VfA's edge ``digest'' is appealing in this regard as it is quick to compute and could possibly be extended to describe a distribution over different alignments.
Supporting more camera degrees of freedom in one RRF is desirable, but likely to require much more training data.
Continuing to target individual RRFs at only $1$ or $2$ degrees
of freedom each, as in this work, seems a promising approach.
A valuable improvement to the interface would be to give live feedback at capture time regarding what parts of the scene require more detailed recording.
The DTAM-based feedback in \cite{davis2012unstructured} may not be possible if scenes lack reliable interest points, but we could provide the RRF the inertial and gyroscope readings that are available in many smartphones.
For now, the system is device agnostic, making it easy to create Swipe Mosaics.
Swipe Mosaics will hopefully encourage content-creators to document and share the details of the world around them.

\bibliographystyle{eg-alpha-doi}
\bibliography{swipe_mosaics}

\newcommand{\etalchar}[1]{$^{#1}$}
\begin{thebibliography}{\uppercase{MAHPB12}}

\bibitem[AAC{\etalchar{*}}06]{agarwala2006photographing}
\textsc{Agarwala A., Agrawala M., Cohen M., Salesin D., Szeliski R.}:
\newblock Photographing long scenes with multi-viewpoint panoramas.
\newblock In \emph{TOG} (2006).

\bibitem[AGP08]{adams2008viewfinderalignment}
\textsc{Adams A., Gelfand N., Pulli K.}:
\newblock Viewfinder alignment.
\newblock \emph{Eurographics} (2008).

\bibitem[ATP{\etalchar{*}}10]{Adams:2010:Frankencam}
\textsc{Adams A., Talvala E.-V., Park S.~H., Jacobs D.~E., Ajdin B., Gelfand
  N., Dolson J., Vaquero D., Baek J., Tico M., Lensch H. P.~A., Matusik W.,
  Pulli K., Horowitz M., Levoy M.}:
\newblock The frankencamera: an experimental platform for computational
  photography.
\newblock \emph{TOG} (2010).

\bibitem[AZP{\etalchar{*}}05]{agarwala2005panoramic}
\textsc{Agarwala A., Zheng K., Pal C., Agrawala M., Cohen M., Curless B.,
  Salesin D., Szeliski R.}:
\newblock Panoramic video textures.
\newblock In \emph{TOG} (2005).

\bibitem[BL03]{Brown03iccv}
\textsc{Brown M., Lowe D.~G.}:
\newblock Recognising panoramas.
\newblock \emph{ICCV} (2003).

\bibitem[BM04]{Benhimane:ESM:2004}
\textsc{Benhimane S., Malis E.}:
\newblock Real-time image-based tracking of planes using efficient second-order
  minimization.
\newblock In \emph{Intelligent Robots and Systems, 2004.(IROS 2004).
  Proceedings. 2004 IEEE/RSJ International Conference on} (2004), vol.~1, IEEE,
  pp.~943--948.

\bibitem[Bre01]{Bleiman:RandomForests2001}
\textsc{Breiman L.}:
\newblock Random forests.
\newblock \emph{Machine Learning} (2001).

\bibitem[Cam04]{Campbell04techniquesfor}
\textsc{Campbell J.}:
\newblock Techniques for evaluating optical flow for visual odometry in extreme
  terrain.
\newblock \emph{IROS} (2004).

\bibitem[COSH11]{Crandall:DCOforSfM:2011}
\textsc{Crandall D., Owens A., Snavely N., Huttenlocher D.}:
\newblock Discrete-continuous optimization for large-scale structure from
  motion.
\newblock In \emph{{CVPR}} (2011).

\bibitem[CSK11]{criminisi2011decision}
\textsc{Criminisi A., Shotton J., Konukoglu E.}:
\newblock Decision forests: A unified framework for classification, regression,
  density estimation, manifold learning and semi-supervised learning.
\newblock \emph{Found. Trends. Comput. Graph. Vis.} (2011).

\bibitem[CW93]{Chen:1993}
\textsc{Chen S.~E., Williams L.}:
\newblock View interpolation for image synthesis.
\newblock \emph{SIGGRAPH} (1993).

\bibitem[Dav98]{Davis:1998:MSM}
\textsc{Davis J.}:
\newblock Mosaics of scenes with moving objects.
\newblock \emph{CVPR} (1998).

\bibitem[DLD12]{davis2012unstructured}
\textsc{Davis A., Levoy M., Durand F.}:
\newblock Unstructured light fields.
\newblock In \emph{Computer Graphics Forum} (2012), vol.~31.

\bibitem[DRB{\etalchar{*}}08]{videomanip-CHI08}
\textsc{Dragicevic P., Ramos G., Bibliowitcz J., Nowrouzezahrai D.,
  Balakrishnan R., Singh K.}:
\newblock Video browsing by direct manipulation.
\newblock In \emph{CHI} (April 2008).

\bibitem[DRMS07]{davison2007monoslam}
\textsc{Davison A., Reid I., Molton N., Stasse O.}:
\newblock Monoslam: Real-time single camera slam.
\newblock \emph{PAMI} (2007).

\bibitem[FB81]{RANSAC:Fischler:1981}
\textsc{Fischler M.~A., Bolles R.~C.}:
\newblock Random sample consensus: a paradigm for model fitting with
  applications to image analysis and automated cartography.
\newblock \emph{Commun. ACM} (1981).

\bibitem[GAF{\etalchar{*}}10]{Goesele:2010}
\textsc{Goesele M., Ackermann J., Fuhrmann S., Haubold C., Klowsky R.}:
\newblock Ambient point clouds for view interpolation.
\newblock \emph{TOG} (2010).

\bibitem[GF12]{Goldstein2012videostabilization}
\textsc{Goldstein A., Fattal R.}:
\newblock Video stabilization using epipolar geometry.
\newblock \emph{ACM Trans. Graph.} (2012).

\bibitem[GGC{\etalchar{*}}08]{GoldmanGCSS08}
\textsc{Goldman D.~B., Gonterman C., Curless B., Salesin D., Seitz S.~M.}:
\newblock Video object annotation, navigation, and composition.
\newblock In \emph{UIST} (2008).

\bibitem[GGSC96]{gortler1996lumigraph}
\textsc{Gortler S., Grzeszczuk R., Szeliski R., Cohen M.}:
\newblock The lumigraph.
\newblock \emph{{SIGGRAPH}} (1996).

\bibitem[GS12]{GargS12}
\textsc{Garg R., Seitz S.~M.}:
\newblock Dynamic mosaics.
\newblock \emph{3DIMPVT} (2012).

\bibitem[HKM09]{Holmes:UKF:2009}
\textsc{Holmes S.~A., Klein G., Murray D.~W.}:
\newblock An o (n$^2$) square root unscented kalman filter for visual
  simultaneous localization and mapping.
\newblock \emph{Pattern Analysis and Machine Intelligence, IEEE Transactions on
  31}, 7 (2009), 1251--1263.

\bibitem[Ho95]{Ho:Random:1995}
\textsc{Ho T.~K.}:
\newblock Random decision forests.
\newblock In \emph{Document aAnalysis and REcognition, 1995. Proceedings of the
  Third International Conference on} (1995).

\bibitem[HZ06]{HartleyZisserman}
\textsc{Hartley A., Zisserman A.}:
\newblock \emph{Multiple View Geometry in Computer Vision (2. ed.)}.
\newblock CUP, 2006.

\bibitem[IAH95]{Irani:1995:MBR}
\textsc{Irani M., Anandan P., Hsu S.}:
\newblock Mosaic based representations of video sequences and their
  applications.
\newblock In \emph{ICCV} (1995).

\bibitem[KCSC10]{Kopf2010}
\textsc{Kopf J., Chen B., Szeliski R., Cohen M.}:
\newblock Street slide: Browsing street level imagery.
\newblock \emph{SIGGRAPH} (2010).

\bibitem[KD04]{klein04tightly}
\textsc{Klein G., Drummond T.}:
\newblock Tightly integrated sensor fusion for robust visual tracking.
\newblock \emph{Image and Vision Computing} (2004).

\bibitem[KHS10]{kalogerakis2010learning}
\textsc{Kalogerakis E., Hertzmann A., Singh K.}:
\newblock Learning 3d mesh segmentation and labeling.
\newblock \emph{TOG} (2010).

\bibitem[KM07]{klein07parallel}
\textsc{Klein G., Murray D.}:
\newblock Parallel tracking and mapping for small {AR} workspaces.
\newblock In \emph{{ISMAR}} (2007).

\bibitem[KUDC07]{KopfGigaPixel07}
\textsc{Kopf J., Uyttendaele M., Deussen O., Cohen M.~F.}:
\newblock Capturing and viewing gigapixel images.
\newblock \emph{SIGGRAPH} (2007).

\bibitem[KWLB08]{karrer2008a}
\textsc{Karrer T., Weiss M., Lee E., Borchers J.}:
\newblock Dragon: A direct manipulation interface for frame-accurate in-scene
  video navigation.
\newblock In \emph{CHI} (2008).

\bibitem[LGW{\etalchar{*}}11]{Liu:2011:SVS}
\textsc{Liu F., Gleicher M., Wang J., Jin H., Agarwala A.}:
\newblock Subspace video stabilization.
\newblock \emph{TOG} (2011).

\bibitem[LH96]{levoy1996light}
\textsc{Levoy M., Hanrahan P.}:
\newblock Light field rendering.
\newblock \emph{{SIGGRAPH}} (1996).

\bibitem[Low04]{Lowe04:ijcv}
\textsc{Lowe D.~G.}:
\newblock Distinctive image features from scale-invariant keypoints.
\newblock \emph{IJCV} (2004).

\bibitem[MAHPB12]{MacAodhaPAMI2012}
\textsc{Mac~Aodha O., Humayun A., Pollefeys M., Brostow G.~J.}:
\newblock Learning a confidence measure for optical flow.
\newblock \emph{PAMI} (2012).

\bibitem[MWJ99]{murphy1999loopy}
\textsc{Murphy K., Weiss Y., Jordan M.}:
\newblock Loopy belief propagation for approximate inference: An empirical
  study.
\newblock In \emph{Proceedings of the Fifteenth conference on Uncertainty in
  artificial intelligence} (1999), pp.~467--475.

\bibitem[NNL13]{Nguyen2013direct}
\textsc{Nguyen C., Niu Y., Liu F.}:
\newblock Direct manipulation video navigation in 3d.
\newblock In \emph{CHI} (2013).

\bibitem[OLT06]{Olson2006fastiterative}
\textsc{Olson E., Leonard J., Teller S.}:
\newblock Fast iterative alignment of pose graphs with poor estimates.
\newblock In \emph{ICRA} (2006).

\bibitem[OT06]{oliva2006building}
\textsc{Oliva A., Torralba A.}:
\newblock Building the gist of a scene: The role of global image features in
  recognition.
\newblock \emph{Progress in brain research} (2006).

\bibitem[Pho12]{PhotosynthURL}
\textsc{Photosynth}:.
\newblock \url{http://www.photosynth.net}, 2012.

\bibitem[Pri12]{PrinceCVMLI2012}
\textsc{Prince S.}:
\newblock \emph{{Computer Vision: Models Learning and Inference}}.
\newblock {Cambridge University Press}, 2012.

\bibitem[PRRAZ00]{Peleg:2000:MAM}
\textsc{Peleg S., Rousso B., Rav-Acha A., Zomet A.}:
\newblock Mosaicing on adaptive manifolds.
\newblock \emph{PAMI} (2000).

\bibitem[RAKRF08]{rav2008unwrap}
\textsc{Rav-Acha A., Kohli P., Rother C., Fitzgibbon A.}:
\newblock Unwrap mosaics: a new representation for video editing.
\newblock In \emph{TOG} (2008), vol.~27.

\bibitem[RAPLP05]{rav2005dynamosaics}
\textsc{Rav-Acha A., Pritch Y., Lischinski D., Peleg S.}:
\newblock Dynamosaics: Video mosaics with non-chronological time.
\newblock In \emph{{CVPR}} (2005), vol.~1, pp.~58--65.

\bibitem[SEE{\etalchar{*}}12]{Sturm2012TUM}
\textsc{Sturm J., Engelhard N., Endres F., Burgard W., Cremers D.}:
\newblock A benchmark for the evaluation of rgb-d slam systems.
\newblock In \emph{Proc. of the International Conference on Intelligent Robot
  Systems (IROS)} (Oct. 2012).

\bibitem[SFC{\etalchar{*}}11]{Shotton:PoseRecognition:2011}
\textsc{Shotton J., Fitzgibbon A., Cook M., Sharp T., Finocchio M., Moore R.,
  Kipman A., Blake A.}:
\newblock {Real-Time Human Pose Recognition in Parts from a Single Depth
  Image}.
\newblock \emph{CVPR} (2011).

\bibitem[SGSS08]{SGSS-siggraph08}
\textsc{Snavely N., Garg R., Seitz S.~M., Szeliski R.}:
\newblock Finding paths through the world's photos.
\newblock \emph{SIGGRAPH} (2008).

\bibitem[SH99]{shum1999rendering}
\textsc{Shum H., He L.}:
\newblock Rendering with concentric mosaics.
\newblock In \emph{{SIGGRAPH}} (1999).

\bibitem[SS97]{Szeliski:1997:CFV}
\textsc{Szeliski R., Shum H.-Y.}:
\newblock Creating full view panoramic image mosaics and environment maps.
\newblock SIGGRAPH.

\bibitem[SSS06]{Snavely:Phototourism:2006}
\textsc{Snavely N., Seitz S., Szeliski R.}:
\newblock Photo tourism: exploring photo collections in 3d.
\newblock \emph{SIGGRAPH} (2006).

\bibitem[Sze96]{Szeliski96videomosaics}
\textsc{Szeliski R.}:
\newblock Video mosaics for virtual environments.
\newblock \emph{IEEE Computer Graphics and Applications 16} (1996), 22--30.

\bibitem[Sze06]{Szeliski:2006:IAS}
\textsc{Szeliski R.}:
\newblock Image alignment and stitching: a tutorial.
\newblock \emph{Found. Trends. Comput. Graph. Vis.} (2006).

\bibitem[TDSL00]{tenenbaum2000isomap}
\textsc{Tenenbaum J., De~Silva V., Langford J.}:
\newblock A global geometric framework for nonlinear dimensionality reduction.
\newblock \emph{Science 290}, 5500 (2000), 2319--2323.

\bibitem[TFBD01]{Thrun:RobustLocalization:2001}
\textsc{Thrun S., Fox D., Burgard W., Dellaert F.}:
\newblock Robust monte carlo localization for mobile robots.
\newblock \emph{AI} (2001).

\bibitem[TM07]{TuytelaarsM07}
\textsc{Tuytelaars T., Mikolajczyk K.}:
\newblock Local invariant feature detectors: A survey.
\newblock \emph{Found. Trends. Comput. Graph. Vis.} (2007).

\bibitem[WACS11]{Wu:MCBA:2011}
\textsc{Wu C., Agarwal S., Curless B., Seitz S.}:
\newblock Multicore bundle adjustment.
\newblock \emph{{CVPR}} (2011).

\bibitem[WMLS10]{WagnerMLS10}
\textsc{Wagner D., Mulloni A., Langlotz T., Schmalstieg D.}:
\newblock Real-time panoramic mapping and tracking on mobile phones.
\newblock VR.

\bibitem[Wu07]{Wu:SIFTGPU:2007}
\textsc{Wu C.}:
\newblock Sift{G}{P}{U}: A {G}{P}{U} implementation of scale invariant feature
  transform ({S}{I}{F}{T}).
\newblock \url{http://cs.unc.edu/~ccwu/siftgpu}, 2007.

\bibitem[YN01]{YouNeumannGyro01}
\textsc{You S., Neumann U.}:
\newblock Fusion of vision and gyro tracking for robust augmented reality
  registration.
\newblock In \emph{VR} (2001).

\end{thebibliography}

\begin{appendices}

\begin{figure*}
 \begin{subfigure}[t]{0.98\textwidth}
  \centering
  \begin{subfigure}[b]{0.48\textwidth}        
   {        
    \setlength{\fboxsep}{0pt}
    \setlength{\fboxrule}{.2pt}
    \fbox{\includegraphics[width=\columnwidth]{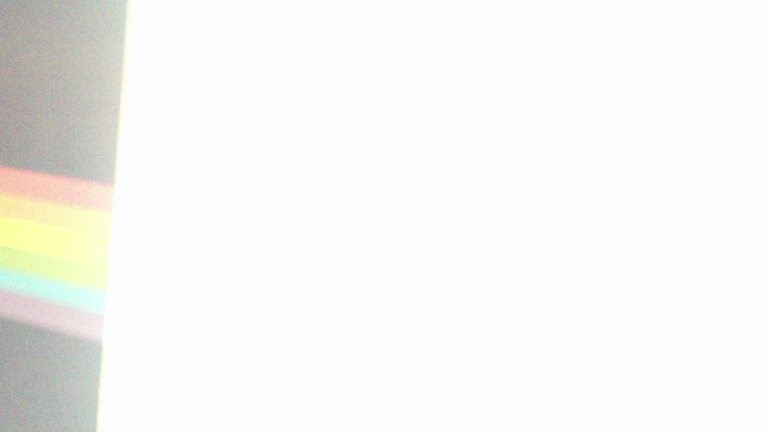}}        
   }        
   \caption{\dataset{prism} frame 165}        
   \label{fig:prism_a}        
  \end{subfigure}        
  \begin{subfigure}[b]{0.48\textwidth}        
   {        
    \setlength{\fboxsep}{0pt}
    \setlength{\fboxrule}{.2pt}
    \fbox{\includegraphics[width=\columnwidth]{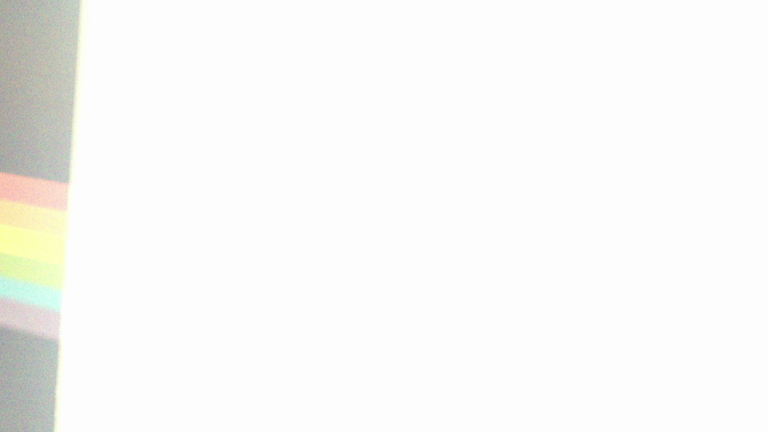}}
   }        
   \caption{\dataset{prism} frame 166}        
   \label{fig:prism_b}             
  \end{subfigure}        
 \end{subfigure}        

 \begin{subfigure}{0.98\textwidth}        
  \centering
  \begin{subfigure}{0.32\textwidth}        
   \includegraphics[width=\columnwidth]{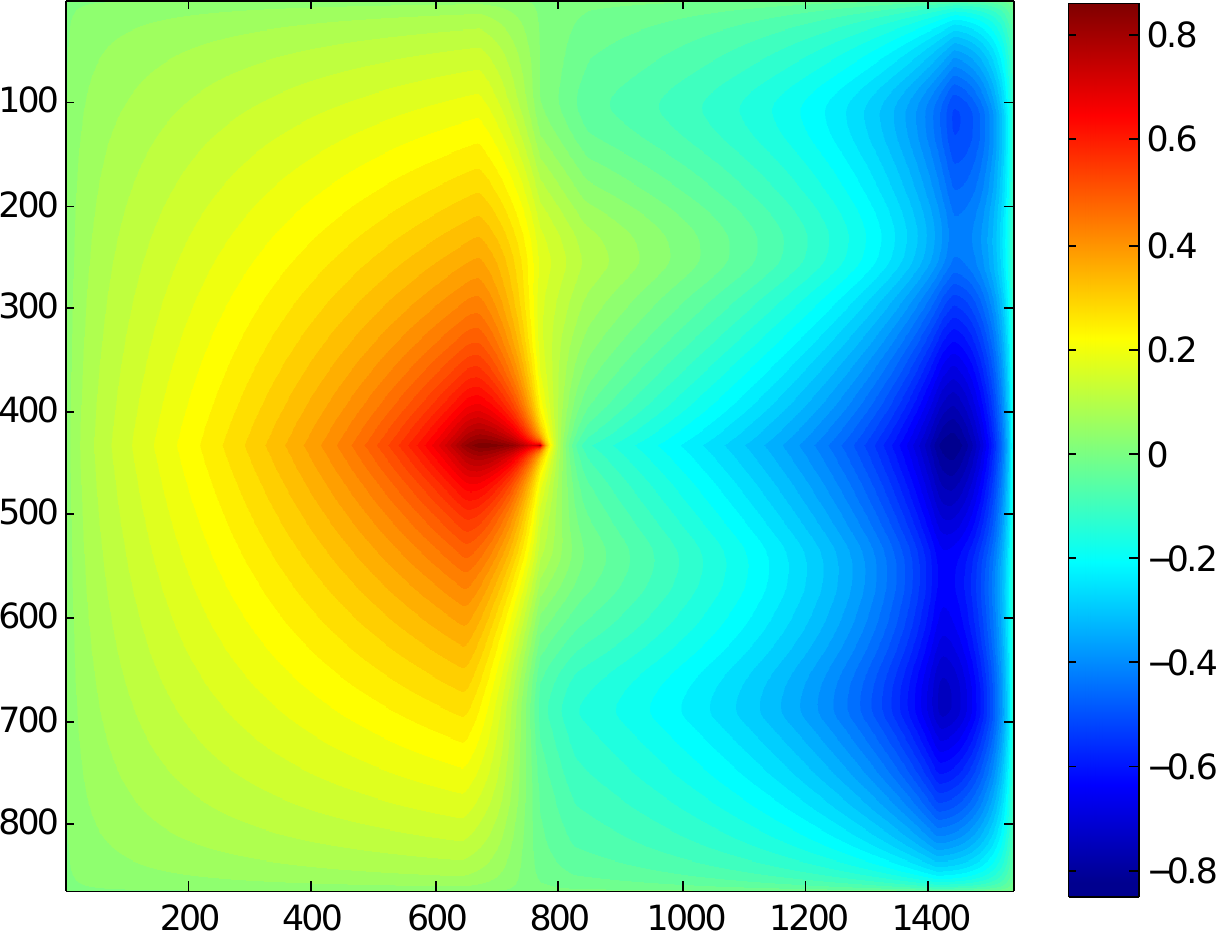}        
   \caption{NCC result from MATLAB \dataset{normxcorr2} between images}        
   \label{fig:prism_ncc_result}        
  \end{subfigure}        
 \end{subfigure}        

 \begin{subfigure}{0.98\textwidth}        
  \begin{subfigure}{0.45\textwidth}        
   \includegraphics[width=\columnwidth]{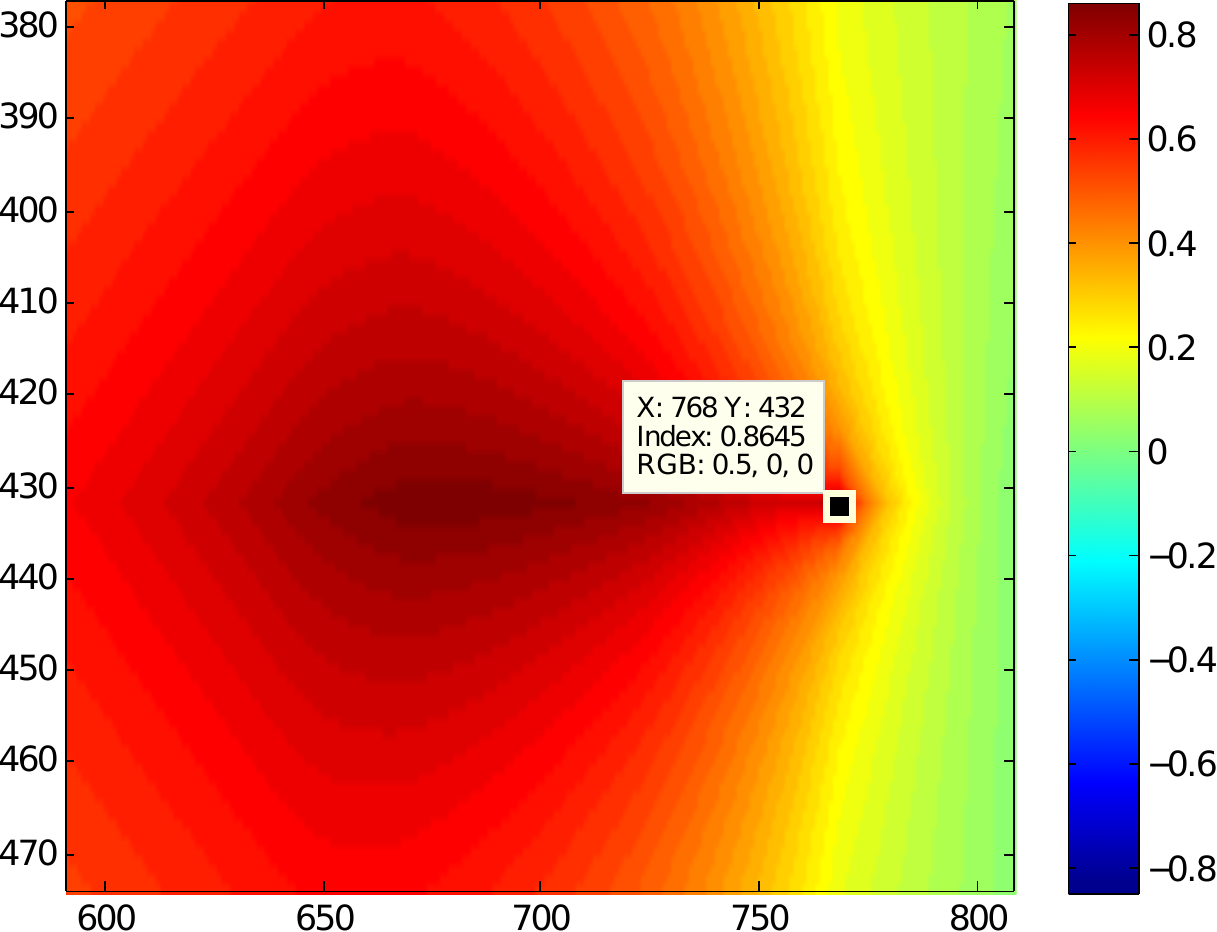}
   \caption{Primary peak (greatest NCC value)}        
   \label{fig:prism_ncc_peak_1}        
  \end{subfigure}        
  \begin{subfigure}{0.45\textwidth}        
   \includegraphics[width=\columnwidth]{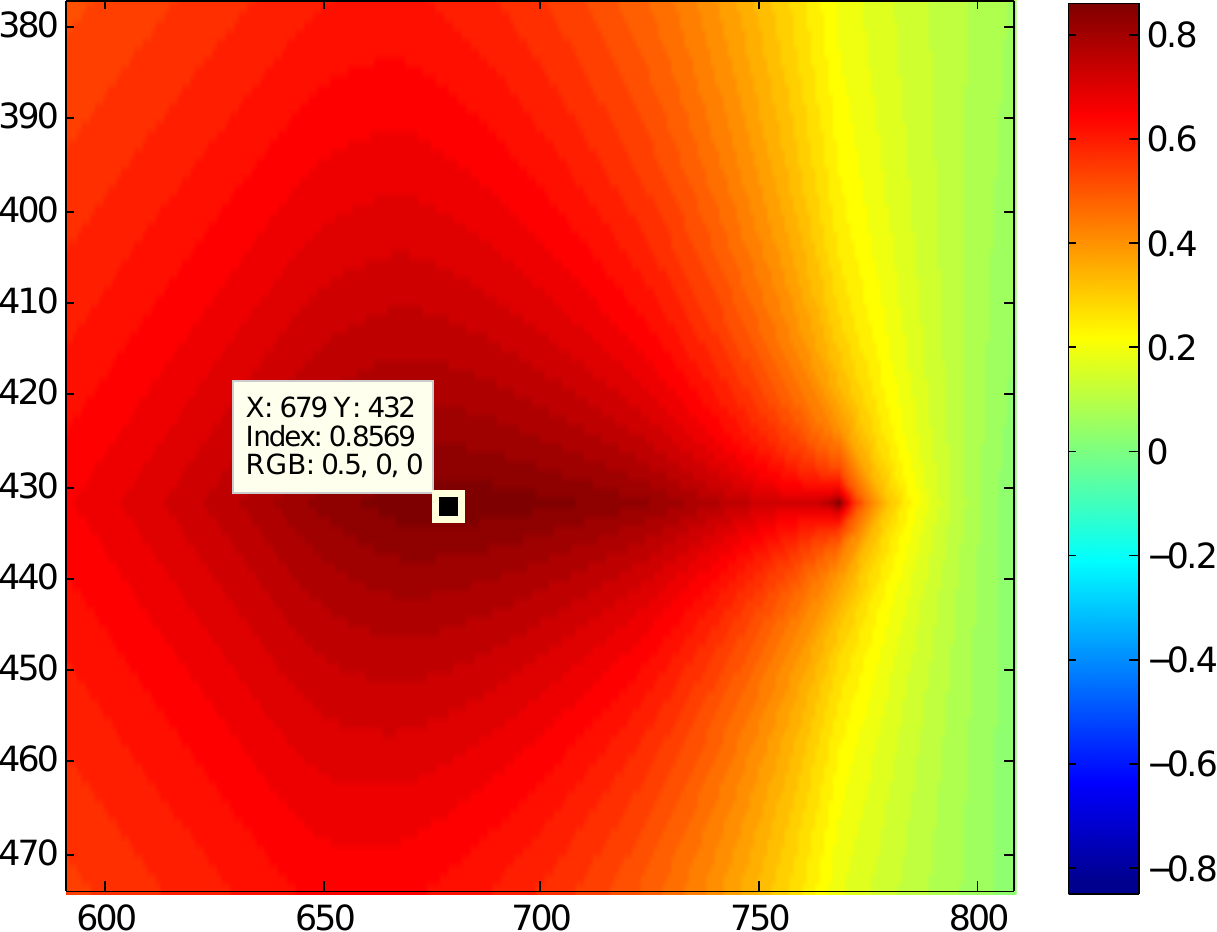}
   \caption{Secondary peak (lower NCC value)}        
   \label{fig:prism_ncc_peak_2}        
  \end{subfigure}        
 \end{subfigure}        
 \caption{NCC failing to compute the correct offset for two frames in \dataset{prism}.
 Black borders added to top images for clarity.}        
 \label{fig:ncc_failure}        
\end{figure*}

\section{Baseline Algorithms}
\label{sec:baseline_algorithms}

In this document we provide further results of comparisons between our system and Wu \etal's VisualSFM~\shortcite{Wu:SIFTGPU:2007,Wu:MCBA:2011}, Viewfinder Alignment (VfA) of Adams~\etal~\shortcite{adams2008viewfinderalignment}, a ``micro-SfM'' method which computes a 2D translation using SIFT matching with RANSAC, and direct (all-pixel) methods as summarized by Szeliski~\cite{Szeliski:2006:IAS}.
For each baseline algorithm we have the capability to load the output into our viewer for qualitative comparison.
To load VisualSfM's output of 6\dof camera positions into the Swipe Mosaic viewer we project each camera to a location and orientation on the $2$D plane.
We do this by computing a normal vector from the average ``forward'' direction over all cameras, then projecting each camera perpendicularly onto a plane defined by this normal. The location on this plane gives a $2$D coordinate to be loaded into the viewer (see main paper).

\subsection{Direct Methods}
\label{sec:comparison_to_direct_methods}

An excellent summary of techniques for Image Alignment and Stitching was presented by Szeliski~\cite{Szeliski:2006:IAS}.
Chapter 3, ``Direct (pixel-based) alignment'', details a number of methods to compute a 2D alignment between image pairs.
The general approach is to define some error metric which can evaluate how well each potential 2D alignment matches the contents of each image.
Given the error metric, one can exhaustively evaluate all possible alignments, or use a coarse to fine method to limit the amount of computation, and the alignment which produces the lowest error is chosen.
Various error metrics can be defined on overlapping pixels, such as Sum of Square Differences or Sum of Absolute Differences.
We compare to the popular method of Normalized Cross Correlation (NCC), which we also used as the basis of our feature vector computation.
NCC is an improvement over improves over Cross Correlation (which has a tendency to give incorrect offsets in the presence of large high intensity areas) by normalizing the overall intensity of each of the regions being compared.
However, as noted by Szeliski, \emph{``[NCC's] performance degrades for noisy low-contrast regions''} so the improved technique is not immune to problems.

Two temporally adjacent frames (\figref{fig:prism_a} and \figref{fig:prism_b}) from the \dataset{prism} sequence were selected.
Some scene geometry is visible on the left of the image, but most of the pixels have been overloaded by the bright light and are reporting close to perfect white.
Nevertheless, it is clear that horizontal camera motion has taken place, given the parts of the geometry which we \emph{can} see.
The NCC image computed in MATLAB is shown in \figref{fig:prism_ncc_result}.
Two closeups of the region around the peak are shown in \figref{fig:prism_ncc_peak_1} and \figref{fig:prism_ncc_peak_2}.
Note that the right hand peak in \figref{fig:prism_ncc_peak_1} has a higher NCC value, but the location $(768, 432)$ implies that \emph{zero} translation is the optimal image alignment.
The left hand peak in \figref{fig:prism_ncc_peak_2} has a slightly lower magnitude, but the offset $(679, 432)$ indicates a horizontal offset of $(768 - 679) / 2 = 39$.
Indeed, a translational shift of $(39, 0)$ does bring the two images into good alignment.
Our RRF based system produced an estimate  with mean $(0.0169, 0.0004)$ and variance $(0.00132, 0.00168)$ for these images (note these results are not in units of pixels as above, so cannot be directly compared), showing primarily horizontal motion with roughly isotropic variance, as we would expect from the fact that the visible texture in the image confirms no vertical motion has taken place.
Any algorithm which simply computes NCC over entire images, finds the single peak and uses the value will be prone to fail in image pairs such as this, whereas our system produces a translation in the correct direction.
Note that un-normalized Cross Correlation for this image actually produced a \emph{purely vertical} translation, reinforcing the idea that un-normalized cross correlation is unsuitable for large high intensity regions, performing even worse than NCC.

\begin{figure*}
 \begin{subfigure}[b]{0.24\textwidth}
  \includegraphics[width=\columnwidth]{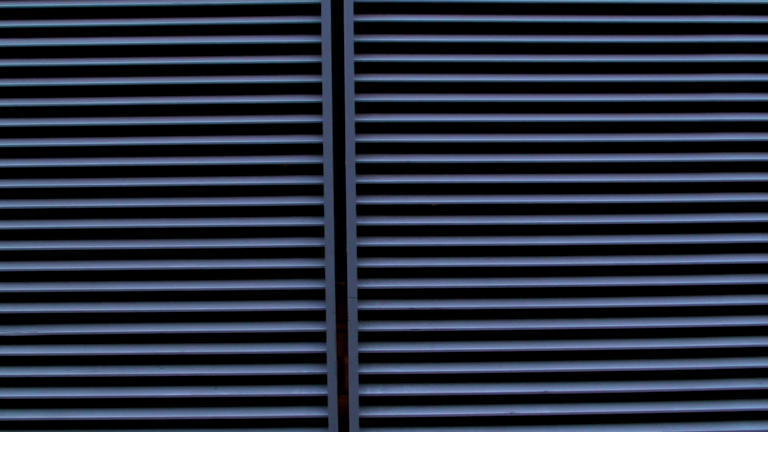}
  \caption{Image A}
  \label{fig:grate_a}
 \end{subfigure}
 \begin{subfigure}[b]{0.24\textwidth}
  \includegraphics[width=\columnwidth]{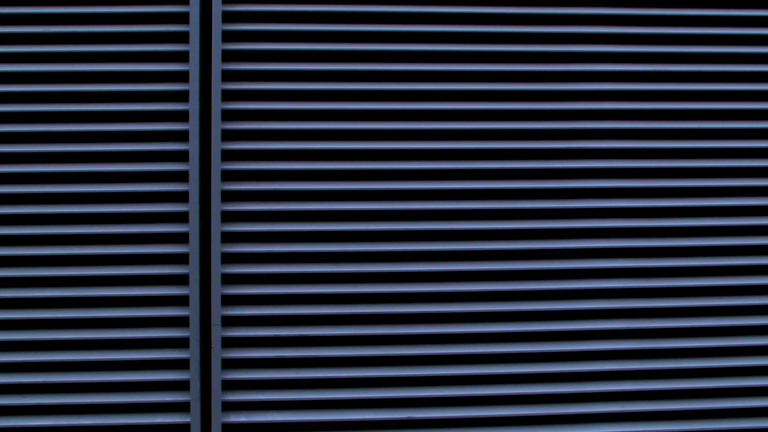}
  \caption{Image B}
  \label{fig:grate_b}     
 \end{subfigure}
 \begin{subfigure}[b]{0.24\textwidth}
  \includegraphics[width=\columnwidth]{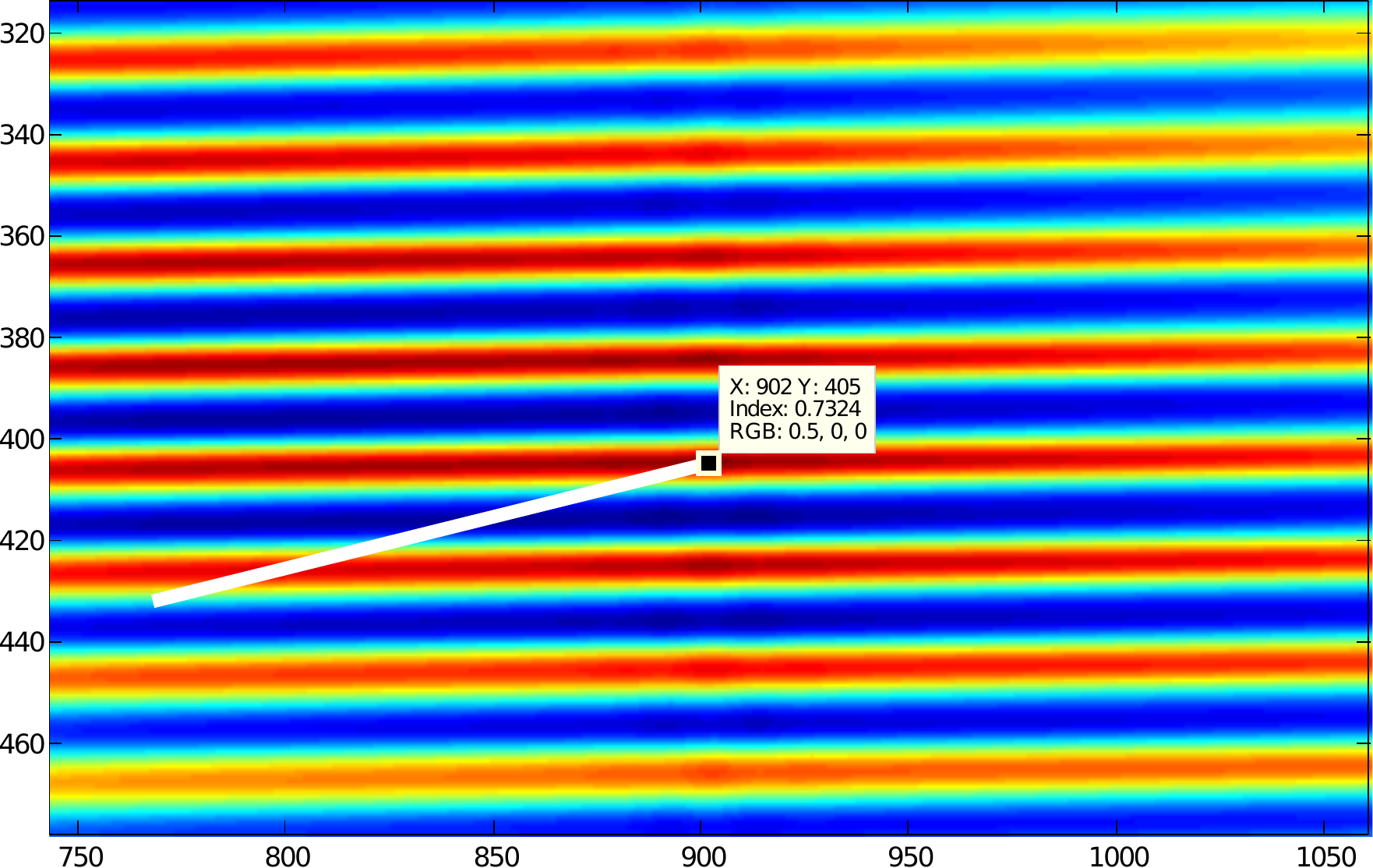}
  \caption{Detail of NCC response with peak and zero location marked.}
  \label{fig:grate_ncc}   
 \end{subfigure}
 \begin{subfigure}[b]{0.24\textwidth}
  \includegraphics[width=\columnwidth]{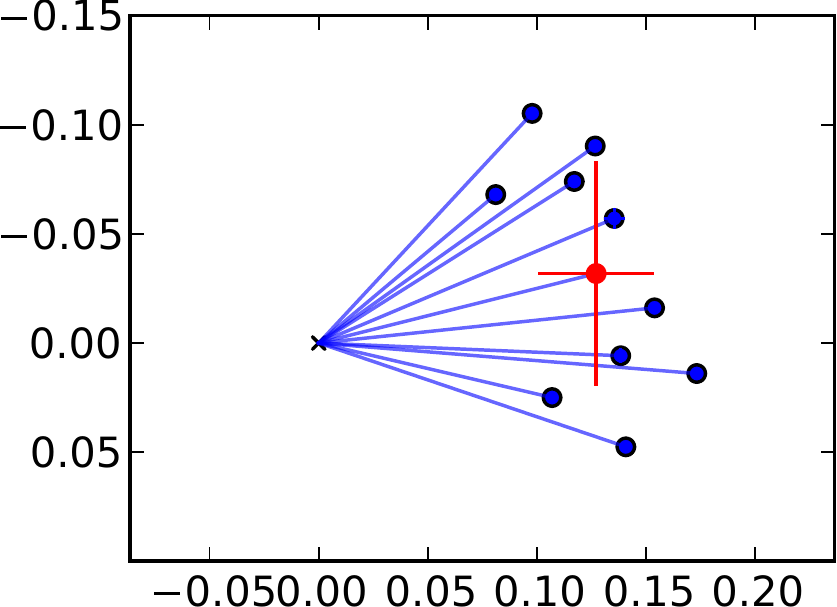}
  \caption{Distribution returned by our method.}
  \label{fig:grate_ours}  
 \end{subfigure}
 \caption{NCC computes a large diagonal offset for 2 frames containing repeated structure.
 The white line in c) connects the NCC peak with the location which would represent zero offset}
 \label{fig:ncc_failure_grate}
\end{figure*}

Our method returns a distribution over possible camera motion, which is a great advantage in cases of ambiguity such as repeated structure. 
\figref{fig:ncc_failure_grate} shows how in the presence of multiple potential alignments.
Another example of how our method is superior to NCC is in  shown in \figref{fig:ncc_failure_grate}.
When repeated structure creates a number of possible alignments, our method (\figref{fig:grate_ours}) returns an anisotropic estimate compared to the deterministic estimate provided by NCC (\figref{fig:grate_ncc}).
Corresponding repeated structure is shown in the NCC response image; the peak happens to be located in a ridge which indicates slight upward motion (as well as rightwards).
The vertically adjacent ridges have similar NCC values and (given the magnitude of the camera motion between the images) are surely almost as likely.
However the single transformation returned by pure NCC alignment will not represent this information at all.
The result from our method is more desirable in this situation.

\newpage
\subsection{Structure from Motion: VisualSfM}
\label{subsub:comparison_sfm}

\begin{figure}
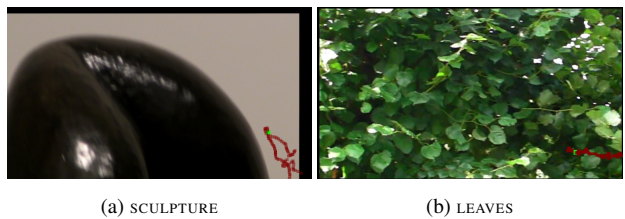

 \begin{subfigure}[t]{0.48\columnwidth}
  \includegraphics[width=\columnwidth]{results/result_sculpture.jpg}
  \caption{\footnotesize \dataset{sculpture}}
  \label{fig:result_sculpture_sfm}
 \end{subfigure}
 \begin{subfigure}[t]{0.48\columnwidth}
  \includegraphics[width=\columnwidth]{results/result_leaves.jpg}
  \caption{\footnotesize \dataset{leaves}}
  \label{fig:result_leaves_sfm}
 \end{subfigure}
 \caption{\footnotesize Screenshots in the Swipe Mosaic interface of the datasets on which we compare our performance to SfM.
 The minimap in the bottom right shows the camera locations.}
 \label{fig:sfm_failure_case}
\end{figure}

We selected the \dataset{sculpture} (\figref{fig:result_sculpture_sfm}) and \dataset{leaves} (\figref{fig:result_leaves_sfm}) sequences as likely to cause SfM failure.
\dataset{sculpture} includes specularities, motion blur, and has few suitable corners for interest point detection.
\dataset{leaves} contains lots of geometry suitable for interest point detection, but most of these areas are on leaves, which are being blown around in the wind, meaning points detected on them may adversely contribute to the optimization.

VisualSfM was run on each sequence in both ordered and unordered mode.
This mode affects which image pairs are compared to find interest points; either all pairs (unordered) or only temporally adjacent pairs (ordered).

Running unordered on \dataset{sculpture}, $38$ camera locations were reconstructed from the $101$ input images, leading to an incomplete Swipe Mosaic.
Running ordered mode yields an even worse result, reconstructing $35$ camera locations but in three independent groups.
The locations that VisualSfM did produce were accurate, but the full camera path produced by our system is preferable, as shown in \figref{fig:sculpture_ours_and_visualsfm}.

The \dataset{leaves} dataset consist of $201$ images.
VisualSfM in ordered mode computed locations for only $70$ of the images, albeit producing a reasonable Swipe Mosaic.
Running SfM in unordered mode computes a location for all of the images, but the placement undergoes a catastrophic failure, with the resulting Swipe Mosaic suffering severe artifacts.
The failure takes the form of one side of the horizontal path being relatively correct, and the image locations gradually worsening as we travel along the video timeline, until the predicted image locations do not even overlap (see supplemental video).
For both the sequences in this section, our system produced easily navigable locations for all cameras (see video and~\figref{fig:sfm_failure_case}).
We surmise that unordered mode producing better results in this case was due to a greater variety of image pairs being run through SIFT matching, rather than merely a few temporal neighbors.
If a several consecutive frames of video are blurred or contain confusing motion, unordered mode will still be able to search for SIFT matches between frames ``either side'' of the problem area, thus providing a more robust solution.

\begin{figure}
 \begin{center}
  \def\svgwidth{\columnwidth}
  \input{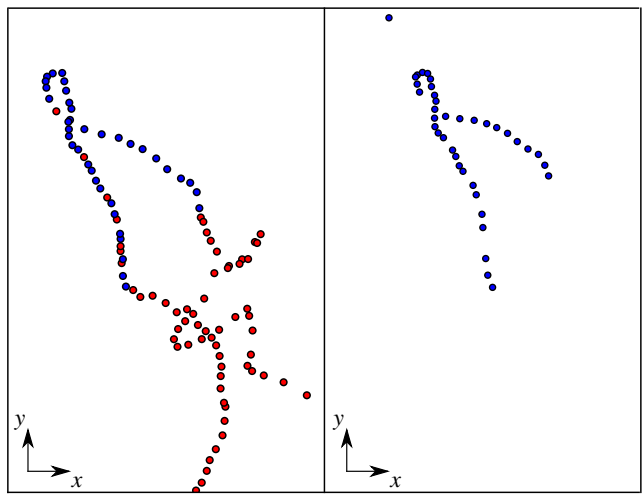}
 \end{center}
 \caption{\footnotesize $2$D camera coordinates (unitless) for \dataset{sculpture} produced by our system (left) and VisualSFM (right).
 Blue points indicate images where both systems gave an estimate of location (note the estimates differ); red points are images where only our system produced an estimate.
 Note the obvious outlier at the top of the SfM result.}
 \label{fig:sculpture_ours_and_visualsfm_second}
\end{figure}

\subsection{$\mu$SfM}
\label{sub:_mu_sfm}

\begin{figure}
 \includegraphics[width=\columnwidth]{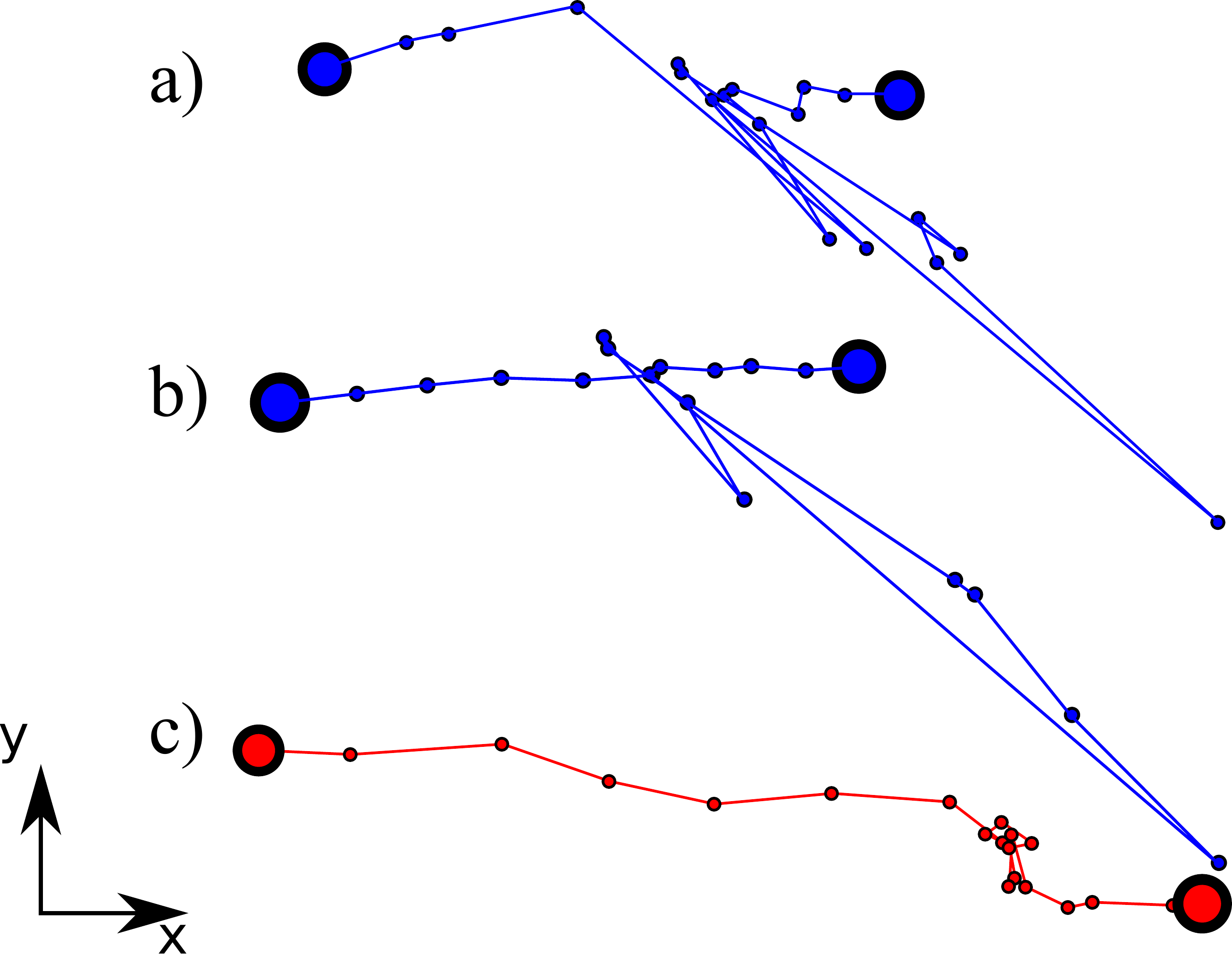}
 \caption{\footnotesize Regularized locations for the \dataset{vinyl} scene, for the images which straddle the obstruction.
 Dots represent image locations and the line joins them in temporal order.
 Larger dots show the start and end of this subsequence.
 As the correct camera motion is approximately a constant horizontal velocity, the ideal result would be equally spaced dots on a horizontal line.
 Scale between the $3$ diagrams is not meaningful.
 a): $\mu$SfM result with unweighted regularization.
 b): $\mu$SfM with weighted regularization.
 c): our system.}
 \label{fig:vinyl_result}
\end{figure}

``micro-SfM'' or ``$\mu$SfM'' is a system which we have developed with the intention of it being an equivalent system to SfM, but without computing any structure, and with camera transforms limited to $2$D translation.
The thinking behind this is that computing a $6$D quantity for each frame is an inherently harder task than computing a 2D quantity for each frame, and so simply comparing our $2$D RRF method to VisualSfM was not a fair comparison.
Rather, we should apply the technique from VisualSfM to the strictly easier problem of computing translations (not fundamental matrices) in order to compare like-for-like results.
``$\mu$SfM'' matches two images by generating SIFT descriptors and performing matching using the standard algorithm of Lowe~\shortcite{Lowe04:ijcv}.
A translation is computed using RANSAC to iteratively select a random SIFT match, compute the corresponding $2$D transform and count the number of inliers.
The transform with the highest inlier count is used to generate a final refined transform from the entire inlier set.
To combine multiple translation estimates across an image sequence, we use the least-squares based layout algorithm developed for our RRF estimates.
As the layout algorithm allows for a weighting to be applied to each relative transform (in our main system we use the inverse variance from the forest) we test two versions of $\mu$SfM -- one unweighted, and one weighted using the ratio of inliers to total number of matches found when generating the transform.
This should ensure that translations for which every single SIFT match agrees are given more weight by the optimization.

Considering the simplicity of the method, $\mu$SfM is surprisingly capable.
In particularly, it can produce just as good a camera path for the \dataset{leaves} sequence as our technique.
However as the method relies entirely on interest point matching, we know it is susceptible to fail in the presence textureless regions, motion blur or repeated structure.
The \dataset{vinyl} sequence contains a blurry obstruction which is very close to the camera, separating to regions containing strong texture information.
The camera travels horizontally, starting in one textured region, passing the obstruction (which takes up the whole screen for a few frames) and ends viewing the second textured region.
Surprisingly, inside the (apparently) textureless region, SIFT \emph{is} able to detect a few interest points.
Matching these interest points prooves difficult however; most of them have extremely similar appearances, and despite implementing Lowe's technique for avoiding ambiguous matches, the translations returned from the middle frames in this sequence were extremely noisy.
Note the results in \figref{fig:vinyl_result}, bearing in mind the ideal answer would be almost pure horizontal motion.
Both $\mu$SfM results display problems with some frames ending up at a large displacement to the lower right corner of the map, with the subsequent frames on the normal timeline.
It can be seen that the weighted version displays a smoother timeline at the beginning and end of the sequence, but for both a) and b) the mistakes in the middle of the sequence make this difficult to navigate in our interface (frames displaying the obstruction are incorrectly displayed amongst the frames of texture objects).
By contrast our result, c), whilst by no means perfect, is a vast improvement on both $\mu$SfM results.
For the images where texture is available it computes a consistent horizontal motion.
For the frames containing no texture, there is insufficient information to state which (if any) direction the camera has moved, so our system returns a number of zero mean, wide variance offsets.
The optimization places these roughly on top of each other, generating the point cluster in the result.
Obviously this is not actually correct, as the camera was always moving, but as there is no way to tell this simply from the images pairs we produce a reasonable result, which allows the sequence to be browsed as a Swipe Mosaic without artifacts.
It may be possible to improve this aspect of the system by using a camera motion model in the layout algorithm, meaning that when we knew the first few frames had the camera move to the right, then when presented with insufficient visual information our estimate would be some kind of rightwards motion, rather than zero mean motion.
Another failure case for $\mu$SfM was \dataset{prism}, where the camera autogain causes whiteout for a few frames.
Similarly to the previous sequence, no reliable feature matches could be detected during this central part of the image sequence, leading to incorrect matches in the middle of the sequence again.

\subsection{Viewfinder Alignment}

Our final baseline comparison is to Viewfinder Alignment (VfA) of Adams~\etal~\shortcite{adams2008viewfinderalignment}.
VfA is a method to compute constrained transforms between temporally close video frames.
VfA computes a ``digest'' for each frame by encoding edge information at $4$ equally spaced orientations using gradient integral projection arrays, as well as detecting the top $k$ peaks in the image.
Two image digests are aligned by first calculating a single $2$D shift which best aligns the edge information stored for each image.
This $2$D shift is applied to the detected corners of one of the digests.
The number of inliers (a pair of points, one from each image, landing within $3$ pixels from each other) between the two points sets is counted and taken as the confidence that the images have been aligned correctly.
The set of inliers is used to generate a similarity transform, giving $4$ degrees of freedom (translation, rotation and scale) between pairs of frames.
Note that our VfA test scenes were chosen intentionally so that the rotation and scale change was negligible so these parameters are ignored, \ie we are only interested in the relative accuracy of the \emph{translations} computed by different methods.
Experiments showed that when compute the scale change was typically between $0.98$ and $1.02$, and the rotation on the order of $0.1$ radians, justifying this decision.

VfA has a number of attractive properties, including computational efficiency and being extremely resistant to noise.
A disadvantage of the algorithm is that it is completely deterministic, in that only one $2$D translation between each frame pair is considered, when the digest edge information could be used to produce a distribution of translations.
Additionally, the corners returned from the corner detector are simply stored as 2D locations, without any kind of descriptor, allowing corners which represent different scenes points to potentially be aligned with each other and treated as an inlier.
We compared our system to our own re-implementation of VfA.
This code is supplied in the appendix. We now present detailed analysis of VfA on various test scenes.

All the sequences in this document were run through our re-implementation of Viewfinder Alignment.
We tried to match each digest with the digests from other images which were within $6$ frames (forward or backwards).
If a complete graph could be constructed, we used all the inferred translation values as input to our linear least squares regularization (see main paper).
Inlier matrices are shown using the standard Jet colormap, except that pairs which either produced zero inliers or were not compared (\ie they were too far apart temporally) are left blank.

\subsection*{Successes}

\subsubsection*{Lobby}

\begin{figure}
 \includegraphics[width=0.49\columnwidth]{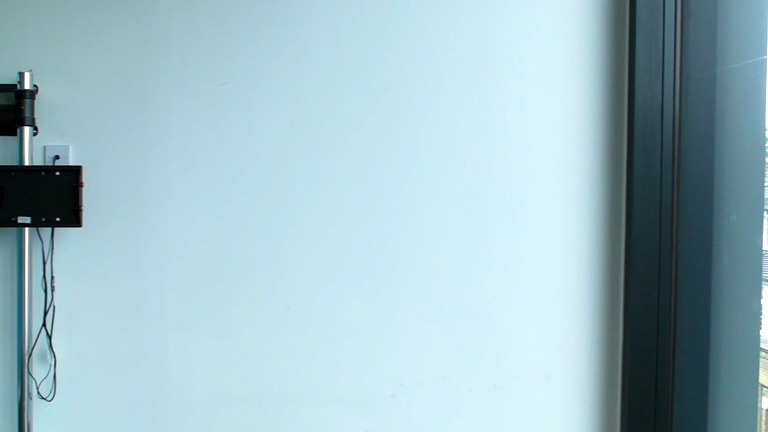}
 \includegraphics[width=0.49\columnwidth]{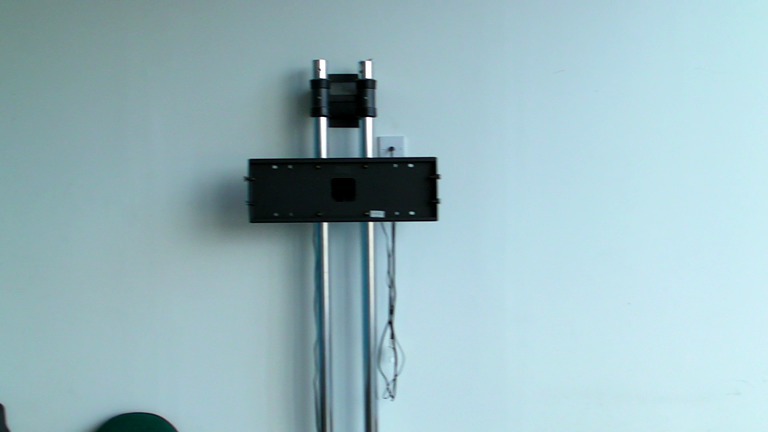}
 \includegraphics[width=0.49\columnwidth]{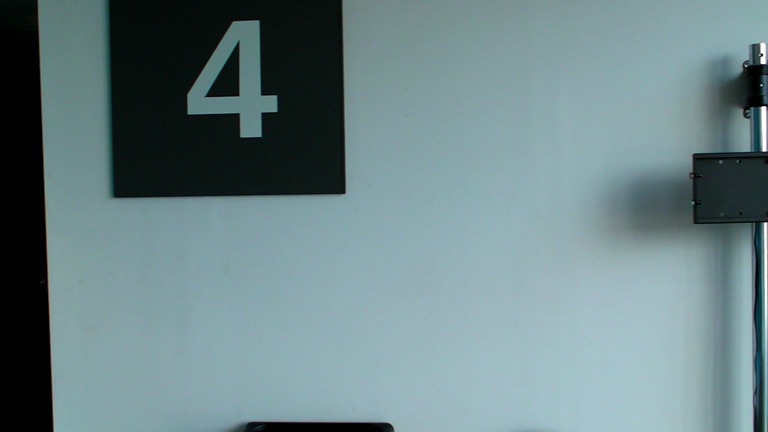}
 \includegraphics[width=0.49\columnwidth]{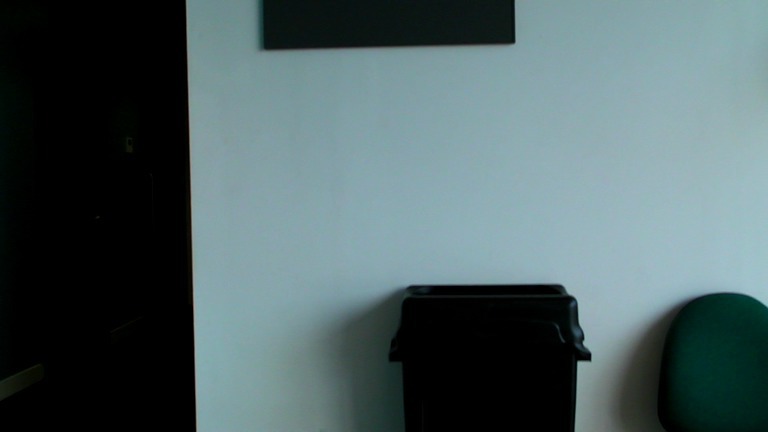}
 \caption{Sample Images from \texttt{lobby} sequence.}
 \label{fig:lobby_images}
\end{figure}

\begin{figure}
 \includegraphics[width=\columnwidth]{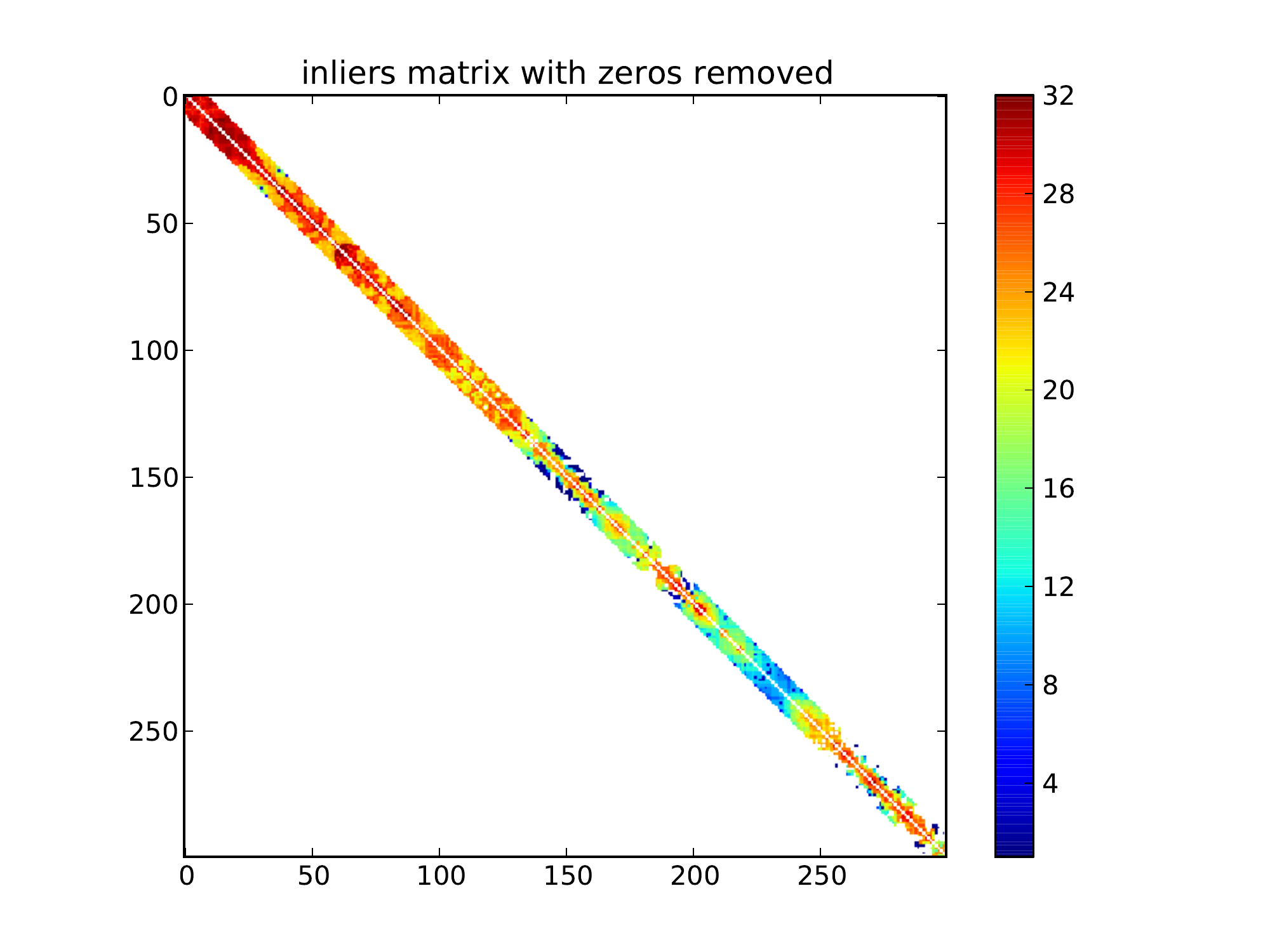}
 \caption{Number of inliers from lobby sequence.}
 \label{fig:lobby_inliers}   
\end{figure}

The lobby sequence is largely featureless, but the objects that are seen display strong vertical and horizontal edges, resulting in an excellent output when browsed as a Swipe Mosaic.

\newpage
\subsubsection*{Fence}

\begin{figure}
    \includegraphics[width=0.49\columnwidth]{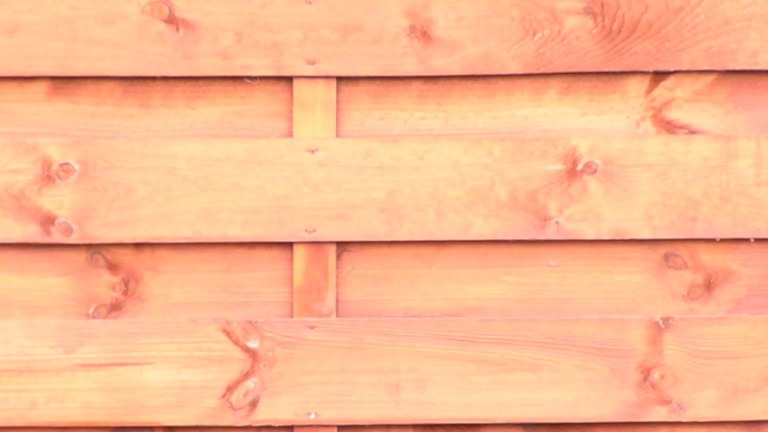}
    \includegraphics[width=0.49\columnwidth]{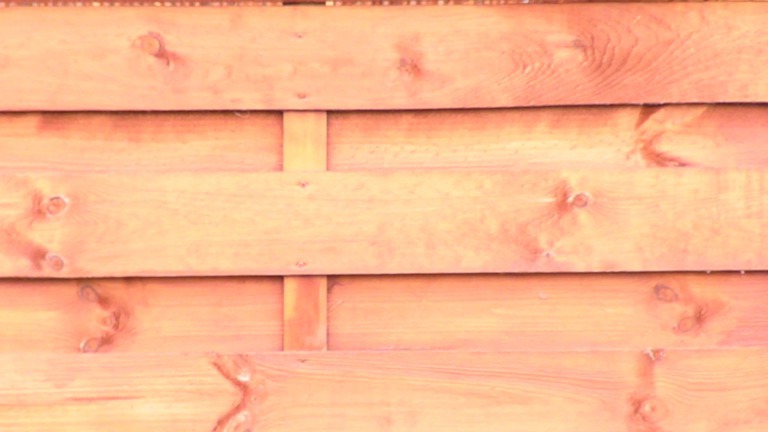}
    \includegraphics[width=0.49\columnwidth]{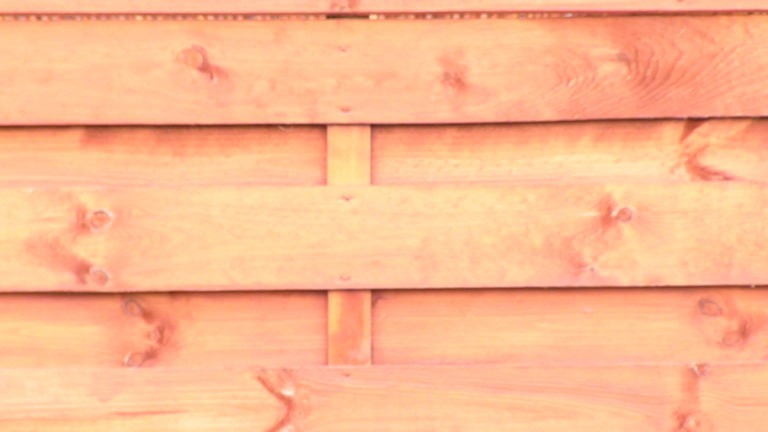}
    \includegraphics[width=0.49\columnwidth]{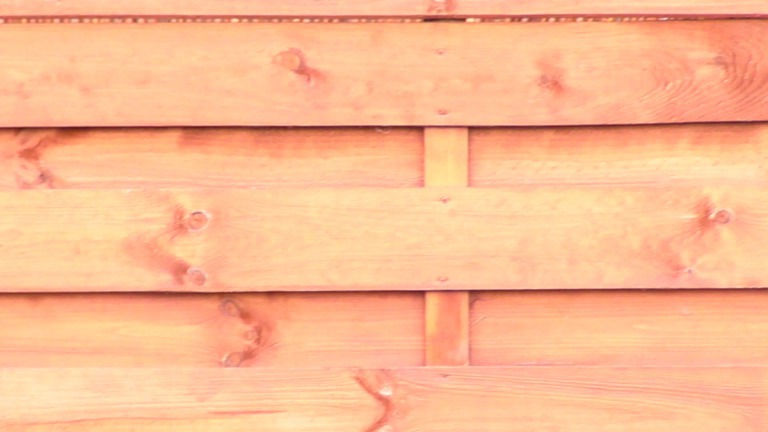}
    \caption{Sample Images from \texttt{fence} sequence.}
    \label{fig:fence_images}
\end{figure}

\begin{figure}
    \includegraphics[width=\columnwidth]{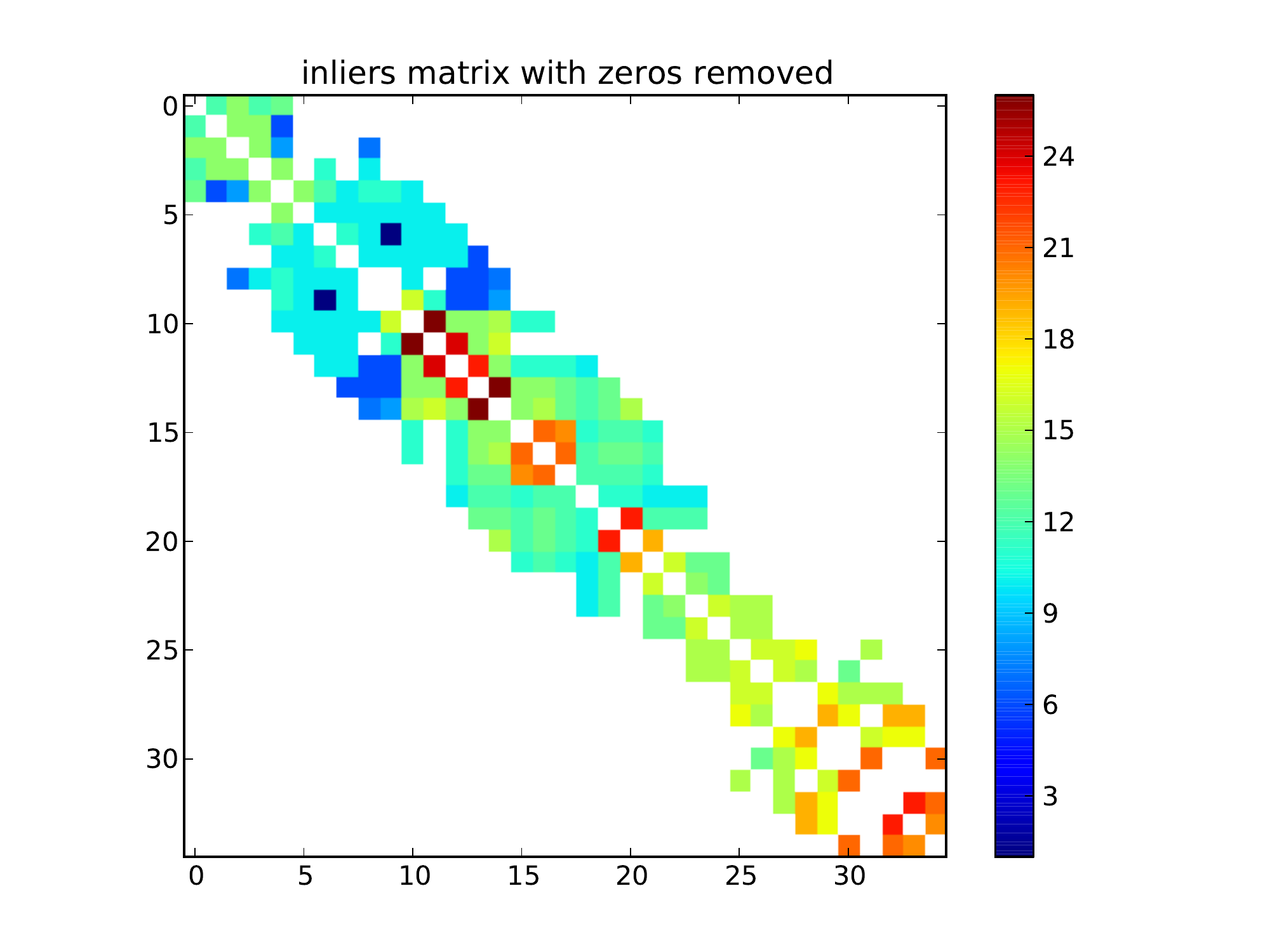}
    \caption{Number of inliers from fence sequence.}
    \label{fig:fence_inliers}   
\end{figure}

The fence contains repeated structure with many similar looking horizontal edges but VfA is robust to this, producing a fully connected set of inliers (\figref{fig:fence_inliers}) and a correct reconstruction.

\newpage
\subsection*{Failure cases}

\subsubsection*{Vinyl}

\begin{figure}
 \includegraphics[width=0.49\columnwidth]{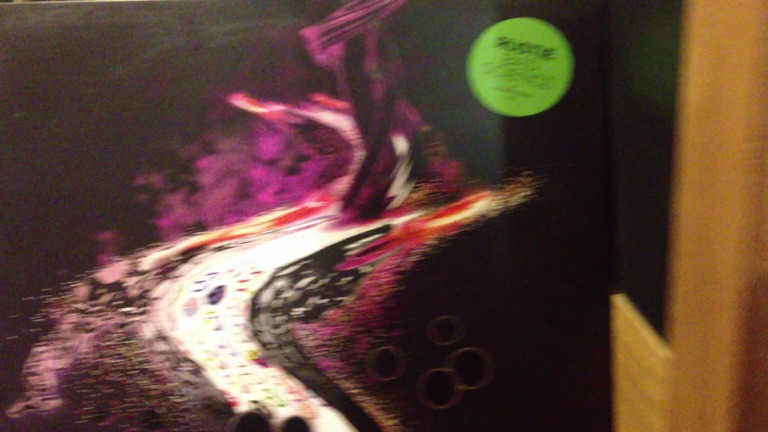}
 \includegraphics[width=0.49\columnwidth]{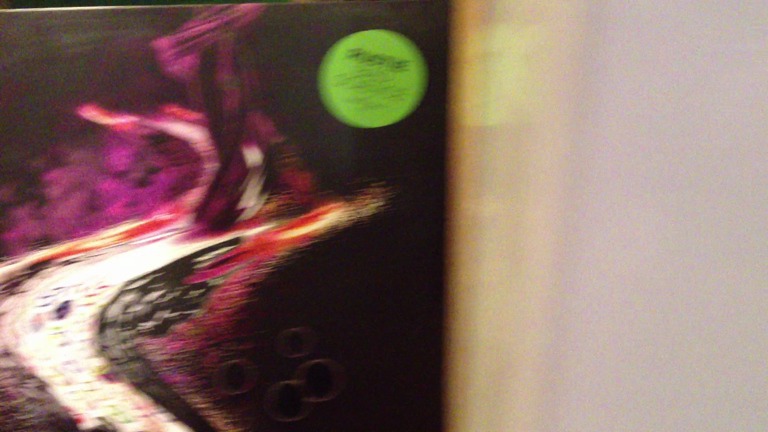}
 \includegraphics[width=0.49\columnwidth]{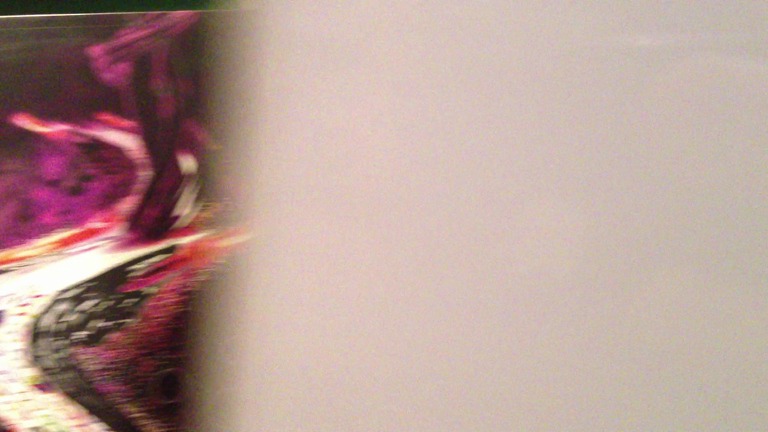}
 \includegraphics[width=0.49\columnwidth]{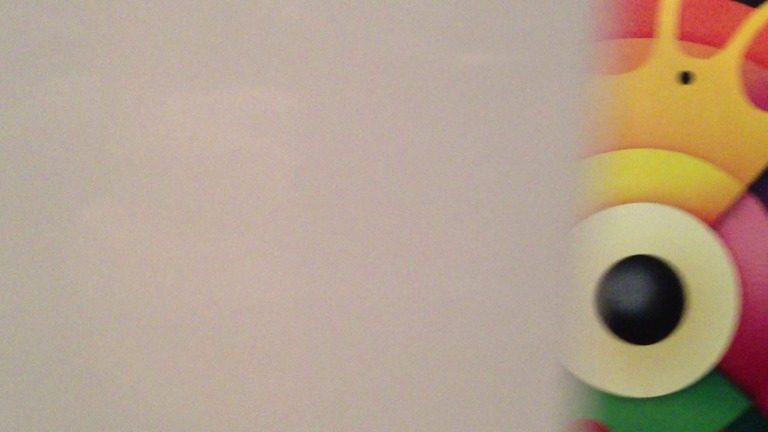}
 \includegraphics[width=0.49\columnwidth]{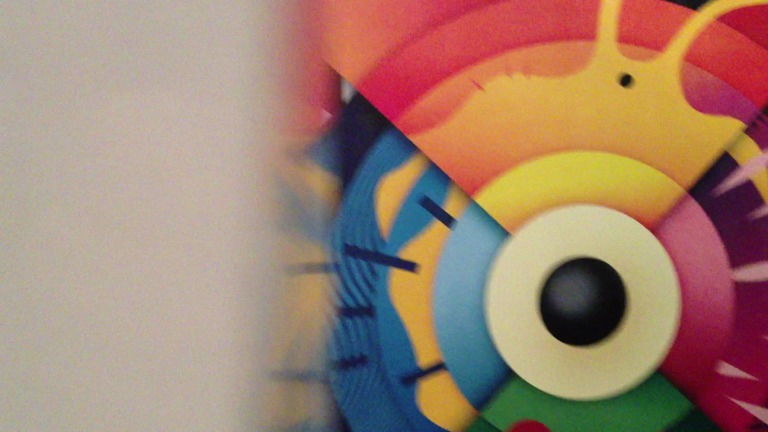}
 \includegraphics[width=0.49\columnwidth]{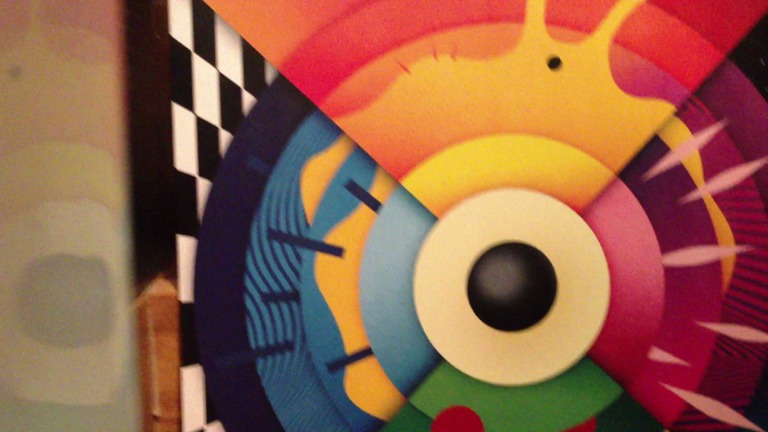}
 \caption{Sample images from \texttt{vinyl} sequence.}
 \label{fig:vinyl_images}
\end{figure}

\begin{figure}
 \includegraphics[width=\columnwidth]{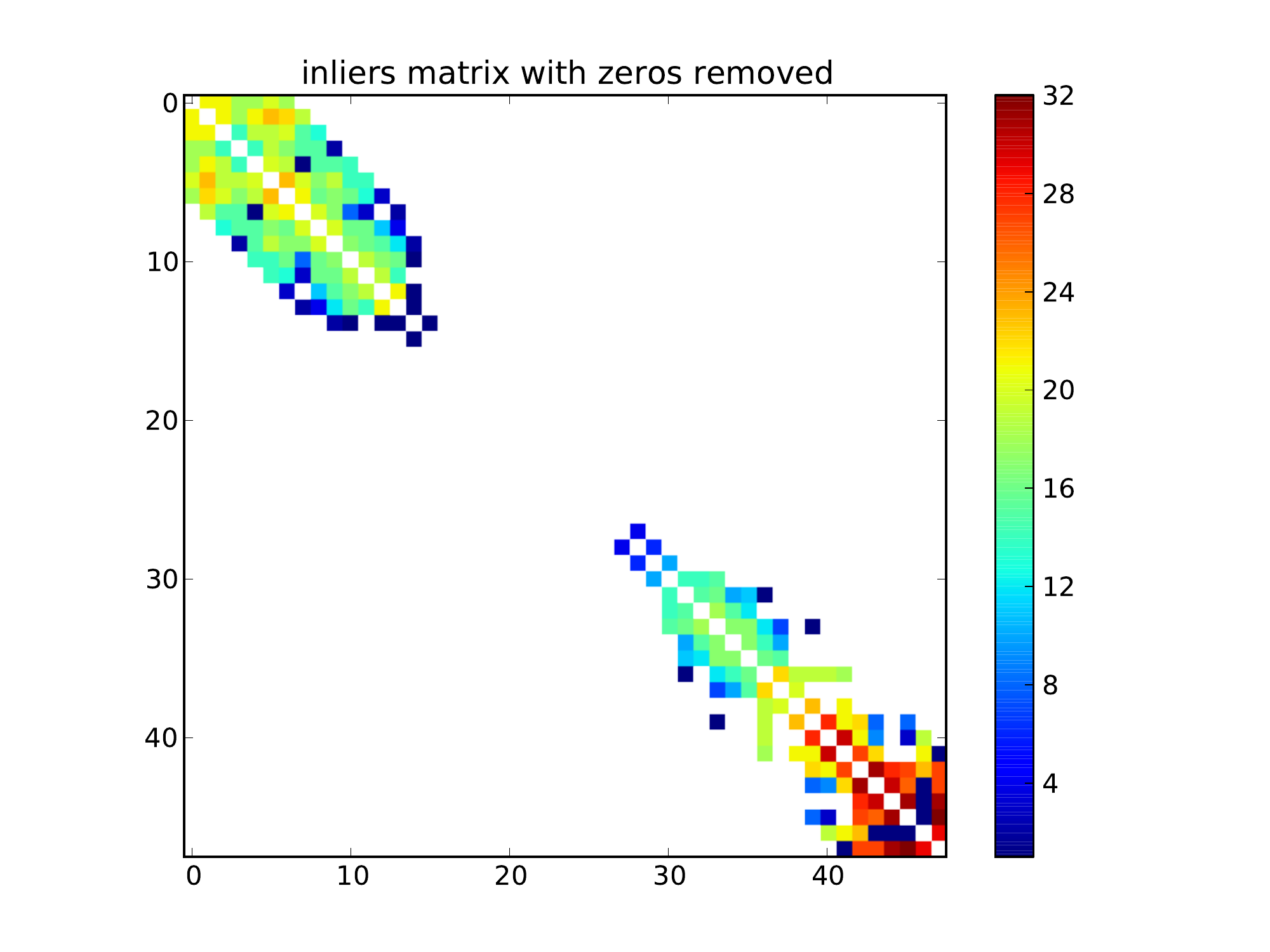}
 \caption{Number of inliers from vinyl sequence.
 Note two separate ``islands''.}
 \label{fig:vinyl_inliers}
\end{figure}

The Vinyl sequence contains an obstruction which does not trigger Viewfinder Alignment's corner detection (\figref{fig:vinyl_images}).
This causes a large region with zero inliers (\figref{fig:vinyl_inliers}) which prevents the start and end of this sequence from connecting to each other, and therefore renders it impossible to create a single set of camera paths to be loaded into our viewer.

\newpage
\subsubsection*{Grating}

\begin{figure}
 \includegraphics[width=0.49\columnwidth]{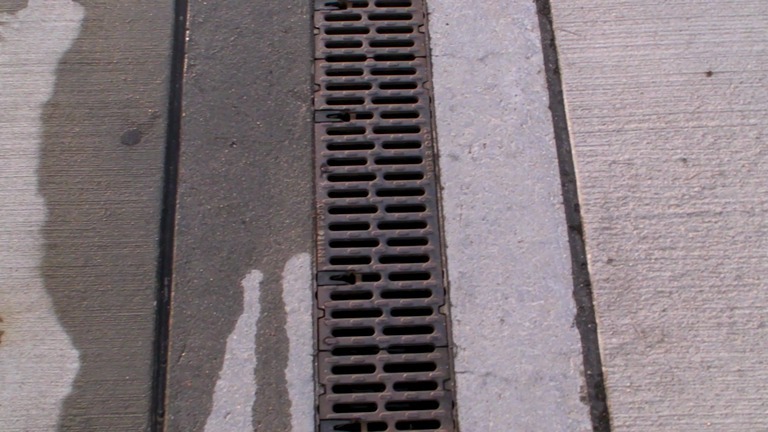}
 \includegraphics[width=0.49\columnwidth]{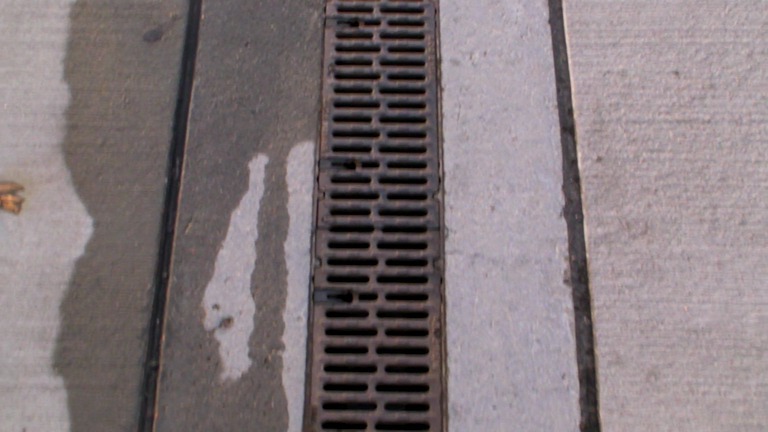}
 \includegraphics[width=0.49\columnwidth]{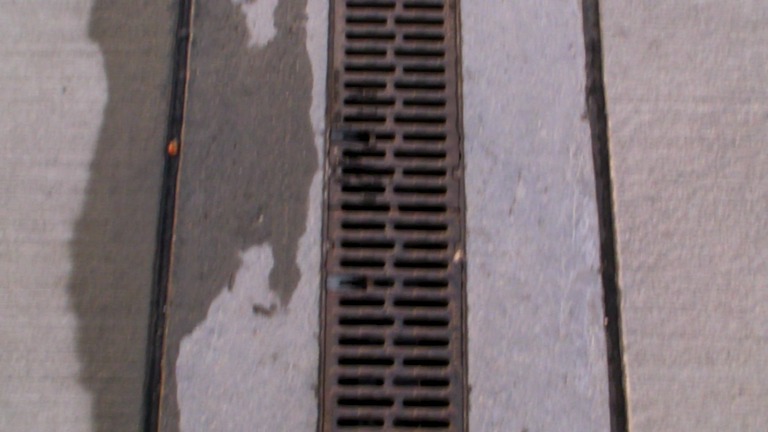}
 \includegraphics[width=0.49\columnwidth]{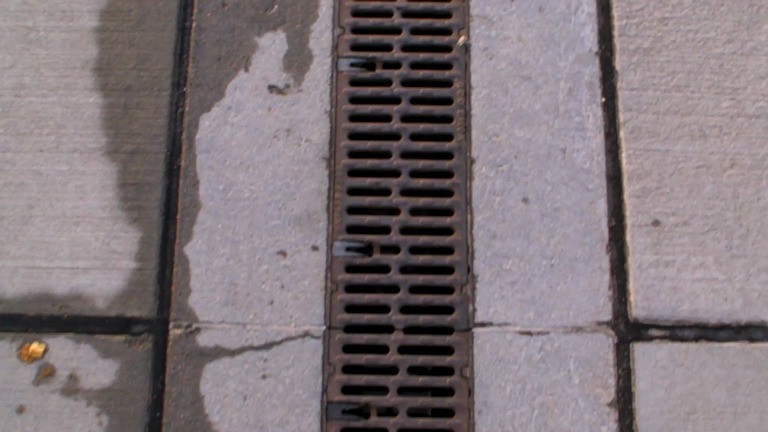}
 \includegraphics[width=0.49\columnwidth]{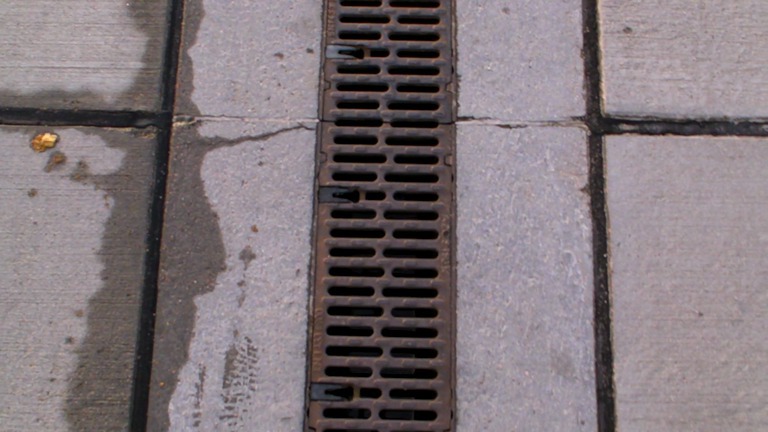}
 \includegraphics[width=0.49\columnwidth]{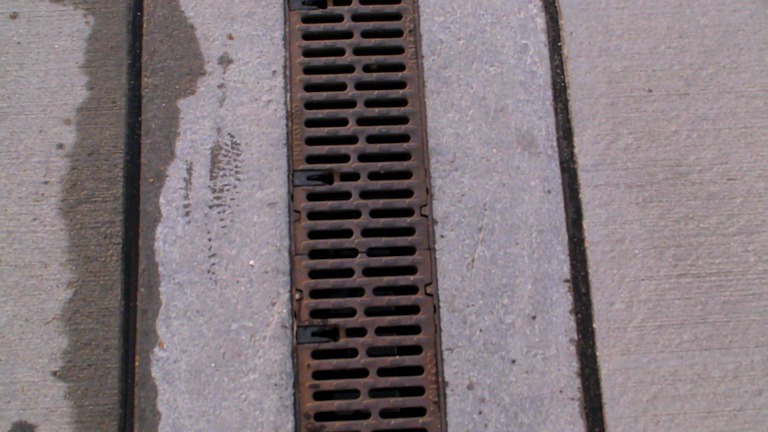}
 \caption{Sample images from \texttt{grating} sequence.}
 \label{fig:grating_images}
\end{figure}

\begin{figure}
 \includegraphics[width=\columnwidth]{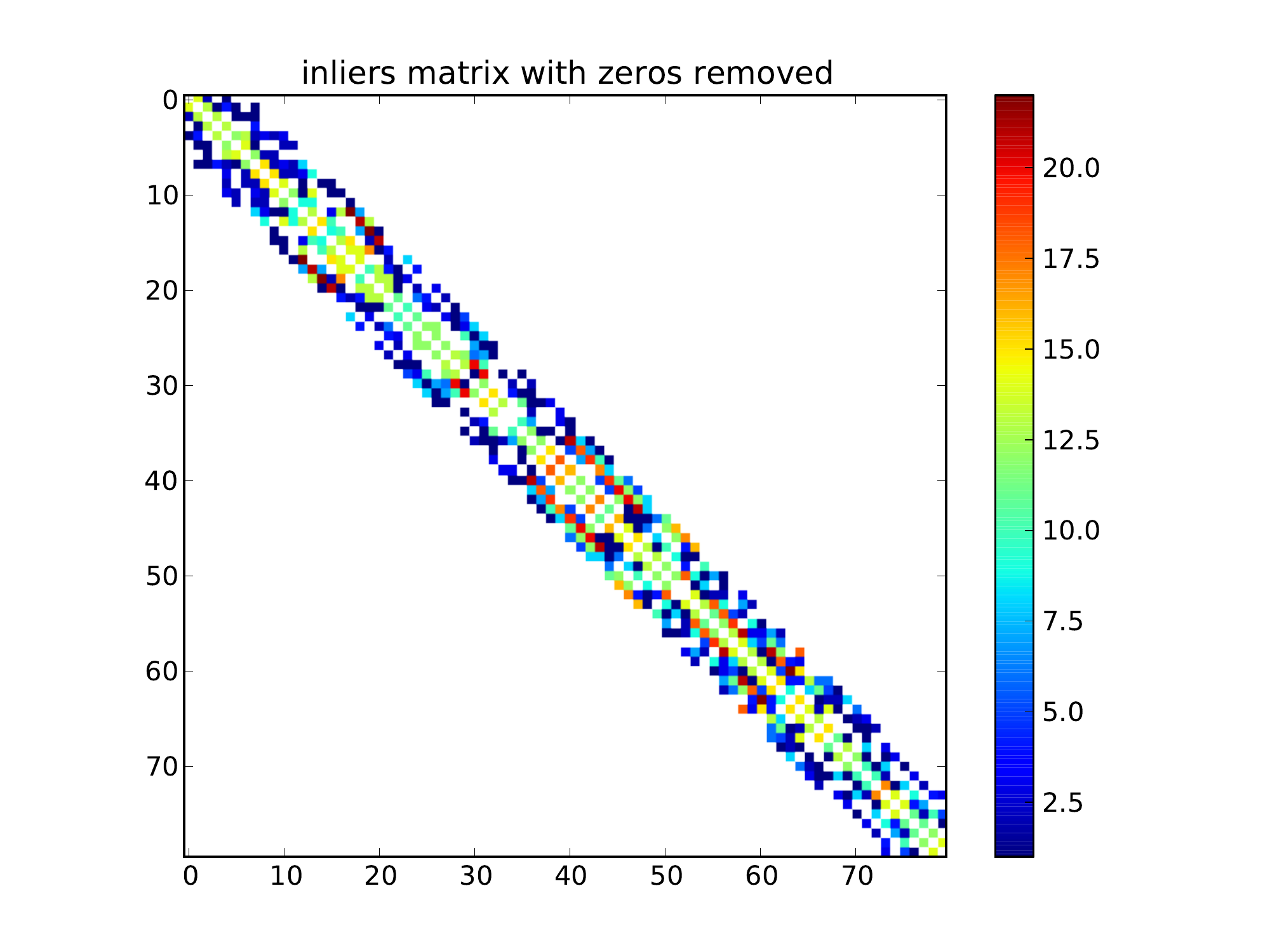}
 \caption{Number of inliers from grating sequence.}
 \label{fig:grating_inliers}
\end{figure}

\begin{figure}
 \includegraphics[width=\columnwidth]{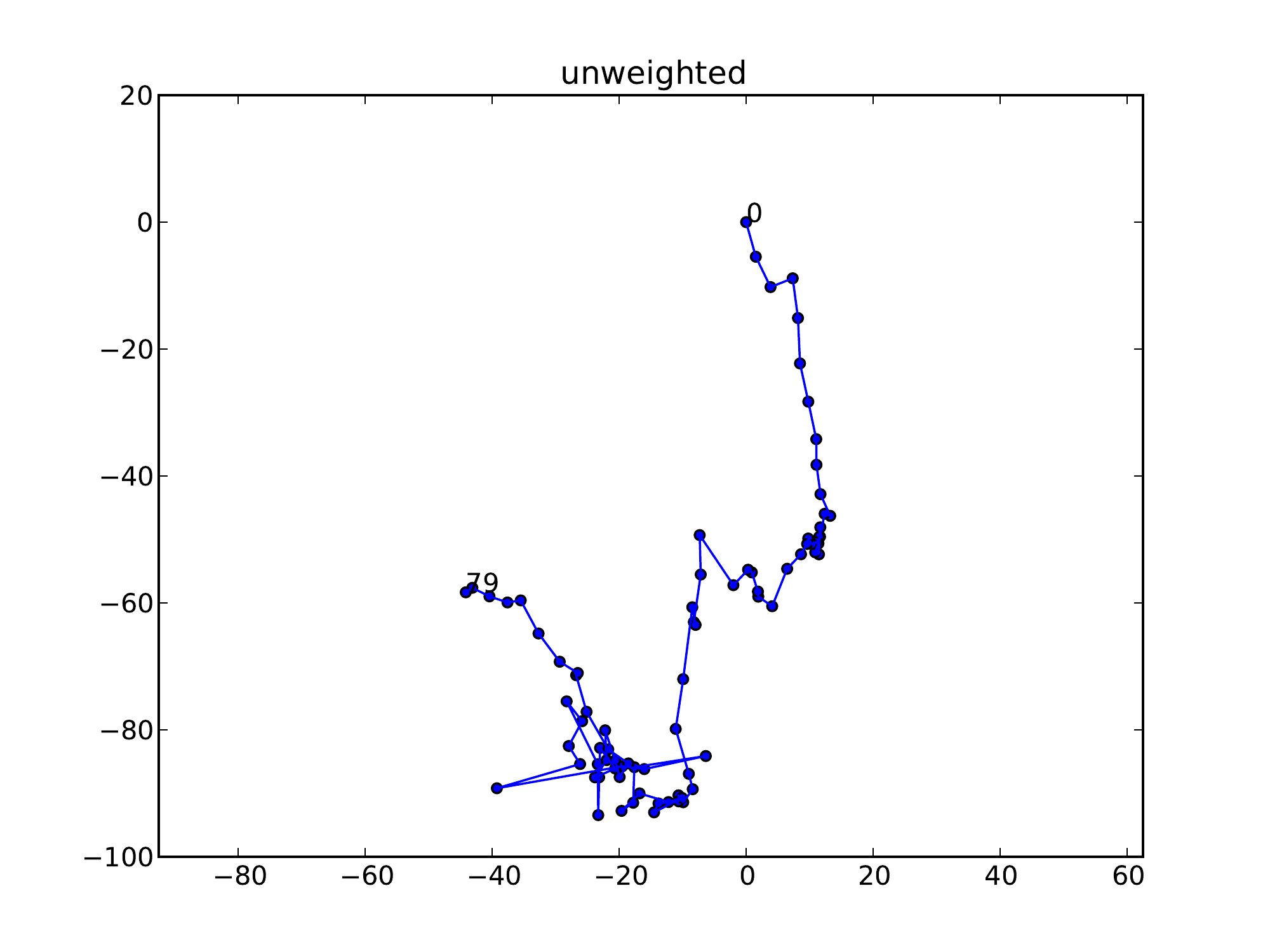}
 \caption{Regularized camera locations for grating, unweighted.
 $0$ is the first frame; $79$ the final frame.}
 \label{fig:grating_unweighted}
\end{figure}

\begin{figure}[h]
 \includegraphics[width=\columnwidth]{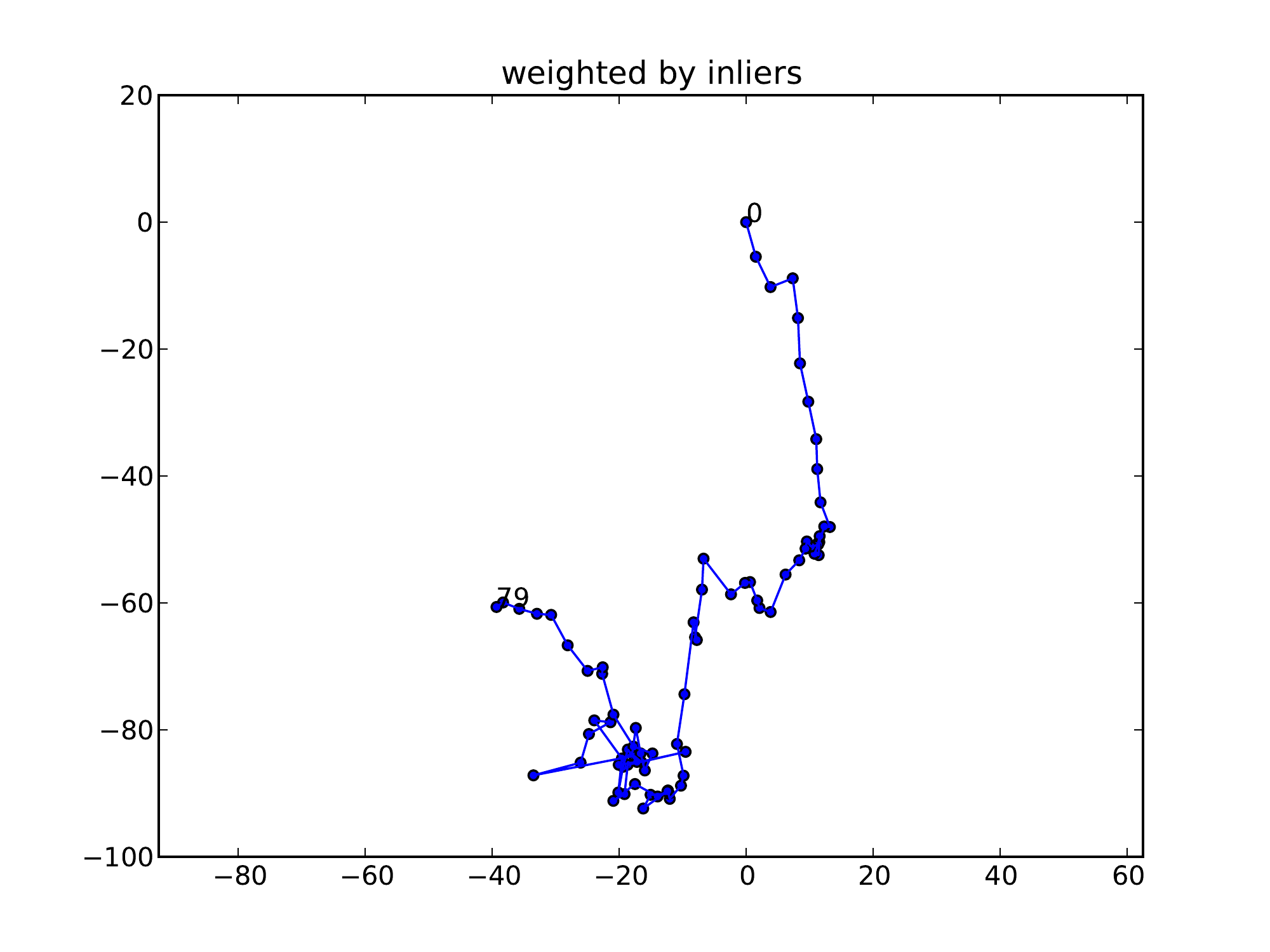}
 \caption{Regularized camera locations for grating, weighted by inlier count.
 $0$ is the first frame; $79$ the final frame.}
 \label{fig:grating_weighted}
\end{figure}

The Grating sequence is an interesting case because it contains very easily localisable vertical edges but relatively few unambiguous horizontal edges (\figref{fig:grating_images}).
Viewfinder Alignment finds sufficient transforms to regularise the 80 frame segment all together as one connected cluster, leading a promising looking inliers graph (\figref{fig:grating_inliers}).
However, a small number of incorrect matches corrupt the whole regularisation, resulting in final camera locations shown in \figref{fig:grating_unweighted} and \figref{fig:grating_weighted}.
The correct arrangement should be a roughly straight vertical line.
It seems likely that the lack of strong horizontal edges meant that when an incorrect alignment was proposed and approved by the corner correspondence stage of VfA.
Corners are deemed as inliers based on whether a given shift puts them on top of each other, not based on any kind of descriptor based on the visual appearance of whatever was originally detected as a corner.

In this situation potentially even adding corner descriptors to the algorithm would not remedy the situation, because it seems like most corners are liable to be detected on either the drain or the two grooves next to it, and any hypothetical descriptor computed on these locations is likely to be visually similar to another descriptor computed somewhere else on the drain / groove.
The best movement cues in this scene are the water stains on the floor, which are non-repeating, but these are not captured well by the VFA digest.

\section{Further Swipe Mosaic Results}
\label{sec:further_results}

\subsection{Synthetic Satellite footage dataset}

\begin{figure*}
 \begin{subfigure}[t]{0.98\textwidth}
  \begin{subfigure}[t]{0.32\textwidth}
   \includegraphics[width=\columnwidth]{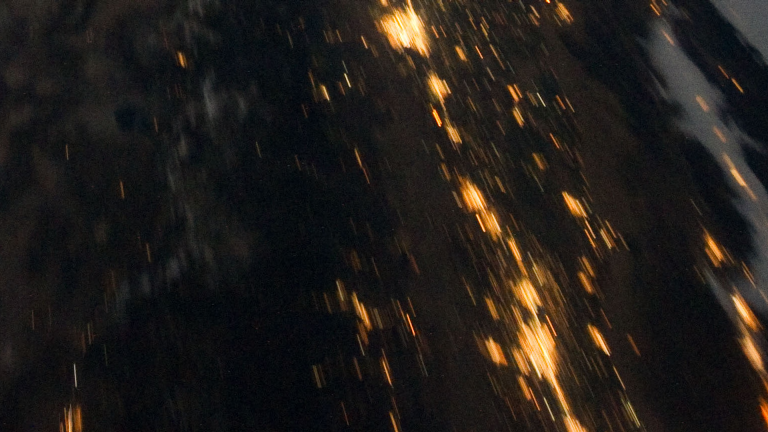}
   \label{fig:iss_example_1}
  \end{subfigure}
  \begin{subfigure}[t]{0.32\textwidth}
   \includegraphics[width=\columnwidth]{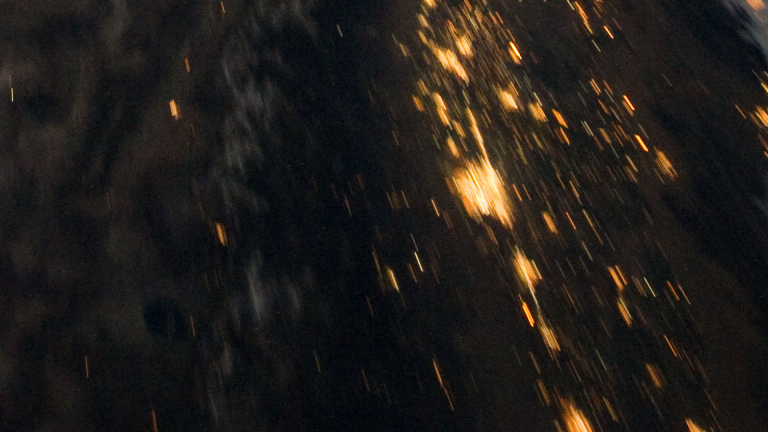}
   \label{fig:iss_example_2}
  \end{subfigure}
  \begin{subfigure}[t]{0.32\textwidth}
   \includegraphics[width=\columnwidth]{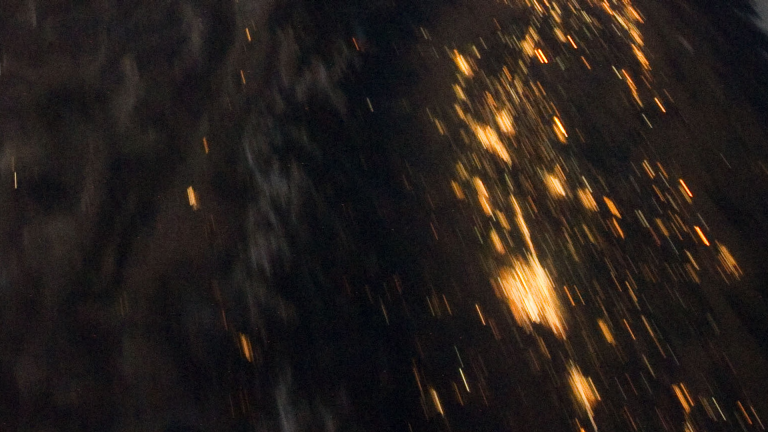}
   \label{fig:iss_example_3}
  \end{subfigure}
 \end{subfigure}
 \caption{Example images from \dataset{ISS} sequence, from the right hand edge of the image grid.
 Note how structures travel diagonally down to the right between frames -- this is due to the curvature of the earth from the ISS' vantage point.}
 \label{fig:iss_images}
\end{figure*}

As well as the real video sequences, we converted a timelapse video of the Earth recorded from the International Space Station into a Swipe Mosaic.
We first constructed an intermediate video by cropping out a thin horizontal strip from the bottom of each image and splitting each strip into eight overlapping images.
A virtual camera was moved back and forth along the strips, moving forward in time upon reaching the end, to give these cut up frames a nominal temporal ordering.
We run pairwise prediction, regularization and translational loop closure on this sequence, and know that the images should ideally be estimated to form a regular rectangular grid.

\begin{figure}
 \centering
 \begin{subfigure}[t]{0.9\columnwidth}
  \includegraphics[width=\columnwidth]{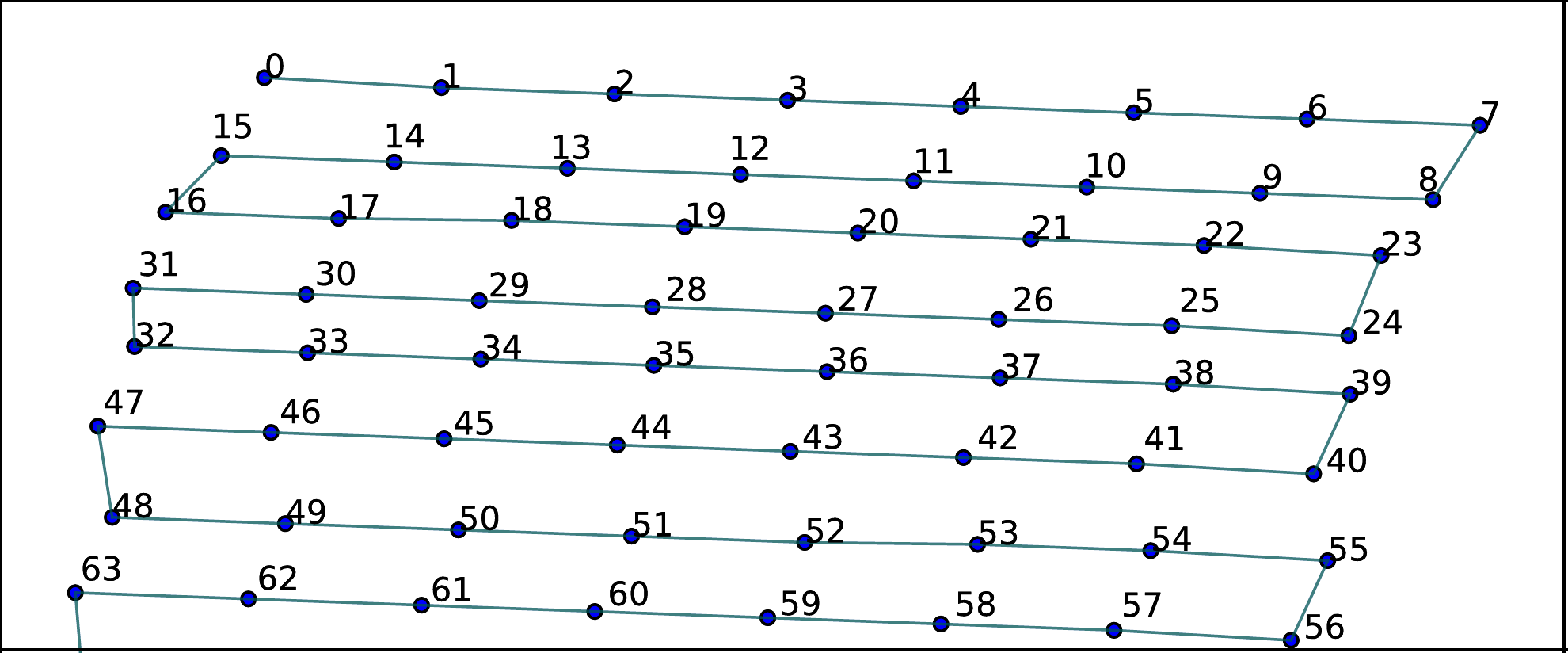}
  \caption{\footnotesize Without loop closure.}
  \label{fig:iss_without_loop_closure}
 \end{subfigure}
 \begin{subfigure}[t]{0.9\columnwidth}
  \includegraphics[width=\columnwidth]{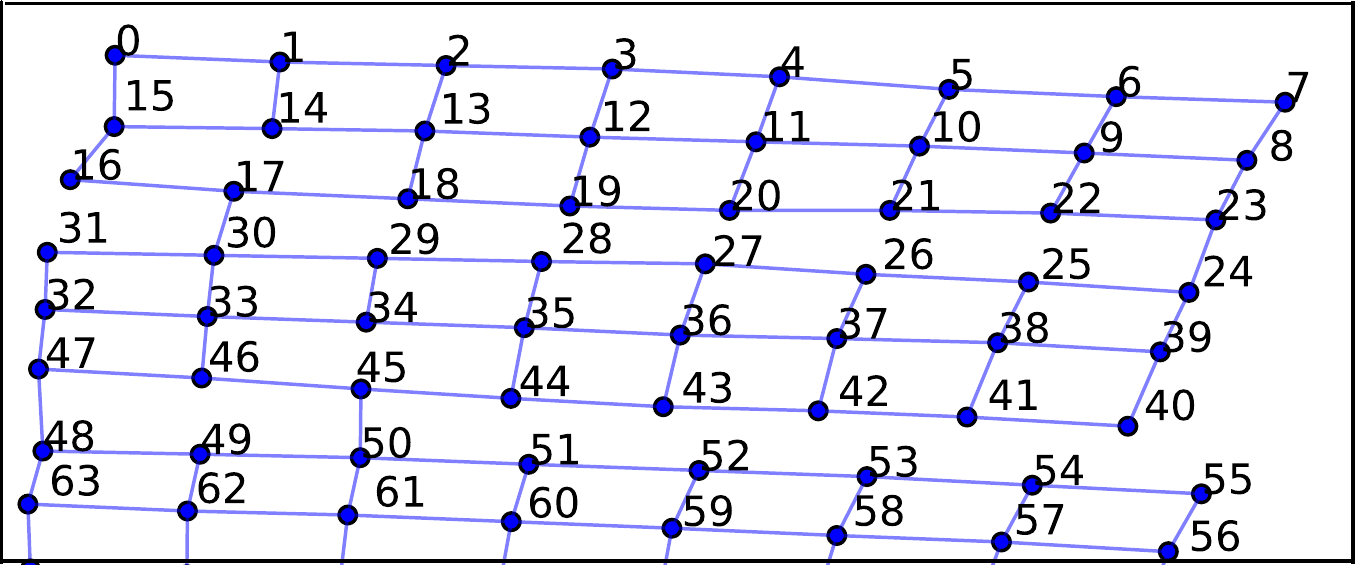}
  \caption{\footnotesize With loop closure.}
  \label{fig:iss_with_loop_closure}
 \end{subfigure}
 \caption{\footnotesize Loop closure on \dataset{iss}.
 For this sequence only, the loop point detection parameter was set to $10$ to allow for shorter loops.
 Only locations of the initial $64$ frames are shown for clarity.}
 \label{fig:iss}
\end{figure}

It is hoped each image would know from the translational RRF that their neighbors on the same strip were at a purely horizontal offset, and neighbors on a different strip were at a purely vertical offset.
The output of the first regularization step is shown in \figref{fig:iss_without_loop_closure}.
The arrangement is approximately what we would have hoped for, but the locations as a whole ``lean'' to one side.
This can be explained by noting that for the images at either end of the strip, when the virtual camera moves ``up'' or ``down'', the overlapping pixel data between the image at either end of this link will actually move diagonally, because all the earth's surface appears to be moving away from the focus of expansion.
Because the strips were not symmetrically cropped (the main goal of the cropping was to remove the visible parts of the ISS which appeared in the frame, of which there was more on one side) we see that the ``up-down'' links  such as $(7, 15)$ push the entire system to the left to a greater extent than the links on the other side such as $40, 48$.

Ideally, our loop closure step should (with slightly modified thresholds to account for the shorter loops present in this artificial scene versus a real scene) detect loop points between every image and its corresponding vertical neighbors, \ie link $(0, 15), (1, 14), (2, 13)$ \etc..
After incorporating a pairwise prediction from each and re-regularizing, we would hope to see the same overall grid structure, but with less of the horizontal skew visible in \figref{fig:iss_without_loop_closure}.
The result of automatic  loop closure is shown in \figref{fig:iss_with_loop_closure}.
Loop points have been found in the majority of places we hoped to see them.
The output is not perfect but the transitions between, for example, images $3$, $12$ and $19$ are closer to vertical than before the loop closure, and so are correspondingly improved in the visualization.
For this dataset, vertical would be the ideal answer.
Even with the imperfect results, the \dataset{iss} sequence is easy to navigate as a Swipe Mosaic.

\section{Implementation Details}

\subsection{Gabor Filter Bank}

As mentioned in the the \emph{Feature Computation} section of the main paper, a bank of Gabor filters are used as part of the feature computation process.
Each filter is computed from the product of a Gaussian and a sinusoid, according to \eqref{eq:gabor_filter}. 
\begin{align}
 \label{eq:gabor_filter}
 g(x, y; \lambda, \theta, \sigma, \gamma) & = \exp\left( -\frac{\hat{x} ^ 2}{2\sigma^2} - \frac{\hat{y} ^ 2}{2\sigma_y^2}\right)
  \exp\left( \frac{2\pi\hat{x}}{\lambda}\right) \\
  \hat{x} & = x \cos \theta + y \sin \theta\\
  \hat{y} & = -x \sin \theta + y \cos \theta\\
  \sigma_y & = \frac{\sigma}{\gamma} \\
\end{align}
$\lambda$ represents the wavelength of the sinusoid, $\theta$ is the orientation of the sinusoid (the orientation parameters allows the detection of multimodal ridges at different angles), $\sigma$ represents the standard deviation of the Gaussian, and $\gamma$ controls how this standard deviation varies in the $x$ and $y$ directions (\ie creating an elliptical function).
The ranges of values used for these parameters is specified in \ref{table:gabor_params}.
For each configuration of parameters, the filter is created as wide (in pixels) as necessary to encompass 3 standard deviations for the Gaussian.

\begin{table}
 \begin{tabular}{l|l|l|l|l}
  $\lambda$ & $\theta$ & $\sigma$ & $\gamma$ \\ \hline
  $100$ & $0$ & $4$ & $1$ \\
  $10$ & ${0, \frac{\pi}{4}, \dots, \frac{7}{4}\pi}$ & $2$ & $1$ & \\
  $10$ & ${0, \frac{\pi}{4}, \dots, \frac{7}{4}\pi}$ & $2$ & $0.5$ & \\
  $10$ & ${0, \frac{\pi}{4}, \dots, \frac{7}{4}\pi}$ & $3$ & $1$ & \\
  $10$ & ${0, \frac{\pi}{4}, \dots, \frac{7}{4}\pi}$ & $3$ & $0.5$ & \\
 \end{tabular}
 \caption{Parameters used to generate the Gabor filter bank.}
 \label{table:gabor_params}
\end{table}

\end{appendices}

\end{document}